\documentclass[11pt,a4paper]{article}
\usepackage[margin=25mm]{geometry}
\usepackage{fontspec}
\usepackage{amsmath,amssymb,bm}
\usepackage{graphicx,float,caption}
\newdimen\PulseNeededSpace
\newdimen\PulseRemainingSpace
\newcommand{\Needspace}[1]{\par\penalty-100\begingroup
  \PulseNeededSpace=#1\relax
  \PulseRemainingSpace=\pagegoal
  \advance\PulseRemainingSpace by -\pagetotal
  \ifdim\PulseRemainingSpace<\PulseNeededSpace\newpage\fi
  \endgroup}
\usepackage{algorithm}
\usepackage{algpseudocode}
\algrenewcommand{\algorithmicrequire}{\textbf{Input:}}
\algrenewcommand{\algorithmicensure}{\textbf{Output:}}
\usepackage{booktabs,longtable,array,calc}
\usepackage{enumitem}
\setlist{nosep,leftmargin=*}
\usepackage[numbers,sort&compress]{natbib}
\usepackage[table]{xcolor}
\definecolor{PulseBlue}{HTML}{1764D8}
\definecolor{PulseTeal}{HTML}{008D8A}
\definecolor{PulseViolet}{HTML}{7435BB}
\definecolor{PulseInk}{HTML}{20324D}
\definecolor{PulsePale}{HTML}{E8F2FF}
\usepackage{titlesec}
\titleformat{\section}{\normalfont\Large\bfseries\color{PulseBlue}}{\thesection}{.7em}{}
\titleformat{\subsection}{\normalfont\large\bfseries\color{PulseBlue}}{\thesubsection}{.7em}{}
\titleformat{\subsubsection}{\normalfont\normalsize\bfseries\color{PulseBlue}}{\thesubsubsection}{.7em}{}
\arrayrulecolor{PulseBlue}
\usepackage{hyperref}
\usepackage{xurl}
\hypersetup{colorlinks=true,linkcolor=PulseBlue,citecolor=PulseBlue,
  urlcolor=PulseBlue,pdftitle={LingSante-Pulse: Unified Physiological Pretraining}}
\usepackage{bookmark}
\usepackage{fancyhdr}
\renewcommand{\headrule}{\hbox to\headwidth{\color{PulseBlue}\leaders\hrule height .6pt\hfill}}
\fancypagestyle{PulseTitle}{%
  \fancyhf{}%
  \fancyhead[L]{\raisebox{-.2\height}{\includegraphics[height=26pt]{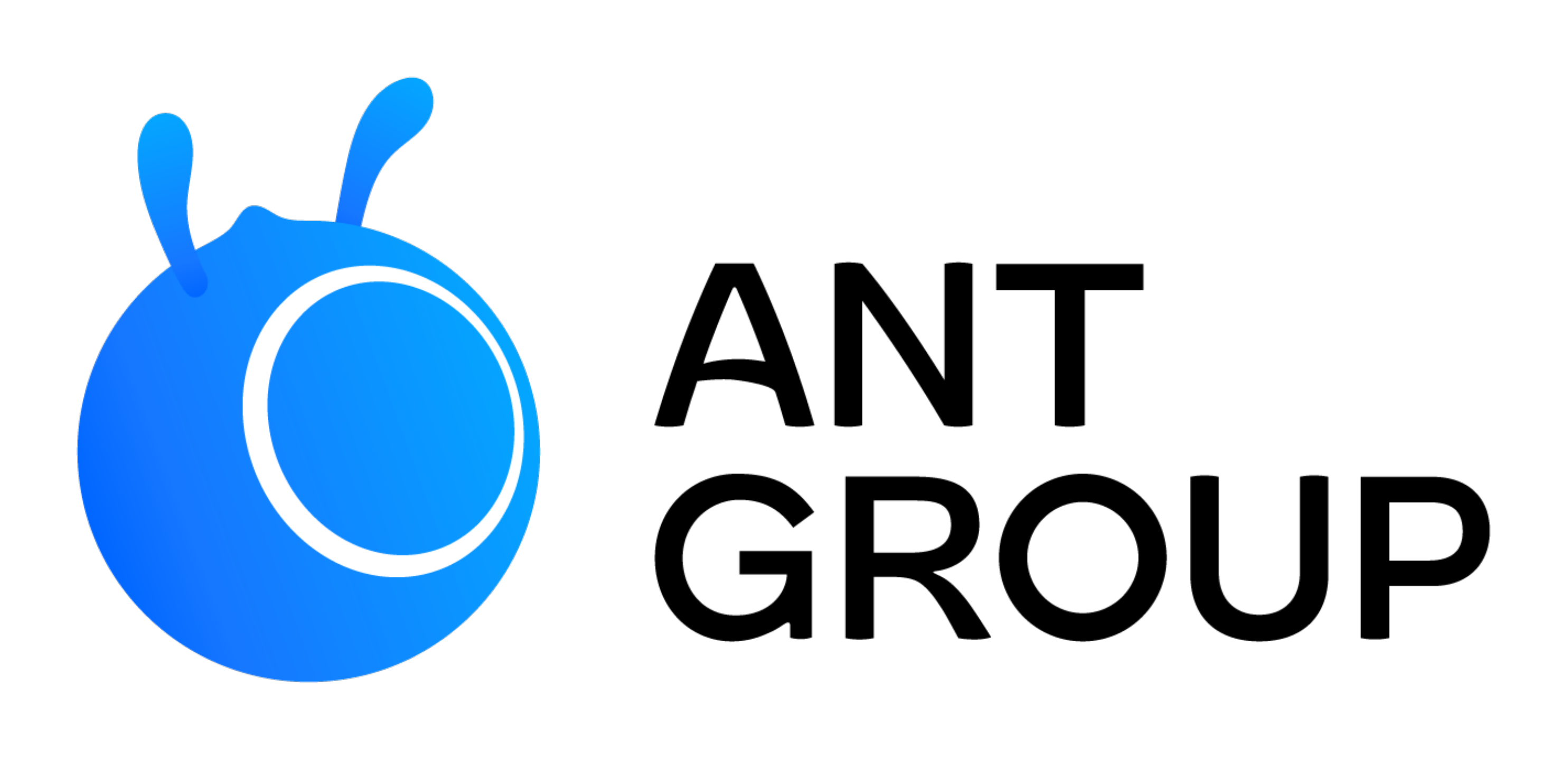}}}%
  \fancyhead[R]{\small\sffamily\color{black}Technical Report}%
  \fancyfoot[C]{\thepage}%
}
\makeatletter
\renewcommand{\maketitle}{\begingroup\centering
  {\LARGE\bfseries\color{black}\@title\par}
  \vspace{.7em}{\color{PulseBlue}\hrule height .6pt}\vspace{.6em}
  \endgroup}
\renewenvironment{abstract}{\small\begin{center}\bfseries\color{black}\abstractname\end{center}\quotation}{\endquotation}
\makeatother
\renewcommand{\arraystretch}{1.14}
\AddToHook{env/longtable/before}{\begingroup\footnotesize}
\AddToHook{env/longtable/after}{\endgroup}
\providecommand{\tightlist}{\setlength{\itemsep}{0pt}\setlength{\parskip}{0pt}}

\usepackage{newunicodechar}
\newunicodechar{₂}{\ensuremath{_2}}
\newunicodechar{↑}{\ensuremath{\uparrow}}
\newunicodechar{↓}{\ensuremath{\downarrow}}
\newunicodechar{→}{\ensuremath{\rightarrow}}
\newunicodechar{×}{\ensuremath{\times}}
\newunicodechar{−}{\ensuremath{-}}

\title{\textcolor{PulseBlue}{LingSante-Pulse}: Unified Physiological Pretraining through Random-Feature Prediction and Cross-Modal Alignment}
\author{Linxiao Yang, Wenwei Wang, Tian Zhou, Bingqing Peng, Zihao Zhang, Zuyi Zhu, Jinjie Gu, Liang Sun}
\hypersetup{pdfauthor={Linxiao Yang, Wenwei Wang, Tian Zhou, Bingqing Peng, Zihao Zhang, Zuyi Zhu, Jinjie Gu, Liang Sun}}
\date{}
\begin{document}
\maketitle
\thispagestyle{PulseTitle}
\begin{center}
\makeatletter
{\fontsize{9.5}{11.5}\selectfont\begin{tabular}{@{}c@{}}\@author\end{tabular}}
\makeatother
\end{center}

\begin{abstract}
Clinical monitoring and wearable devices capture complementary physiological signals, but their partially overlapping modality coverage makes within-signal information and cross-modal relationships difficult to exploit jointly. We introduce LingSante-Pulse, a unified, general-purpose and flexible physiological representation framework that learns from these heterogeneous recordings. A shared, signal-conditioned encoder processes each physical channel independently and supports different numbers and combinations of signals for multimodal assessment and single-sensor inference. Pairwise cross-modal alignment, structured random-feature prediction (RFP) and masked waveform reconstruction jointly learn physiological relationships and preserve informative within-signal structure. Across 38 general physiological task configurations spanning PPG, ECG, EEG and EMG, the frozen encoder achieves the best dataset-balanced mean rank in each domain. In sleep assessment, multimodal staging on the HMC dataset and EEG-only staging on the Sleep-EDF dataset reach Macro-F1 scores of 0.7784 and 0.7941, respectively, while single-lead ECG apnea screening achieves an AUROC of 0.9183. Controlled experiments show that function prediction and waveform reconstruction provide complementary supervision for retaining temporal and rhythmic information. Representations learned under continually refreshed random supervision generalize to unseen teacher realizations, and correct auxiliary modality correspondence improves respiration-only representations. These results demonstrate that a single encoder can learn reusable physiological features across signal types, prediction targets and sensor configurations from partially overlapping recordings.
\par\noindent
\raisebox{-1pt}{\includegraphics[height=8pt]{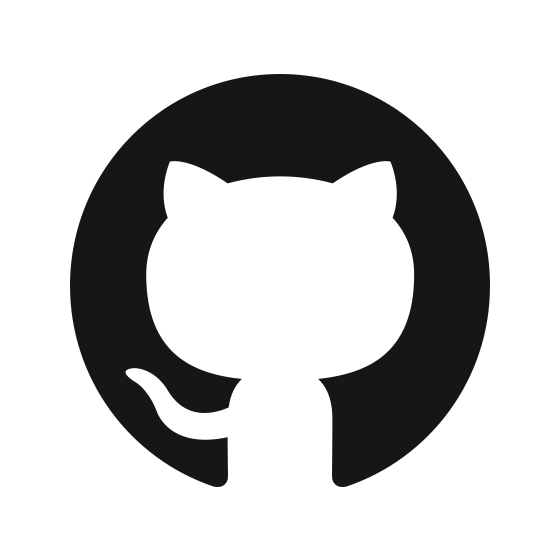}}\,\url{https://github.com/inclusionAI/LingSante-Pulse}\\
\raisebox{-1pt}{\includegraphics[height=8pt]{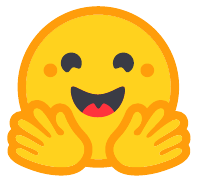}}\,\url{https://huggingface.co/inclusionAI/LingSante-Pulse-1.0}
\end{abstract}

\begin{figure}[H]
\phantomsection\label{fig:1}
\centering
\captionsetup{font=footnotesize,skip=4pt}
\includegraphics[width=\linewidth]{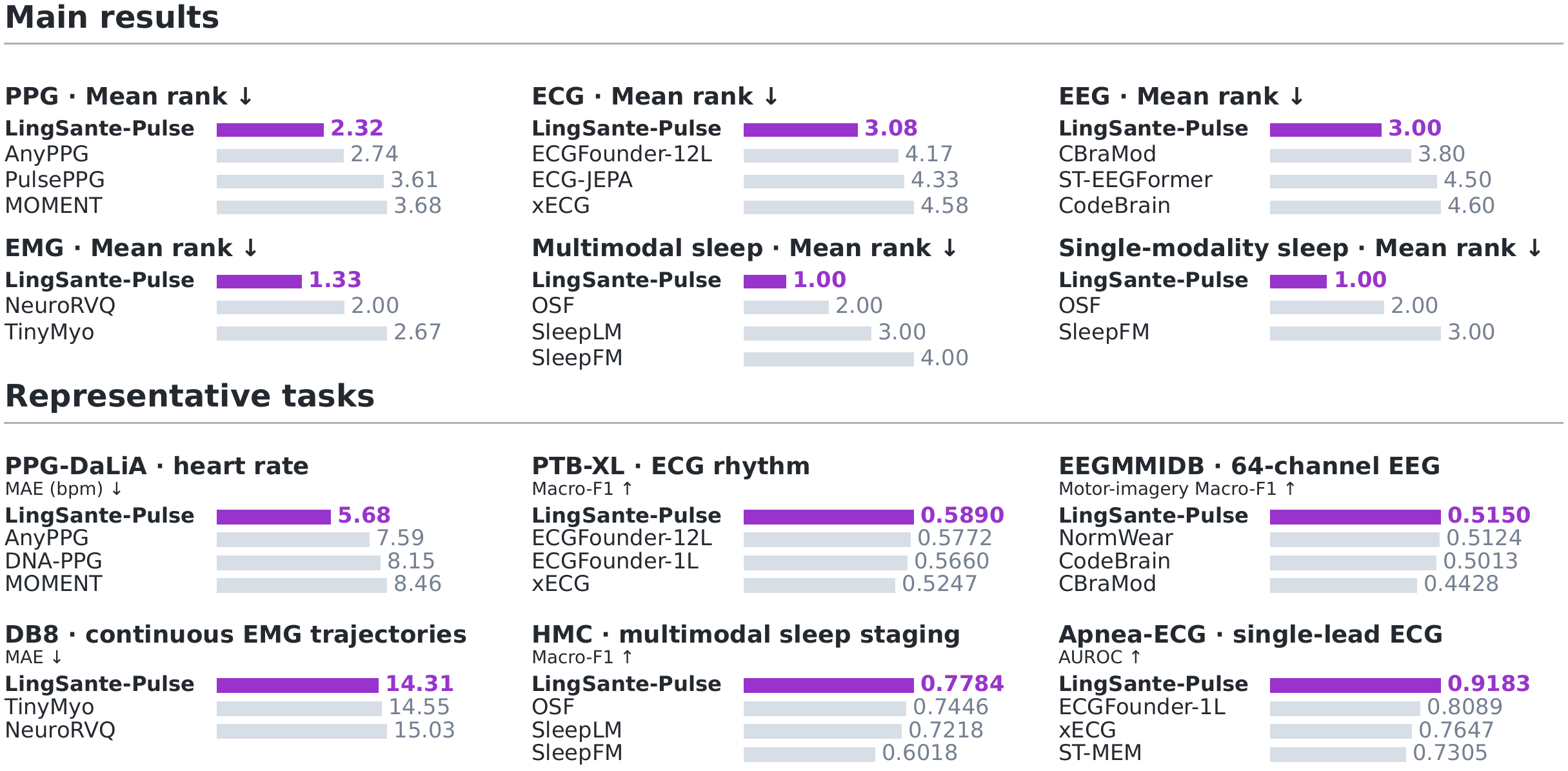}
\caption*{{\color{black}\textbf{Figure 1. Main results.}} Domain-level and sleep-staging ranks (top) and representative downstream tasks (bottom).}
\end{figure}

\clearpage
\begingroup
\setlength{\parskip}{0pt}
\tableofcontents
\endgroup
\clearpage
\section{Introduction}\label{introduction}

Clinical monitoring and wearable devices are accumulating physiological data at increasing scale. Recent foundation-model studies have used up to 40 million hours of minute-level wearable measurements and over 585,000 hours of polysomnography (PSG) recordings \citep{ref81,ref03}. These recordings provide complementary observations of human physiology, in which neural, cardiovascular, respiratory and neuromuscular processes interact across time scales. Electroencephalography (EEG) captures neural rhythms and transient events, electrocardiography (ECG) records cardiac electrical activity, photoplethysmography (PPG) measures peripheral blood-volume changes, and electromyography (EMG) reflects muscle activation. Each signal reveals a different aspect of physiological state, while relationships among signals provide information about interactions among physiological systems. Learning from both individual signals and their relationships is therefore an important goal for physiological representation learning. \hyperref[fig:1]{Figure~1} summarizes the main downstream results.

These complementary observations are distributed across recording settings with different sensor configurations. For example, PSG captures synchronized neural, cardiac, respiratory, and muscular signals, whereas a wrist wearable may provide PPG and movement measurements. Physiological datasets therefore contain partially overlapping collections of modalities, and downstream applications range from multimodal assessment to single-sensor monitoring. Alongside modality-specific foundation models that learn transferable features within individual signal domains \citep{ref05,ref06,ref07,ref08}, researchers have developed multimodal physiological models that learn relationships among synchronized signals \citep{ref03,ref04,ref02}. These approaches motivate exploiting both the information within individual signals and the relationships across modalities.

Cross-modal learning offers two complementary benefits. First, when multiple modalities are available, their complementary information can improve physiological assessment through joint modeling of signals and their relationships. Second, during pretraining, a related modality can provide additional supervision that improves representations for subsequent single-modality inference. For example, AnyPPG \citep{ref05} demonstrates the value of ECG-guided pretraining for learning transferable PPG representations, drawing on the physiological coupling between cardiac electrical activity and peripheral hemodynamics. This motivates extending cross-modal learning across partially overlapping recordings, using shared modalities to connect supervision across data sources. A target signal could then benefit from a broader set of physiological observations, including recordings in which that signal is absent.

Our goal is to build a general-purpose framework for multimodal physiological representation learning with a single shared encoder. First, by learning from signals and their relationships across data sources, the encoder should match or surpass modality-specific foundation models when given a single modality, such as dedicated EEG models on EEG-only tasks. Second, the framework should learn cross-modal relationships within a unified architecture that accommodates different numbers and combinations of channels and can incorporate new signal types during pretraining. Third, it should support applications ranging from full PSG to single-sensor monitoring and remain useful when entire modalities are unavailable or individual recordings contain gaps.

\clearpage
\begin{figure}[H]
\phantomsection\label{fig:2}
\centering
\captionsetup{font=footnotesize,skip=4pt}
\includegraphics[width=\linewidth,height=.43\textheight,keepaspectratio]{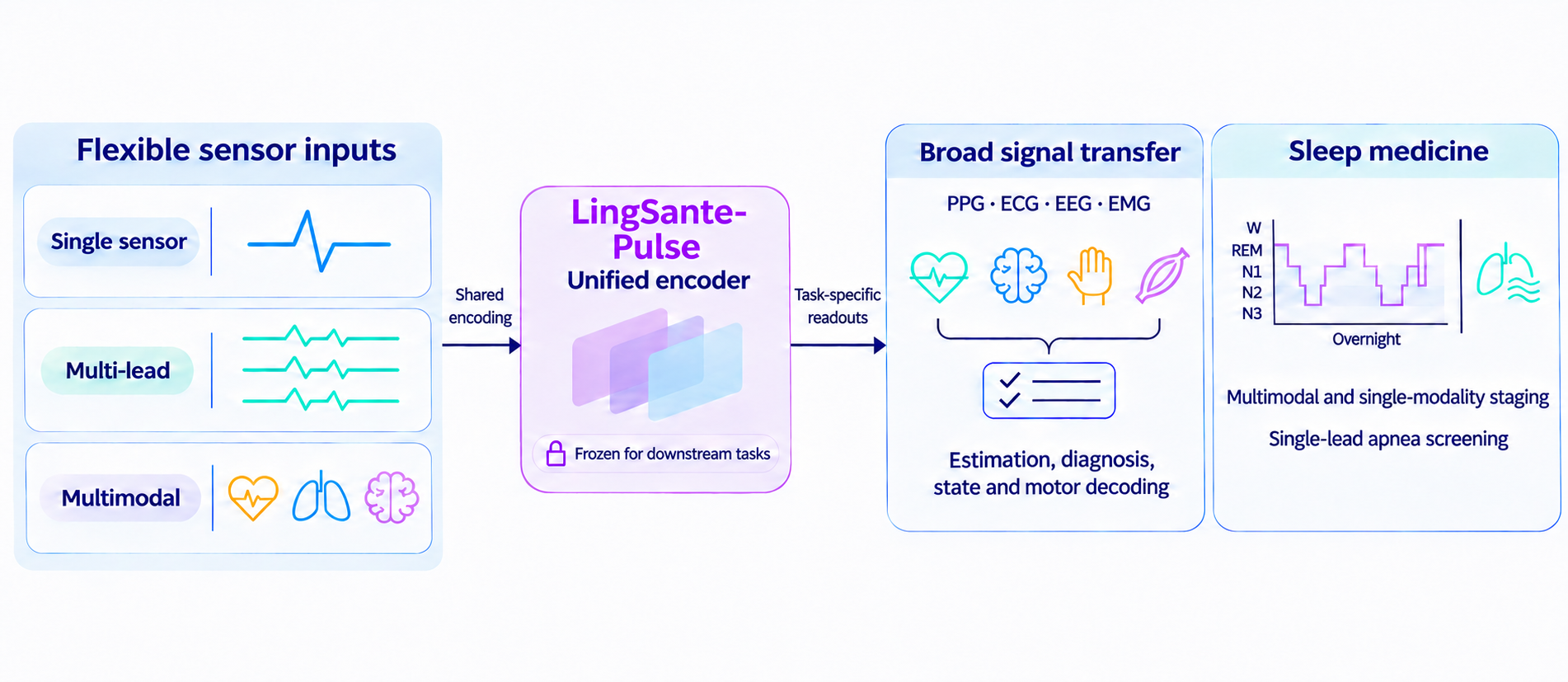}
\caption*{{\color{black}\textbf{Figure 2. Flexible inputs and downstream capabilities of LingSante-Pulse.}} The unified encoder supports single-sensor, multi-lead and multimodal inputs. Frozen representations serve physiological estimation, diagnostic and state classification, motor decoding, multimodal and single-modality sleep staging, and single-lead apnea screening through task-specific readouts.}
\end{figure}

Two challenges arise from this goal. First, how can cross-modal relationships be learned across datasets that record different subsets of signals? Each dataset supplies direct evidence for the relationships among its observed channels. Shared signals must connect these relationships so that supervision from different recordings contributes to a common representation space. Second, how can cross-modal learning preserve the distinctive information within each signal? Related physiological events produce different observations across modalities, while individual signals contain patterns such as QRS morphology in ECG and transient spindles in EEG. Pretraining must capture cross-modal relationships while retaining these signal-specific structures. Enforcing invariance between temporal crops can further suppress event timing or amplitude context relevant to downstream prediction.

We introduce LingSante-Pulse, a unified, general-purpose and flexible physiological representation framework (\hyperref[fig:2]{Figure~2}). A shared, signal-conditioned encoder processes each physical channel independently and accommodates different numbers and combinations of input signals. Pairwise synchronized alignment learns relationships from the signal pairs available in each recording. 

Beyond directly paired signals, LingSante-Pulse enables transitive relational bridging through modalities shared across recording sources. Its open pretraining framework learns from the synchronized signal pairs available in each source, allowing a shared modality to connect supervision across partially overlapping recordings. In our controlled EEG--ECG--respiration experiment (Section~\ref{transitive-transfer-through-pairwise-alignment}), pretraining uses aligned EEG--ECG and ECG--respiration pairs, while EEG and respiration are never directly paired. Adding correctly aligned auxiliary EEG--ECG observations improves representations for respiration-only inference, providing evidence that supervision can transfer through the shared ECG modality. Partially overlapping recordings therefore contribute both additional waveform data and relational supervision, extending the benefits of cross-modal learning to signals absent from some recording sources.

To preserve informative within-signal structure and improve its linear accessibility, LingSante-Pulse combines structured random-feature prediction (RFP) with masked waveform reconstruction. RFP repeatedly samples random teachers built around waveform, time-frequency and autocorrelation structure. Each teacher defines a deterministic measurement function shared across the examples within an episode. A linear readout fitted on support examples predicts teacher outputs for held-out query examples \citep{ref12,ref14}. Learning across these changing functions encourages the encoder to keep underlying signal properties broadly accessible. Masked \mbox{reconstruction} complements this objective by preserving finer temporal detail. Our task-kernel analysis characterizes the expected prediction objective and connects it to downstream regression.

We evaluate LingSante-Pulse across physiological domains and sensor configurations. The frozen encoder achieves the best dataset-balanced mean rank in every evaluated domain across 38 non-sleep task configurations spanning PPG, ECG, EEG and EMG. In sleep medicine, it supports multimodal staging on the HMC dataset and single-modality staging on the HMC dataset and Sleep-EDF dataset. Single-lead ECG representations also support minute-level apnea screening on the Apnea-ECG dataset. These evaluations demonstrate reuse of a unified physiological encoder for signal-specific decoding and multimodal sleep assessment.

Our main contributions are summarized as follows:

\begin{enumerate}
\def\labelenumi{\arabic{enumi}.}
\tightlist
\item
  \textbf{A unified physiological pretraining framework.} We develop LingSante-Pulse to integrate heterogeneous, partially overlapping recordings through a unified signal-conditioned encoder and pairwise synchronized alignment. The framework supports arbitrary numbers of input signals and both multimodal and single-sensor use.
\item
  \textbf{Structured random-feature prediction.} We introduce an episodic objective based on freshly sampled, signal-structured random functions. Support-to-query prediction encourages broad within-signal accessibility alongside cross-modal alignment and masked reconstruction.
\item
  \textbf{Theoretical characterization and mechanistic evidence.} We derive an exact task-kernel characterization and a conditional downstream-regression bound for RFP. Controlled experiments examine temporal information preservation, adaptation to refreshed teachers and transitive cross-modal transfer.
\item
  \textbf{Broad physiological transfer.} We demonstrate leading domain-level performance with frozen representations across four physiological signal domains. We further establish their reuse in multimodal sleep staging, single-modality sleep assessment and apnea screening.
\end{enumerate}

\Needspace{9\baselineskip}

\section{Related work: modeling physiological time series}\label{related-work-modeling-physiological-time-series}

Physiological representation learning combines signal-specific structure with cross-modal relationships. Existing approaches differ in how they share encoders across signals, exploit synchronized observations and define self-supervised targets.

\subsection{Signal diversity and shared physiological encoding}\label{signal-diversity-and-shared-physiological-encoding}

Signal-specific foundation models incorporate priors about the organization of their inputs. In EEG, LaBraM learns a neural tokenizer through spectrum prediction and pretrains Transformers to recover masked neural codes \citep{ref06}. CBraMod instead separates spatial and temporal dependencies through criss-cross attention and uses masked waveform reconstruction \citep{ref07}. Both address heterogeneous EEG formats while exploiting the organization of brain recordings. ECG-FM combines contrastive and generative pretraining for transferable cardiac representations \citep{ref08}. These approaches establish strong within-domain references through signal-specific tokenization, channel structure and target construction.

Shared physiological encoders extend transfer across signal types. NormWear uses channel-aware attention and a shared liaison token to learn within-sensor and between-sensor patterns across heterogeneous wearable measurements \citep{ref18}. PanLUNA represents EEG, ECG and PPG channels through a unified query set with sensor-type embeddings, enabling early fusion within a shared encoder \citep{ref19}. Its compact architecture, quantization analysis and deployment on low-power devices emphasize efficient multimodal biosignal inference. LingSante-Pulse follows this shared-encoding direction while processing each physical channel independently, using semantic channel conditioning to specialize shared transformations. Downstream tasks combine the resulting channel features according to their available sensor configuration. \hyperref[tab:1]{Table~1} summarizes these approaches alongside cross-modal learning methods.

\begin{table}[H]
\phantomsection\label{tab:1}
\centering\small
\setlength{\tabcolsep}{5pt}\renewcommand{\arraystretch}{1.18}
\caption*{{\color{black}\textbf{Table 1. Representative physiological encoders and multimodal learning approaches.}} Models discussed in Sections 2.1 and 2.2, grouped by signal coverage and representation-learning focus.}
\begin{tabular}{@{}p{3.1cm}p{2.5cm}p{\dimexpr\linewidth-5.6cm-4\tabcolsep\relax}@{}}\toprule
\rowcolor{PulsePale}Model & Signal coverage & Learning focus\\\midrule
LaBraM \citep{ref06} & EEG & Neural tokenization for generic EEG representations.\\
CBraMod \citep{ref07} & EEG & Spatial and temporal dependencies in brain signals.\\
ECG-FM \citep{ref08} & ECG & Foundation representations for electrocardiographic tasks.\\
AnyPPG \citep{ref05} & PPG with ECG guidance & Synchronized cardiac supervision for PPG representations.\\
NormWear \citep{ref18} & Wearable biosignals & Shared learning across heterogeneous wearable signals.\\
PanLUNA \citep{ref19} & EEG, ECG and PPG & Query-based early fusion in a compact shared encoder, with missing-modality inference and quantization for edge deployment.\\
SleepFM \citep{ref11,ref03} & Sleep physiology & Representation learning from synchronized sleep modalities.\\
sleep2vec \citep{ref04} & Nocturnal biosignals & Cross-modal alignment across heterogeneous sleep recordings.\\
OSF \citep{ref02} & Sleep physiology & Channel heterogeneity, objective design and multisource scaling.\\
ImageBind \citep{ref09} & General multimodal data & Shared embeddings learned from incomplete cross-modal pairing.\\
\bottomrule\end{tabular}
\end{table}

\subsection{Coupled physiology across fragmented recordings}\label{coupled-physiology-across-fragmented-recordings}

Synchronized physiological measurements provide supervision beyond the information available within an individual waveform \citep{ref83}. AnyPPG uses ECG guidance to learn PPG representations \citep{ref05}, illustrating how a paired training modality can benefit a signal used independently downstream. SleepFM learns from synchronized brain, cardiac and respiratory observations through a leave-one-out contrastive objective \citep{ref11,ref03}. The sleep2vec model develops unified cross-modal alignment for heterogeneous nocturnal biosignals and studies inference with incomplete modality sets \citep{ref04}. OSF systematically examines objective families, channel heterogeneity and multisource scaling for sleep foundation models \citep{ref02}. Together, these studies establish joint assessment and restricted-sensor deployment as complementary goals.

Partial pairing also appears in general multimodal learning. ImageBind learns a joint embedding using image-paired observations, connecting modalities without requiring every modality combination to be observed \citep{ref09}. This suggests organizing fragmented physiological recordings through shared signals: each dataset contributes the synchronized relationships it contains, and overlapping modalities connect learning across datasets. LingSante-Pulse uses observed pairs to supervise a common channel-conditioned encoder and combines this relational objective with within-signal prediction and reconstruction. Its auxiliary-alignment study examines the value of an additional training relationship by comparing correct correspondence with matched auxiliary observations under disabled or shuffled alignment.

\subsection{Temporal structure, self-supervised targets and task accessibility}\label{temporal-structure-self-supervised-targets-and-task-accessibility}

The choice of self-supervised target determines which signal distinctions receive training pressure. SimCLR learns agreement between augmented views, making augmentation design central to the learned representation \citep{ref20}. LeJEPA combines embedding prediction with isotropic Gaussian regularization \citep{ref01}. For time series, TS2Vec learns contextual timestamp representations through hierarchical contrastive learning and supports subsequence representations by aggregation \citep{ref13}. These approaches motivate examining the temporal context retained by a view: a crop may preserve recurring rhythms while changing an event's location within the original window. 

Masked autoencoders provide complementary supervision by reconstructing hidden input content from visible observations \citep{ref21}. Applied to physiological waveforms, sample-level reconstruction trains local morphology and temporal detail. Random-target prediction offers another way to define supervision directly from the input. Learning from random data projectors demonstrates that random projections can support representation learning without requiring handcrafted augmentation invariances \citep{ref12}. LingSante-Pulse builds on this direction with refreshed functions structured around waveform, time-frequency and autocorrelation measurements. It combines their prediction with reconstruction to train both broader signal attributes and local waveform information.

Task construction connects these objectives to representation meta-learning. CACTUs constructs tasks from unlabeled data \citep{ref22}, and R2-D2 differentiates through support-fitted ridge regression \citep{ref14}. Analyses of shared linear representations and train--validation splitting explain how related tasks and query supervision shape transferable features under explicit assumptions \citep{ref16,ref15}. Transferable evidence reconstruction (TER) trains longitudinal glucose representations by fitting a ridge regressor to label-free signal summaries in one group and scoring recovery on identity-disjoint individuals \citep{ref82}. RFP similarly trains cross-example recoverability, using refreshed random waveform, spectral and rhythmic functions to define its tasks. LingSante-Pulse combines this supervision with pairwise alignment and masked reconstruction.

\section{LingSante-Pulse framework}\label{ling-pulse-framework}

\subsection{Overview of unified physiological pretraining}\label{overview-of-unified-physiological-pretraining}

LingSante-Pulse learns a unified physiological representation from heterogeneous datasets that contain partially overlapping sets of signals. These observations form a sparse observation graph: signal modalities or physical leads constitute the nodes, and synchronously recorded pairs provide the observed connections. This perspective motivates learning from the relationships available within each dataset, while sharing an encoder across the entire collection.

\begin{figure}[H]
\phantomsection\label{fig:3}
\centering
\captionsetup{font=footnotesize,skip=4pt}
\includegraphics[width=\linewidth,height=.70\textheight,keepaspectratio]{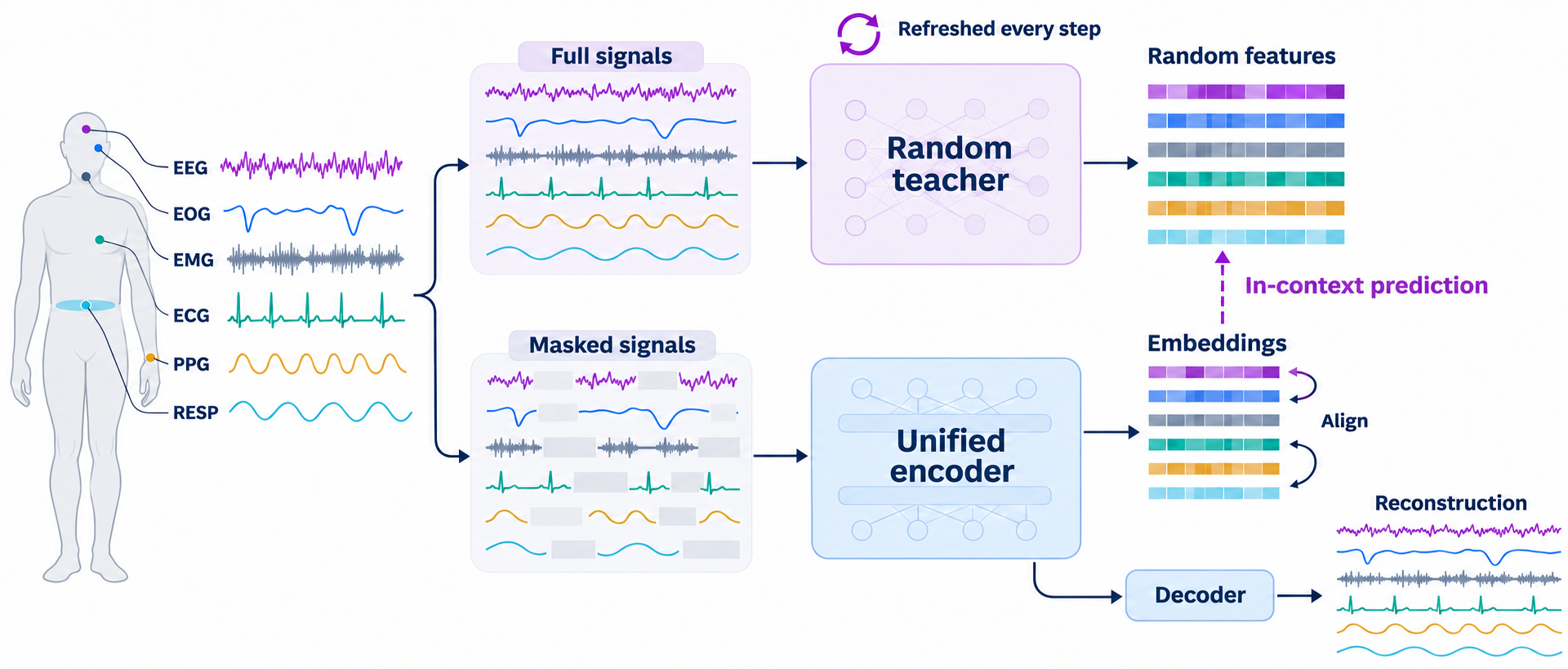}
\caption*{{\color{black}\textbf{Figure 3. High-level overview of LingSante-Pulse pretraining.}} Each channel is processed independently by the shared encoder. Complete signals enter a random teacher refreshed at every training step, while masked signals enter the encoder. Synchronized channel embeddings are aligned, and RFP predicts each channel's own teacher targets across examples. A separate decoder reconstructs masked waveform segments. Alignment uses only pairs observed in each dataset. At deployment, only the unified encoder is retained from pretraining; the random teacher, adaptation solver, alignment projector and reconstruction decoder are discarded.}
\end{figure}

Inputs comprise physiological waveform channels, such as EEG, ECG, PPG, EMG, electrooculography (EOG) and respiratory effort, together with observation masks and channel descriptions. The shared interface is not restricted to these examples or to a fixed number or combination of channels. As illustrated in \hyperref[fig:3]{Figure~3}, LingSante-Pulse processes each physical channel independently through the same signal-conditioned encoder. Channel descriptions identify the physiological modality and acquisition lead, allowing a shared backbone to accommodate diverse inputs. Pretraining combines pairwise cross-modal alignment with two within-signal objectives, structured random-feature prediction (RFP) and masked waveform reconstruction. Together, they connect complementary physiological observations while preserving the information carried by each individual signal.

\textbf{Learning relationships through pairwise alignment.} Synchronized signals provide complementary measurements of the same underlying physiological state. LingSante-Pulse aligns their representations along the observed connections, integrating relational supervision from different recording environments into a shared representation space. Training therefore draws on the available signal pairs without requiring complete multimodal recordings across datasets.

\textbf{Preserving signal information through prediction and reconstruction.} Cross-modal agreement emphasizes shared physiological information, while individual signals also contain distinctive morphology, spectral structure and rhythmic patterns. RFP encourages the encoder to retain these attributes through prediction tasks generated by continually refreshed random teachers. Although their weights are random, each sampled teacher defines a deterministic, structured function of the input waveform. Learning representations that support prediction of these functions across examples encourages diverse signal attributes to remain linearly accessible, without prescribing invariance between temporal crops. Masked waveform reconstruction complements this functional supervision by recovering hidden waveform segments, encouraging preservation of local morphology and precise temporal structure.

\textbf{Reusing one encoder across sensor configurations.} After pretraining, the unified encoder is retained for downstream use. It independently represents the channels available for a given task: individual representations support single-sensor inference, while their combination supports multi-lead and multimodal prediction. This connects heterogeneous pretraining to flexible deployment through the same learned encoder.

\subsection{Unified signal-conditioned encoder}\label{unified-signal-conditioned-encoder}

\textbf{Patch tokenization.} Let \(x_i^m\) denote waveform window \(i\) from physical channel \(m\). During pretraining, signals are resampled to 128 Hz, and each window is standardized using its observed samples. Each waveform is divided into non-overlapping temporal patches and mapped to token embeddings of width \(D\) through a shared patch projection. Natural observation masks identify missing intervals, while artificial patch masking removes additional patches to construct partially observed inputs for pretraining. The retained patches preserve their original temporal coordinates, which are incorporated through rotary position embeddings (RoPE) \citep{ref26}.

\textbf{Signal-conditioned Transformer.} LingSante-Pulse processes each physical channel independently using a shared bidirectional Transformer. Within each channel sequence, retained patch tokens can attend to both preceding and subsequent patches. A learnable summary token, CLS, aggregates window-level information, while register tokens provide additional capacity for internal computation \citep{ref25}.

To specify the physiological meaning of each input, the encoder receives a concise natural-language channel description \(c_m\), such as an EEG montage, an ECG lead or a respiratory effort channel. A frozen text encoder embeds the description \citep{ref27}, and a shared adapter maps it to a conditioning vector. Within each Transformer block \(\ell\), learned modulation heads convert this vector into channel-dependent scale and shift vectors \(\gamma_{\ell,m},\beta_{\ell,m}\in\mathbb R^D\). For a sublayer input \(v_{\ell,m}\in\mathbb R^D\), feature-wise linear modulation (FiLM) \citep{ref10} gives

\[
\widetilde v_{\ell,m}
=(\mathbf 1+\gamma_{\ell,m})\odot\operatorname{LN}(v_{\ell,m})+\beta_{\ell,m},
\]

where \(\operatorname{LN}\) denotes layer normalization, \(\odot\) denotes element-wise multiplication, and \(\mathbf 1\in\mathbb R^D\) is the all-ones vector. Separate modulation heads condition the attention and feed-forward sublayers. Channel conditioning thus allows the shared backbone to apply signal-dependent transformations while reusing its parameters across modalities and acquisition leads.

\textbf{Encoder outputs and pretraining supervision.} Let \(F_\theta\) denote the unified encoder with shared parameters \(\theta\), and let \(x_{i,\mathrm{in}}^m\) denote the waveform input after artificial patch masking during pretraining. Here, \(i\) indexes waveform windows, \(m\) identifies the physical channel, and the subscript \(\mathrm{in}\) distinguishes the masked encoder input from the original observed waveform \(x_i^m\). The encoder produces a summary representation \(h_{i,\mathrm{CLS}}^m\), \(R\) register representations \(r_{i,k}^m\), and contextualized patch representations \(u_{i,j}^m\):

\[
\left(h_{i,\mathrm{CLS}}^m,\;
\{r_{i,k}^m\}_{k=1}^{R},\;
\{u_{i,j}^m\}_{j\in\mathcal I_i}\right)
=F_\theta\!\left(x_{i,\mathrm{in}}^m;c_m\right).
\]

Here, \(k\) indexes register tokens, \(j\) indexes temporal patch positions, and \(\mathcal I_i\) contains the original temporal indices of the patch positions supplied to the encoder. All output representations have width \(D\). During downstream inference, artificial masking is disabled, and \(\mathcal I_i\) spans the full patch grid.

The summary and patch tokens exchange information throughout the Transformer. Cross-modal alignment and structured random-feature prediction (RFP) supervise the summary representation, while masked waveform reconstruction supervises the patch representations. Together, these objectives shape the encoder's window-level and temporal features.

\textbf{Readouts and multichannel composition.} From the final normalized encoder outputs, the primary window-level readout is the uniform mean of valid patch representations. Let \(\mathcal V_i\subseteq\mathcal I_i\) contain the patch positions with observed samples, excluding prefix tokens, padding and wholly missing intervals. For a channel with at least one valid patch, the patch-mean representation is

\[
h_{i,\mathrm{mean}}^m
=\frac{1}{|\mathcal V_i|}\sum_{j\in\mathcal V_i}u_{i,j}^m.
\]

The final CLS representation \(h_{i,\mathrm{CLS}}^m\) provides a complementary window-level readout, while the unpooled patch sequence supports dense temporal prediction.

For tasks with multiple available channels, each channel is encoded independently, and its window-level representation is concatenated with those of the other channels in a fixed physical-channel order. The same frozen backbone therefore supports single-channel recordings, multi-lead arrays and multimodal observations. Complete architectural configurations and preprocessing details are provided in Appendix A.

\subsection{Pairwise cross-modal alignment}\label{pairwise-cross-modal-alignment}

LingSante-Pulse learns physiological relationships from the synchronized channel pairs available within each dataset. For a batch of \(B\) pairs, let \(a\) and \(b\) denote the paired signal sides, and let \(h_{i,\mathrm{CLS}}^m\) denote the masked summary representation of example \(i\) on side \(m\in\{a,b\}\). A shared projector \(P_\phi\), with learned parameters \(\phi\), maps these representations into a \(K\)-dimensional alignment space:

\[
z_i^m=P_\phi\!\left(\operatorname{SyncBN}(h_{i,\mathrm{CLS}}^m)\right)\in\mathbb R^K,
\]

where \(\operatorname{SyncBN}\) denotes parameter-free batch normalization synchronized across the training devices.

We directly minimize the squared distance between the paired embeddings:

\[
\mathcal L_{\mathrm{align}}
=\frac{1}{4BK}\sum_{i=1}^{B}\left\|z_i^a-z_i^b\right\|_2^2.
\]

Here, \(\|\cdot\|_2\) denotes the Euclidean norm. The factor \(1/4\) gives a within-pair variance normalization: the expression equals the average squared deviation of the two embeddings from their midpoint, averaged over samples and coordinates. Minimizing this loss brings synchronized observations into a shared representation space, allowing heterogeneous datasets to contribute relational supervision through their observed signal pairs. Appendix B relates the pairwise penalty to alignment over larger observation sets.

To prevent collapse, we apply Sketched Isotropic Gaussian Regularization (SIGReg) \citep{ref01} to each side's projected representations. SIGReg compares their distributions with a standard isotropic Gaussian along randomly sampled directions. Averaging the penalties across both sides gives the distribution regularization term \(\mathcal L_{\mathrm{reg}}\). Appendix A.4 provides its exact formulation.

\subsection{Structured random-feature prediction}\label{structured-random-feature-prediction}

Structured random-feature prediction (RFP) trains the encoder through prediction tasks generated by randomly initialized, signal-structured teachers. At each training update, a fresh teacher measures the complete observed waveform, while the student encoder processes its masked counterpart. A differentiable linear learner \citep{ref14} fits a mapping from student representations to teacher outputs on support examples, and the encoder is optimized through prediction error on held-out query examples. The teacher remains fixed throughout this episode and is redrawn at the next update.

\begin{figure}[!t]
\phantomsection\label{fig:4}
\centering
\captionsetup{font=footnotesize,skip=4pt}
\includegraphics[width=\linewidth,height=.70\textheight,keepaspectratio]{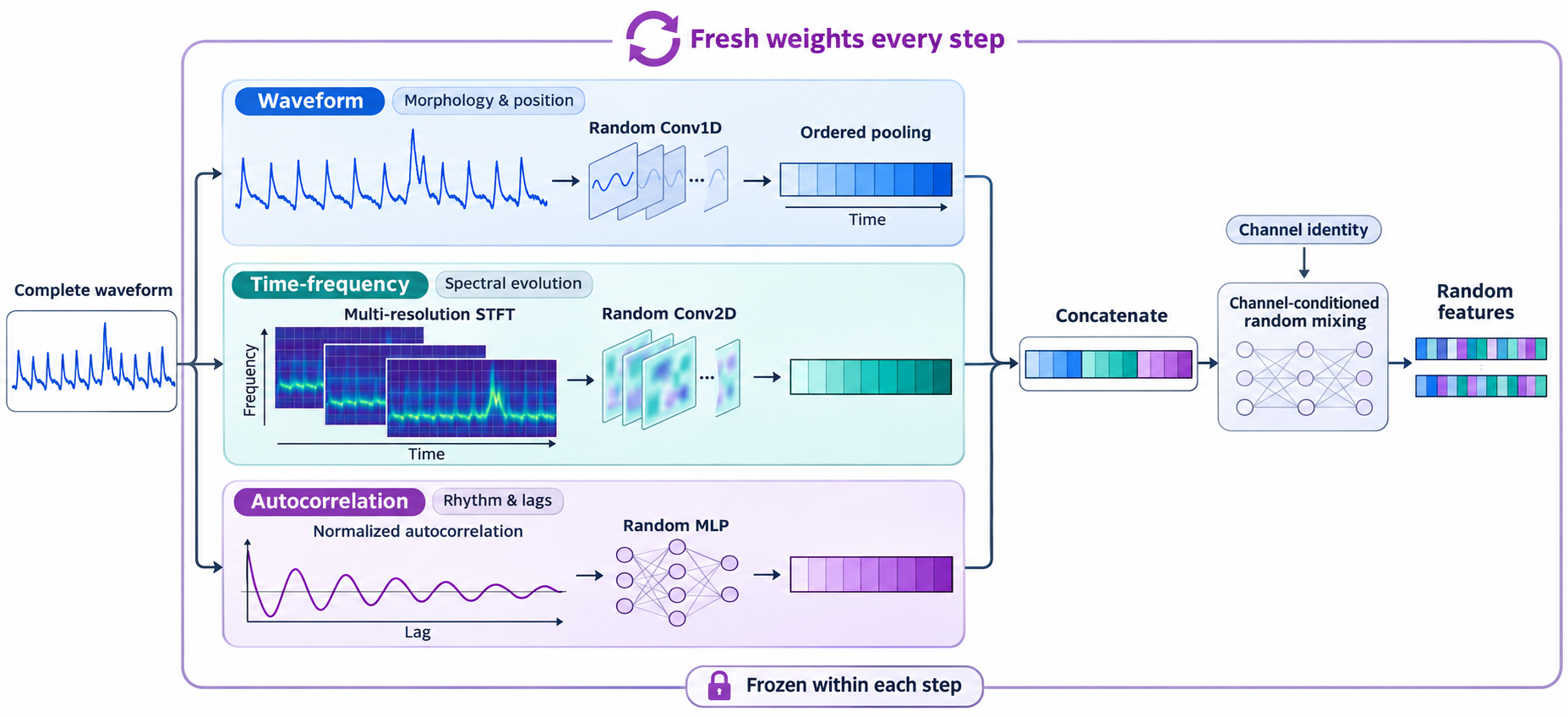}
\caption*{{\color{black}\textbf{Figure 4. Architecture of the structured random teacher.}} The complete waveform enters three parallel measurement branches: random temporal convolutions with ordered pooling, multiresolution time-frequency analysis with random spectrogram convolutions, and normalized autocorrelation with a random nonlinear network. Their features are concatenated and mixed by a channel-conditioned random residual transform and projection to form the prediction targets. Random network and mixing weights are redrawn at every training step and held fixed within that step; the signal-processing structure remains the same.}
\end{figure}

\textbf{Structured random teacher.} Let \(\tau\) index the teacher realization sampled at the current update. For waveform \(x_i^m\) and channel description \(c_m\), the teacher \(g_\tau\) produces a target vector

\[
y_i^{\tau,m}=g_\tau(x_i^m;c_m)\in\mathbb R^{d_T},
\]

where \(d_T=576\) is the target dimension. The teacher processes the complete observed waveform through three complementary branches (\hyperref[fig:4]{Figure~4}; \hyperref[tab:2]{Table~2}).

The \textbf{waveform branch} applies randomly initialized temporal convolutions to local signal patterns. Coverage-aware pooling aggregates the resulting responses into ordered temporal bins, preserving both local morphology and coarse event position within the window. The resulting measurements therefore distinguish patterns occurring at different temporal locations.

The \textbf{time-frequency branch} constructs log-magnitude spectrograms using short-time Fourier transforms (STFT) at multiple temporal resolutions and processes them through randomly initialized two-dimensional convolutions. Pooling over an ordered frequency-time grid preserves spectral composition and its evolution across the window. Multiple resolutions expose both relatively brief oscillatory events and more sustained spectral organization.

The \textbf{autocorrelation branch} computes a normalized lag-dependent correlation profile, correcting for the number of observed sample pairs at each lag. A randomly initialized nonlinear network maps this profile into rhythmic features. These measurements emphasize recurring temporal relationships and characteristic intervals, complementing the local waveform and time-frequency branches.

The three branch outputs are concatenated and combined through a channel-conditioned random nonlinear residual transform and projection. All random network and mixing weights are freshly drawn at each update, shared across support and query examples, and frozen during the episode. Appendix A.5 provides the architectural configurations and exact mixing formulation.

\Needspace{15\baselineskip}

\phantomsection\label{tab:2}
{\color{black}\textbf{Table 2. Signal structure exposed by the structured random teacher.}} All branches process the complete observed waveform. Their outputs are combined into 576 mixed target coordinates, using one teacher realization across support and query examples within each update.

{
\begin{longtable}[]{@{}
  >{\raggedright\arraybackslash}p{(\linewidth - 1pt - 6\tabcolsep) * \real{0.2308}}
  >{\raggedright\arraybackslash}p{(\linewidth - 1pt - 6\tabcolsep) * \real{0.2308}}
  >{\raggedright\arraybackslash}p{(\linewidth - 1pt - 6\tabcolsep) * \real{0.2308}}
  >{\raggedleft\arraybackslash}p{(\linewidth - 1pt - 6\tabcolsep) * \real{0.3077}}@{}}
\toprule\noalign{}
\rowcolor{PulsePale}
\begin{minipage}[b]{\linewidth}\raggedright
Branch
\end{minipage} & \begin{minipage}[b]{\linewidth}\raggedright
Input and processing
\end{minipage} & \begin{minipage}[b]{\linewidth}\raggedright
Information emphasized
\end{minipage} & \begin{minipage}[b]{\linewidth}\raggedleft
Coordinates before mixing
\end{minipage} \\
\midrule\noalign{}
\endhead
\bottomrule\noalign{}
\endlastfoot
Waveform & Conv1D and ordered temporal pooling & Local morphology and coarse event position & 256 \\
Time-frequency & Multiresolution log-STFT and Conv2D & Spectral content and its temporal organization & 256 \\
Autocorrelation & Normalized autocorrelation and MLP & Lag-dependent rhythms & 64 \\
\end{longtable}
}

\textbf{Episodic differentiable prediction.} For one waveform channel, we partition an episode into disjoint support and query sets \(S\) and \(Q\), containing \(n=|S|\) and \(q=|Q|\) examples, respectively. Stacking the student's masked CLS representations gives feature matrices \(H_S\in\mathbb R^{n\times D}\) and \(H_Q\in\mathbb R^{q\times D}\), where \(D\) is the encoder width. Corresponding teacher targets form \(Y_S^\tau\in\mathbb R^{n\times d_T}\) and \(Y_Q^\tau\in\mathbb R^{q\times d_T}\).

Features and targets are standardized coordinate-wise using support-set means and standard deviations only; tildes denote the standardized matrices. With positive ridge penalty \(\lambda_r\), the support-fitted mapping \(W_{\tau,S}^{*}\in\mathbb R^{D\times d_T}\) is

\[
\begin{aligned}
W_{\tau,S}^{*}
&=\arg\min_W\frac{1}{n}
\left\|\widetilde H_SW-\widetilde Y_S^\tau\right\|_F^2
+\lambda_r\|W\|_F^2\\
&=\left(\widetilde H_S^\top\widetilde H_S+n\lambda_r I_D\right)^{-1}
\widetilde H_S^\top\widetilde Y_S^\tau,
\end{aligned}
\]

where \(W\) is the candidate linear mapping, \(\|\cdot\|_F\) denotes the Frobenius norm and \(I_D\) is the \(D\)-dimensional identity matrix. The query prediction error is

\[
\ell_\tau(S,Q)
=\frac{1}{q\,d_T}
\left\|\widetilde H_QW_{\tau,S}^{*}-\widetilde Y_Q^\tau\right\|_F^2.
\]

The RFP objective \(\mathcal L_{\mathrm{RFP}}\) averages this error across both support-to-query directions, eligible waveform channels and training episodes. Gradients pass through the query predictions, support-fitted standardization and ridge solution into the encoder; teacher targets receive no gradients. Each update therefore rewards representations that support prediction across examples, with the linear mapping adapted to the current teacher. Appendix A.6 describes ridge adaptation and loss aggregation.

\begin{figure}[H]
\phantomsection\label{fig:5}
\centering
\captionsetup{font=footnotesize,skip=4pt}
\includegraphics[width=\linewidth,height=.70\textheight,keepaspectratio]{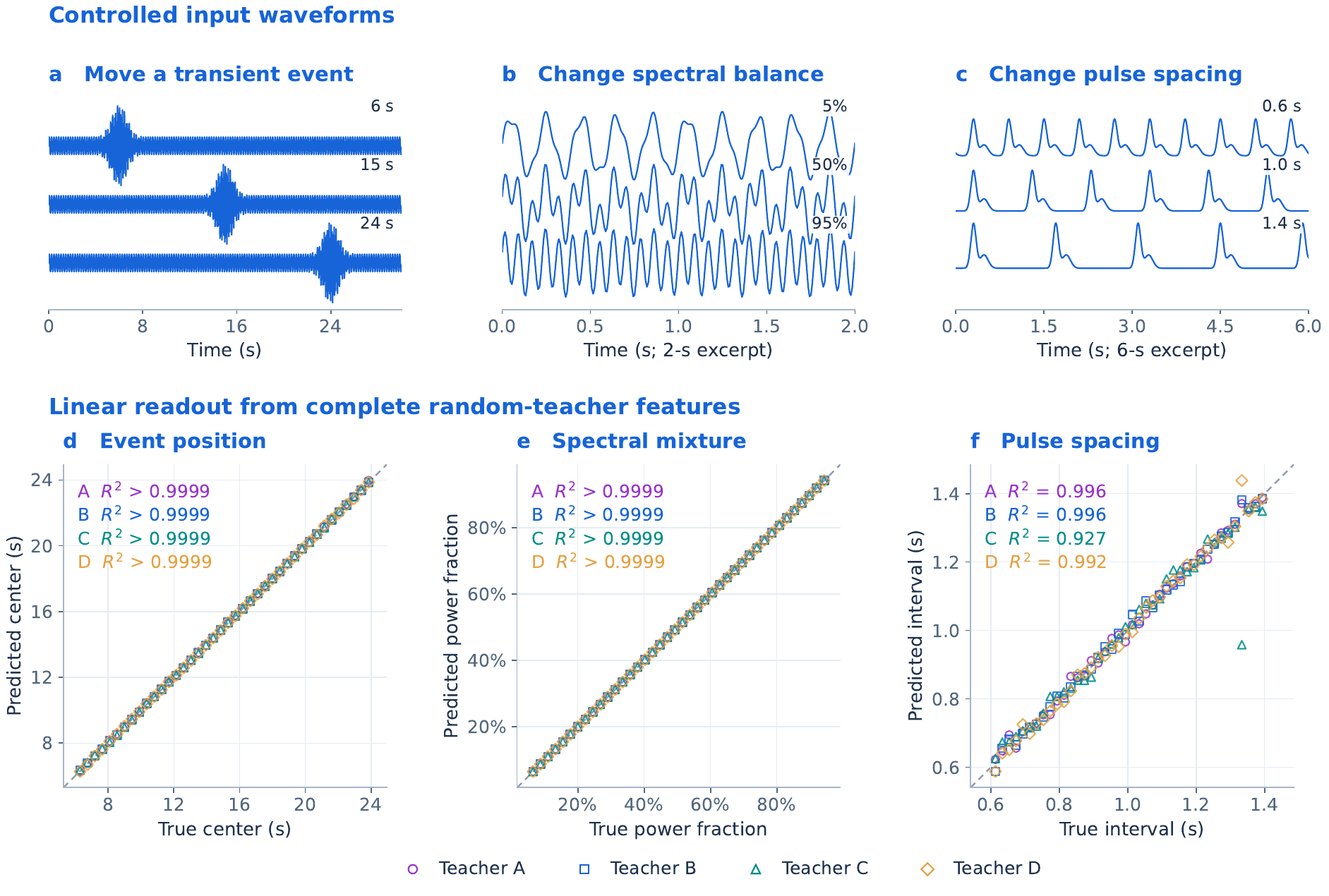}
\caption*{{\color{black}\textbf{Figure 5. Readable signal structure across random teachers.}} (a-c) Controlled waveform examples: a transient event moves from 6 to 15 to 24 seconds; high-frequency power increases from 5\% to 50\% to 95\% in a mixture of 5- and 13-Hz components; and pulse intervals increase from 0.6 to 1.0 to 1.4 seconds. All inputs are 30-second standardized waveforms; (b,c) show 2- and 6-second excerpts, respectively, with traces scaled and offset vertically for display. (d-f) Corresponding attribute recovery on held-out inputs. Colors and symbols distinguish four independently sampled, frozen structured teachers. Each teacher supplies its full 576-dimensional mixed output to a separately fitted ridge readout. Points show predictions on 40 held-out query inputs; 41 support inputs fit the readout and 40 validation inputs select its penalty. The dashed line denotes exact recovery. The diagnostic evaluates interpolation within each controlled family; attribute labels are not used during RFP pretraining.}
\end{figure}

\textbf{Readable signal structure across random teachers.} To visualize the information retained in these targets, we construct controlled waveform families in which transient event position, spectral power balance or pulse spacing varies independently. For each of four independently sampled teachers, we fit a ridge readout from its complete 576-dimensional output to the controlled attribute and evaluate on held-out parameter values (\hyperref[fig:5]{Figure~5}). The readout uses 41 support examples, with regularization selected on 40 separate validation examples and performance evaluated on 40 query examples.

All four teachers permit near-exact recovery of event position and spectral mixture (\(R^2>0.9999\)), while pulse-interval recovery reaches \(R^2=0.927\text{--}0.996\). Here, the known attribute values supervise the diagnostic ridge readout; during RFP pretraining, the prediction targets are the random teacher outputs. Across these realizations, the same temporal, spectral and rhythmic attributes remain linearly readable as the measurement functions change. Individual teachers differ in pulse-interval recovery (\hyperref[fig:5]{Figure~5f}), motivating learning across multiple realizations of the measurement family.

The support-fitted mapping adapts to each realization, while the shared encoder learns representations that make recurring signal structure accessible across examples. Section 3.6 characterizes the expected prediction objective through the teacher's task kernel, and Section 5 evaluates how this training translates into representation learning and downstream transfer.

\subsection{Masked waveform reconstruction and joint objective}\label{masked-waveform-reconstruction-and-joint-objective}

\textbf{Masked waveform reconstruction.} Alongside relational alignment and RFP, reconstruction supplies local supervision for waveform morphology and precise temporal structure \citep{ref21}. A lightweight, signal-conditioned decoder receives the contextualized patch representations, inserts learned mask tokens at artificially hidden positions, and predicts the corresponding waveform samples. All three objectives share the same masked encoder output.

Let \(\mathcal A\subseteq\{a,b\}\) contain the waveform sides eligible for reconstruction, and let \(\mathcal R_i^m\) contain the sample positions that were originally observed but artificially hidden in window \(i\) of channel \(m\). Denote the normalized target and decoder prediction at sample position \(t\) by \(x_{i,t}^m\) and \(\widehat x_{i,t}^m\), respectively. The reconstruction loss is

\[
\mathcal L_{\mathrm{MAE}}
=\begin{cases}
\displaystyle\frac{1}{|\mathcal A|B}\sum_{m\in\mathcal A}\sum_{i=1}^{B}
\frac{\displaystyle\sum_{t\in\mathcal R_i^m}
\left(\widehat x_{i,t}^m-x_{i,t}^m\right)^2}
{\max(1,|\mathcal R_i^m|)},&|\mathcal A|>0,\\[6pt]
0,&|\mathcal A|=0
\end{cases}.
\]

Targets retain the window-level normalization without additional per-patch normalization, preserving relative amplitude variation across the window. Naturally missing samples and padding are excluded. Slowly sampled numerical monitor tracks contribute to alignment and distribution regularization, while waveform-specific RFP and reconstruction apply to eligible waveform channels. Waveform-specific terms are zero when neither side is eligible.

\textbf{Joint pretraining objective.} We combine alignment, distribution regularization, RFP and reconstruction using

\[
\mathcal L
=(1-\lambda_s)\mathcal L_{\mathrm{align}}
+\lambda_s\mathcal L_{\mathrm{reg}}
+\beta\mathcal L_{\mathrm{RFP}}
+\gamma\mathcal L_{\mathrm{MAE}}.
\]

Here, \(\mathcal L_{\mathrm{align}}\) is the pairwise alignment loss defined in Section~\ref{pairwise-cross-modal-alignment}, \(\mathcal L_{\mathrm{reg}}\) is the SIGReg distribution regularizer (Appendix~\ref{a.8-exact-alignment-and-sigreg}), \(\mathcal L_{\mathrm{RFP}}\) averages held-out query prediction errors after support-set ridge fitting (Section~\ref{structured-random-feature-prediction}; loss aggregation in Appendix~\ref{a.7-ridge-implementation-and-exact-weighting}), and \(\mathcal L_{\mathrm{MAE}}\) is the masked waveform reconstruction loss defined above. The weight \(\lambda_s\) balances alignment and distribution regularization, while \(\beta\) and \(\gamma\) weight prediction and reconstruction. Our primary configuration uses \(\lambda_s=0.02\) and \(\beta=\gamma=1\). Together, these objectives connect synchronized physiological observations while retaining broadly readable within-signal structure and local waveform detail. Appendix A provides the complete training configuration and update algorithm.

\subsection{Theoretical analysis of random-feature prediction}\label{theoretical-analysis-of-random-feature-prediction}

RFP exposes the encoder to prediction tasks whose defining functions change throughout training. Using the CLS representation and ridge learner of Section 3.4, we analyze what common objective this changing supervision induces, under what conditions it supports downstream prediction, and how continual sampling extends learning beyond a single teacher realization.

\Needspace{12\baselineskip}

\subsubsection{A common objective under changing teachers}\label{a-common-objective-under-changing-teachers}

Let \(\nu\) denote the distribution of input episodes, including the sampled waveform examples, channel identities, student patch masks and support/query splits. Fix the encoder parameters \(\theta\) and one episode from \(\nu\) on an eligible signal side. Let \(S\) and \(Q\) contain \(n=|S|\) support examples and \(q=|Q|\) query examples. Retain the support-standardized student feature matrices \(\widetilde H_S\in\mathbb R^{n\times D}\) and \(\widetilde H_Q\in\mathbb R^{q\times D}\) of Section 3.4, where \(D\) is the representation width. With ridge penalty \(\lambda_r>0\), the operator mapping support targets to query predictions is

\[
A_\theta
=\widetilde H_Q\widetilde H_S^\top
\left(\widetilde H_S\widetilde H_S^\top+n\lambda_r I_n\right)^{-1}
\in\mathbb R^{q\times n},
\]

where \(I_n\) is the \(n\)-dimensional identity matrix.

A scalar generated task \(\xi=(\tau,k)\) consists of a teacher realization \(\tau\) and an output coordinate \(k\), selected uniformly from its \(d_T\) coordinates. Let \(g_{\xi,S}\in\mathbb R^{n+q}\) stack this task's support and query targets after standardization using support statistics. Define the \textbf{task second-moment matrix}, or task kernel, \(K_{T,S}\), and the residual prediction operator \(M_\theta\) as

\[
K_{T,S}=\mathbb E_\xi\left[g_{\xi,S}g_{\xi,S}^\top\right],
\qquad M_\theta=[-A_\theta,\ I_q],
\]

where \(I_q\) is the \(q\)-dimensional identity matrix. The task kernel is positive semidefinite and describes the cross-example target relationships induced by the teacher family. For a sampled task, its mean query error is

\[
\ell_Q(\theta,\xi)=\frac{1}{q}\left\|M_\theta g_{\xi,S}\right\|_2^2.
\]

\Needspace{10\baselineskip}

\textbf{Proposition 1. Expected generated-task risk.} \emph{If the target second moment exists and the teacher is sampled independently of the fixed student and input episode, then}

\[
\boxed{\mathbb E_\xi\,\ell_Q(\theta,\xi)
=\frac{1}{q}\operatorname{tr}\left(M_\theta K_{T,S}M_\theta^\top\right),}
\]

\emph{where \(\operatorname{tr}\) denotes the matrix trace.}

\textbf{Interpretation.} Individual teachers define different prediction functions, while their distribution defines a common expected learning objective. Signal distinctions repeatedly expressed across the teacher family contribute to this objective through \(K_{T,S}\). Refreshing therefore exposes the encoder to changing measurements governed by persistent waveform, spectral and rhythmic priors. \hyperref[fig:5]{Figure~5} illustrates the accessibility of common signal attributes across individual teacher realizations, while Proposition 1 characterizes the corresponding average prediction criterion over the teacher family. The identity also holds after averaging over signal sides, reverse support/query directions and input episodes (Appendix C.1).

\subsubsection{When generated tasks support downstream prediction}\label{when-generated-tasks-support-downstream-prediction}

Consider a downstream regression task distribution evaluated on the same input episodes. Let \(d_S\in\mathbb R^{n+q}\) stack its support and query targets in a fixed evaluation scale, used consistently for the target kernel and regression risk, and define

\[
K_{D,S}=\mathbb E\left[d_Sd_S^\top\right].
\]

Using the same frozen representation, support budget, feature preprocessing and ridge penalty gives the same prediction operator \(M_\theta\). Let \(\mathcal R_T(\theta)\) and \(\mathcal R_D(\theta)\) denote the expected query risks of the generated and downstream tasks, respectively, averaged over \(\nu\) and their respective task distributions.

\Needspace{10\baselineskip}

\textbf{Proposition 2. Downstream-risk bound under task coverage.} \emph{If, for \(\nu\)-almost every episode,}

\[
K_{D,S}\preceq C K_{T,S}
\]

\emph{for a finite constant \(C>0\), then}

\[
\boxed{\mathcal R_D(\theta)\le C\,\mathcal R_T(\theta).}
\]

\emph{Here, \(\preceq\) denotes positive-semidefinite ordering.}

\textbf{Interpretation.} The coverage condition requires the teacher family to assign sufficient task variation to directions needed by the downstream targets. Under this condition, improving generated-task prediction tightens an upper bound on downstream regression error. For the fixed target scale, the constant \(C\) describes the strength of coverage: stronger teacher variation along relevant target directions permits a smaller bound.

This makes the teacher's structure consequential. Waveform convolutions, multiresolution spectral analysis and autocorrelation supply complementary measurement families, broadening the signal attributes exposed during pretraining. Their purpose is to provide coverage across the morphological, spectral and rhythmic information needed by diverse downstream tasks. Appendix C.2 provides the proof, function-space sufficient conditions and an approximate-coverage extension that accounts for residual target directions.

\subsubsection{Learning beyond one teacher realization}\label{learning-beyond-one-teacher-realization}

A simple linear illustration shows how task diversity affects the information a representation must preserve. Suppose the input contains three independent, unit-variance attributes \(x_1,x_2,x_3\), and the representation retains two dimensions. Both representations \((x_1,x_2)\) and \((x_1,x_3)\) contain all information needed for a fixed target \(y=x_1\). This target leaves the choice of the second represented attribute unconstrained.

Now sample targets

\[
y=a_1x_1+a_2x_2,
\]

where \(a_1\) and \(a_2\) are independent, zero-mean, unit-variance random coefficients. Predicting this family rewards preservation of both \(x_1\) and \(x_2\): a representation that omits \(x_2\) loses information required by a recurring set of tasks. Appendix C.3 formalizes this mechanism in a finite-support Gaussian linear model, relating expected query error to task variation omitted by the represented subspace.

In RFP, each vector-valued teacher already supplies multiple measurement functions. Continual refreshing samples new branch filters and nonlinear measurement functions, extending this exposure across realizations of the structured teacher family. \hyperref[fig:5]{Figure~5} also shows that individual realizations differ in how accurately their features expose particular attributes, especially pulse spacing. This variation motivates sampling across teachers, reducing dependence on the information preferences of one initialization. Episode-specific ridge adaptation accommodates the current realization, while query prediction rewards signal structure that remains useful across examples \citep{ref15}. Sections 5.2 and 5.3 examine the resulting attribute accessibility and adaptation to newly sampled teachers empirically.

\section{Physiological transfer and sleep assessment}\label{physiological-transfer-and-sleep-assessment}

LingSante-Pulse supports the distinct information requirements of different physiological signals and reuse across sensor configurations. We evaluate these capabilities through signal-specific tasks, multimodal sleep staging and single-modality sleep assessment. Throughout, the pretrained encoder is frozen and task-specific heads provide the prediction interface.

\subsection{Evaluation scope and protocols}\label{evaluation-scope-baseline-models-and-protocols}

\textbf{Evaluation scope.} We organize downstream evaluation into two groups that address complementary questions. General physiological transfer comprises 38 task configurations: 22 PPG, six ECG, seven EEG and three EMG configurations covering physiological estimation, signal quality, diagnostic and state classification, motor imagery, gestures and continuous trajectories. This group tests whether one encoder preserves the distinct information required across signal domains (Section 4.2).

Sleep assessment comprises eight configurations: seven sleep-staging configurations on the HMC dataset and Sleep-EDF dataset, and one single-lead apnea-screening configuration on the Apnea-ECG dataset. The staging configurations include multimodal linear and overnight assessment on the HMC dataset, EEG-only and EOG-only assessment on both datasets, and chin-EMG-only assessment on the HMC dataset. This group examines reuse across multimodal and single-modality inputs within a common application area (Section 4.3). Inputs across both groups include individual channels, high-density lead arrays and cross-system multimodal recordings. All downstream waveforms and labels were excluded from LingSante-Pulse pretraining.

\hyperref[tab:3]{Table~3} summarizes the datasets, tasks and input configurations used in these evaluations; Appendix F provides the complete task definitions and numerical results.

\Needspace{9\baselineskip}
\phantomsection\label{tab:3}
{\color{black}\textbf{Table 3. Physiological evaluation domains.}} Each panel retains its stated input configuration and prediction target.

{
\begin{longtable}[]{@{}
  >{\raggedright\arraybackslash}p{(\linewidth - 1pt - 4\tabcolsep) * \real{0.2000}}
  >{\raggedright\arraybackslash}p{(\linewidth - 1pt - 4\tabcolsep) * \real{0.5500}}
  >{\raggedright\arraybackslash}p{(\linewidth - 1pt - 4\tabcolsep) * \real{0.2500}}@{}}
\toprule\noalign{}
\rowcolor{PulsePale}
\begin{minipage}[b]{\linewidth}\raggedright
Domain
\end{minipage} & \begin{minipage}[b]{\linewidth}\raggedright
Datasets and targets
\end{minipage} & \begin{minipage}[b]{\linewidth}\raggedright
Input configuration
\end{minipage} \\
\midrule\noalign{}
\endhead
\bottomrule\noalign{}
\endlastfoot
PPG transfer & BIDMC \citep{ref57}, PPG-DaLiA \citep{ref58}, WESAD \citep{ref59}, BUT-PPG \citep{ref60}, PPG-BP \citep{ref61}, WildPPG \citep{ref62} and ECSMP \citep{ref63}; physiological estimation, quality and state classification & Single-channel PPG \\
ECG transfer & PTB-XL \citep{ref65}, MIT-BIH \citep{ref66}, CPSC2018 \citep{ref67} and European ST-T \citep{ref68}; diagnostic, rhythm, beat and ST/T classification & Single-lead or multilead ECG \\
EEG transfer & EEGMMIDB \citep{ref69}, BCIC-IV2a \citep{ref70}, EEGMAT \citep{ref71} and Mumtaz \citep{ref72}; motor imagery, cognitive and clinical state classification & Sparse to high-density EEG arrays \\
EMG transfer & WESAD and NinaPro DB5/DB8 \citep{ref74,ref75}; state, gesture and continuous trajectory decoding & Single-channel or multichannel EMG \\
Sleep assessment & HMC \citep{ref77}, Sleep-EDF \citep{ref76} and Apnea-ECG \citep{ref78}; staging and minute-level apnea screening & Multimodal or single-modality inputs \\
\end{longtable}
}

\textbf{Baseline models.} Signal-specialized foundation models, general time-series and physiological encoders, and multimodal sleep models provide complementary tests of within-domain quality and sleep assessment (\hyperref[tab:4]{Table~4}). Appendix F records the evaluated releases and task-specific coverage.

\Needspace{11\baselineskip}
\phantomsection\label{tab:4}
{\color{black}\textbf{Table 4. Evaluated foundation-model families.}} References identify the model papers or official releases.

{
\begin{longtable}[]{@{}
  >{\raggedright\arraybackslash}p{(\linewidth - 1pt - 2\tabcolsep) * \real{0.2500}}
  >{\raggedright\arraybackslash}p{(\linewidth - 1pt - 2\tabcolsep) * \real{0.7500}}@{}}
\toprule\noalign{}
\rowcolor{PulsePale}
\begin{minipage}[b]{\linewidth}\raggedright
Model family
\end{minipage} & \begin{minipage}[b]{\linewidth}\raggedright
Baseline models
\end{minipage} \\
\midrule\noalign{}
\endhead
\bottomrule\noalign{}
\endlastfoot
General time-series and cross-signal & MOMENT-small \citep{ref30}, NormWear \citep{ref18}, PanLUNA \citep{ref19}, SPOTR-Base \citep{ref31} \\
PPG & AnyPPG \citep{ref05}, PulsePPG \citep{ref33}, PaPaGei-P/S \citep{ref32}, SIGMA-PPG \citep{ref34}, DNA-PPG \citep{ref35} \\
ECG & ECGFounder-1lead/12lead \citep{ref36}, xECG \citep{ref37}, ST-MEM \citep{ref38}, ECG-JEPA \citep{ref39}, ECG-FM \citep{ref08}, ECG-CPC \citep{ref40}, HuBERT-ECG \citep{ref41}, ECG-LFM \citep{ref42} \\
EEG & CBraMod \citep{ref07}, LaBraM \citep{ref06}, REVE \citep{ref43}, CodeBrain EEGSSM \citep{ref44}, ST-EEGFormer \citep{ref45}, EEGPT \citep{ref46}, BIOT \citep{ref47} \\
EMG & NeuroRVQ-EMG \citep{ref48}, TinyMyo \citep{ref49}, PhysioWave-EMG \citep{ref50} \\
Multimodal sleep & SleepFM Clinical \citep{ref03}, OSF \citep{ref02}, SleepLM signal encoder \citep{ref51} \\
\end{longtable}
}

\textbf{Frozen-encoder evaluation.} Throughout downstream evaluation, the pretrained encoder remains frozen. Window-level tasks use the mean of valid observed patch representations as a fixed prediction interface across signal domains. Multichannel tasks concatenate independently encoded 256-dimensional channel representations in a fixed physical order, while continuous trajectory tasks retain the patch sequence. This protocol evaluates whether the same encoder exposes useful information at both window and local temporal scales. Task-specific heads and hyperparameters are fitted using training and validation data; external models retain their documented preprocessing and representation interfaces. Complete CLS benchmarks appear in Appendix F and are compared with patch mean in Section 5.

The HMC dataset and the single-signal benchmarks use their respective participant partitions. All main LingSante-Pulse downstream panels use the 128-Hz encoder. Appendix F provides the corresponding protocols.

\textbf{Aggregation.} Domain-level summaries first average task ranks within each dataset and then weight datasets equally, using fixed sets of baseline models with complete task coverage. We additionally report first-place and top-two counts separately for each domain against all external results reported for the corresponding primary tasks. These summaries describe complementary aspects of transfer: dataset-balanced ranks measure overall competitiveness, while task-level counts show how frequently the representation reaches the leading group. The DB8 dataset contributes one trajectory task, with MAE as its primary ranking metric.

The sleep ranks in \hyperref[fig:1]{Figure~1} average configuration-level ranks over two multimodal HMC dataset settings (linear and overnight) and five single-modality settings (EEG and EOG on the HMC dataset and Sleep-EDF dataset, plus chin EMG on the HMC dataset). The baseline sets with complete task coverage comprise LingSante-Pulse, OSF, SleepLM and SleepFM for multimodal staging, and LingSante-Pulse, OSF and SleepFM for single-modality staging. The overview displays leading models from each fixed comparison set.

\subsection{Preserving diverse information across physiological signals}\label{preserving-diverse-information-across-physiological-signals}

A general-purpose physiological encoder must accommodate information requirements that differ substantially across signals and prediction targets. Pulse-rate estimation depends on recurring temporal structure, cardiac classification requires localized waveform distinctions, EEG tasks combine neural activity across spatial observations, and motor decoding requires continuous temporal information. We therefore evaluate whether one set of pretrained encoder weights can support these requirements across PPG, ECG, EEG and EMG.

\begin{figure}[H]
\phantomsection\label{fig:6}
\centering
\captionsetup{font=footnotesize,skip=4pt}
\includegraphics[width=\linewidth]{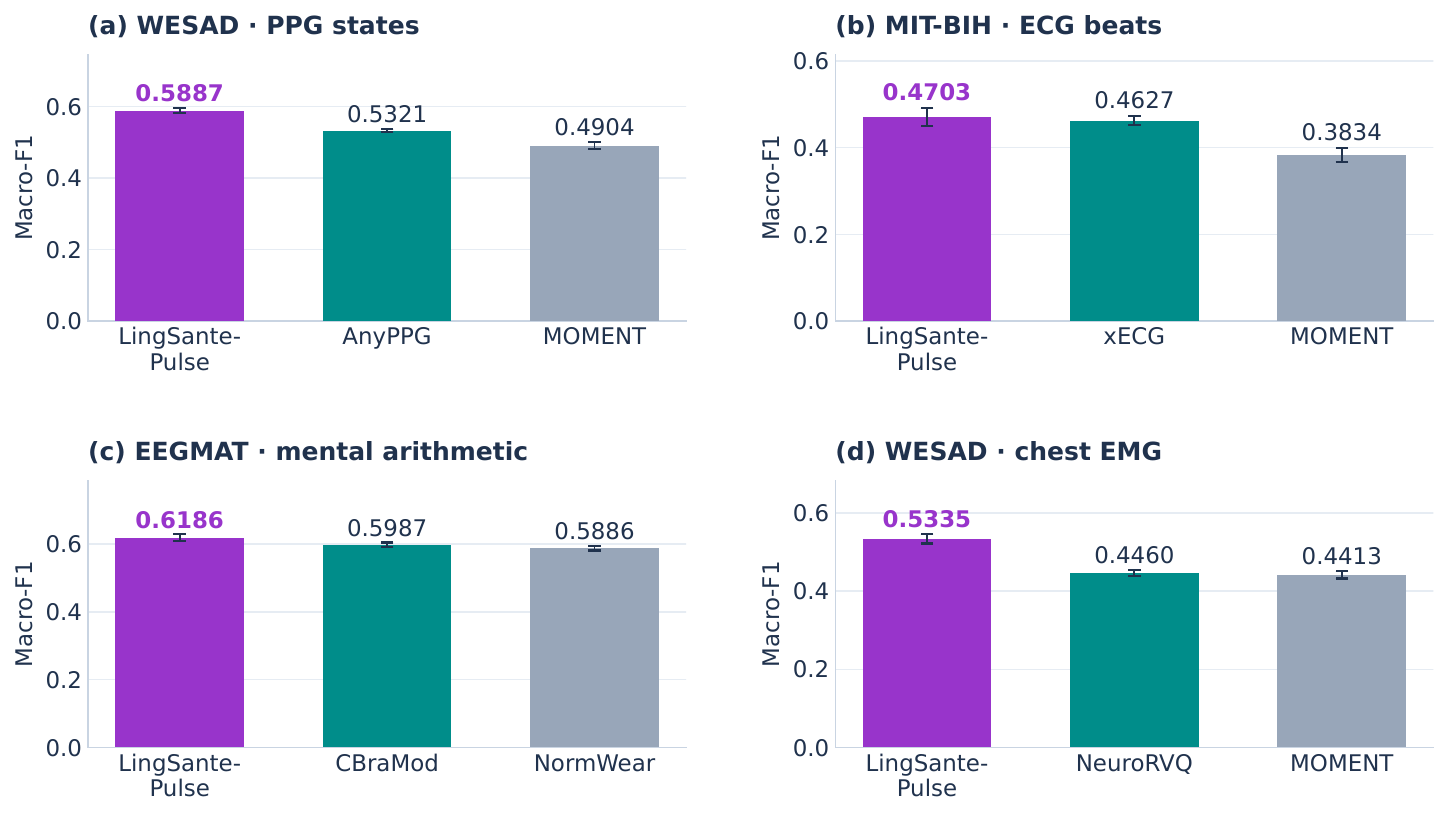}
\caption*{{\color{black}\textbf{Figure 6. Complementary physiological transfer tasks.}} (a) PPG state classification on the WESAD dataset; (b) single-lead beat classification on the MIT-BIH dataset; (c) three-channel mental arithmetic on the EEGMAT dataset; (d) chest-EMG state classification on the WESAD dataset. These examples complement the representative tasks in \hyperref[fig:1]{Figure~1}. All use frozen window-level features and report Macro-F1. Error bars show sample SDs. Appendix F.2 provides the complete results.}
\end{figure}

LingSante-Pulse achieves the best dataset-balanced mean rank in all four evaluated domains: 2.32 across 22 PPG configurations, 3.08 across six ECG configurations, 3.00 across seven EEG configurations and 1.33 across three EMG configurations. Against all reported external comparisons for each primary task, it ranks first on nine PPG, two ECG, two EEG and two EMG configurations. It reaches the top two on 13/22, 2/6, 5/7 and 2/3 configurations, respectively. This breadth is obtained with a shared frozen backbone; the downstream interface changes only through pooling, channel concatenation or retention of the patch sequence. \hyperref[fig:1]{Figure~1} summarizes the domain-level results and representative tasks. \hyperref[fig:6]{Figure~6} supplies complementary examples of physiological state, localized cardiac events and sparse EEG observations.

\textbf{PPG: recovering pulse timing and physiological state under wearable conditions.} For wrist heart-rate estimation on the PPG-DaLiA dataset, LingSante-Pulse achieves an MAE of 5.6843 bpm, reducing error by 25.1\% relative to the strongest external model, AnyPPG. For state classification on the WESAD dataset, it reaches Macro-F1 0.5887 versus 0.5321 for AnyPPG. These tasks require different aspects of the PPG waveform: the former extracts pulse timing from motion-corrupted observations, whereas the latter distinguishes physiological states from window-level signal variation. Strong performance on both shows that the representation retains information useful beyond a single rate-estimation objective. The same encoder therefore supports quantitative physiological estimation and state recognition without task-specific backbone adaptation.

\textbf{ECG: combining localized beat information with multilead observations.} LingSante-Pulse leads the evaluated models on single-lead beat classification in the MIT-BIH dataset and 12-lead rhythm classification in the PTB-XL dataset, reaching Macro-F1 scores of 0.4703 and 0.5890, respectively. These results span two observation scales: short, localized cardiac events and rhythm labels inferred from a multilead recording. For the latter, the prediction head combines independently encoded lead representations, while the backbone continues to process one physical channel at a time. The results demonstrate that shared channel-wise encoding preserves useful cardiac distinctions and supplies features that can be combined effectively across leads.

\textbf{EEG: shared encoding across sparse and high-density observations.} LingSante-Pulse reaches Macro-F1 0.5150 on 64-channel motor imagery in the EEGMMIDB dataset and 0.6186 on three-channel mental arithmetic in the EEGMAT dataset, leading the evaluated models in both configurations. A single channel-conditioned backbone thus supports substantially different channel counts and neural-state targets. Spatially distributed observations enter through the fixed ordering of concatenated channel features, without introducing a task-specific spatial encoder. These findings establish that independently learned channel representations can provide a strong basis for both sparse and high-density EEG prediction under frozen linear evaluation.

\textbf{EMG: transferring from global states to continuous movement trajectories.} LingSante-Pulse leads chest-EMG state classification on the WESAD dataset with Macro-F1 0.5335 and continuous five-degree-of-actuation trajectory decoding on the NinaPro DB8 dataset with MAE 14.3060 and R-squared 0.2674. The DB8 dataset is particularly informative because it changes both prediction granularity and input sampling grid. The encoder was pretrained at 128 Hz, yet receives the published 2,000-Hz, 16-channel arrays from the DB8 dataset directly, without temporal resampling. A fixed Ridge head reads the unpooled patch features to predict trajectories throughout each window. Successful decoding under this 15.6-fold change in sampling rate demonstrates reuse of local temporal features beyond the encoder's primary pretraining grid.

Across these domains, the central result is the reuse of one frozen representation across different physiological sources, channel configurations and prediction granularities. Window pooling supports global labels and scalar estimates, channel concatenation combines distributed observations, and patch-sequence readout supports dense trajectories. Section 5 examines how the pretraining objectives contribute to these representational capabilities.

\subsection{Multimodal and single-modality sleep assessment}\label{multimodal-complementarity-and-single-sensor-sleep-assessment}

Sleep assessment provides a common application in which complementary physiological signals can be evaluated jointly or individually. We use it to examine whether the shared encoder supports both multimodal integration and strong prediction from restricted signal sets. The HMC dataset supplies EEG, EOG, chin EMG and ECG for multimodal staging, together with modality-specific evaluations; the Sleep-EDF dataset provides additional EEG-only and EOG-only staging tests. The Apnea-ECG dataset extends the assessment to sleep-breathing screening from single-lead ECG.

\subsubsection{Multimodal and single-modality sleep staging}\label{multimodal-and-single-sensor-sleep-staging}

\textbf{Four-channel EEG features exceed external multimodal linear results.} On the HMC dataset, LingSante-Pulse reaches Macro-F1 0.7678 using only four EEG channels. This exceeds the multimodal linear-probe results of OSF, SleepLM and SleepFM Clinical, which reach 0.7446, 0.7218 and 0.6018, respectively, on the same participant partition. The comparison highlights the strength of the EEG representation learned within a unified physiological model: restricting its inputs to EEG still yields highly informative features relative to external sleep encoders evaluated with multimodal observations.

\textbf{Multimodal composition and overnight context further improve staging.} Concatenating the frozen EEG, EOG, chin-EMG and ECG representations raises linear-staging Macro-F1 on the HMC dataset to 0.7784, an increase of 1.06 pp over EEG-only assessment. The complementary channel features can therefore improve prediction through a linear head. An overnight BiLSTM head further raises Macro-F1 to 0.7859 using the same frozen patch-mean representations, leading the external sleep models under their corresponding overnight protocols. The encoder supports both epoch-level prediction and sequence modelling, with overnight context supplied by the downstream head.

\begin{figure}[H]
\phantomsection\label{fig:7}
\centering
\captionsetup{font=footnotesize,skip=4pt}
\includegraphics[width=\linewidth,height=.68\textheight,keepaspectratio]{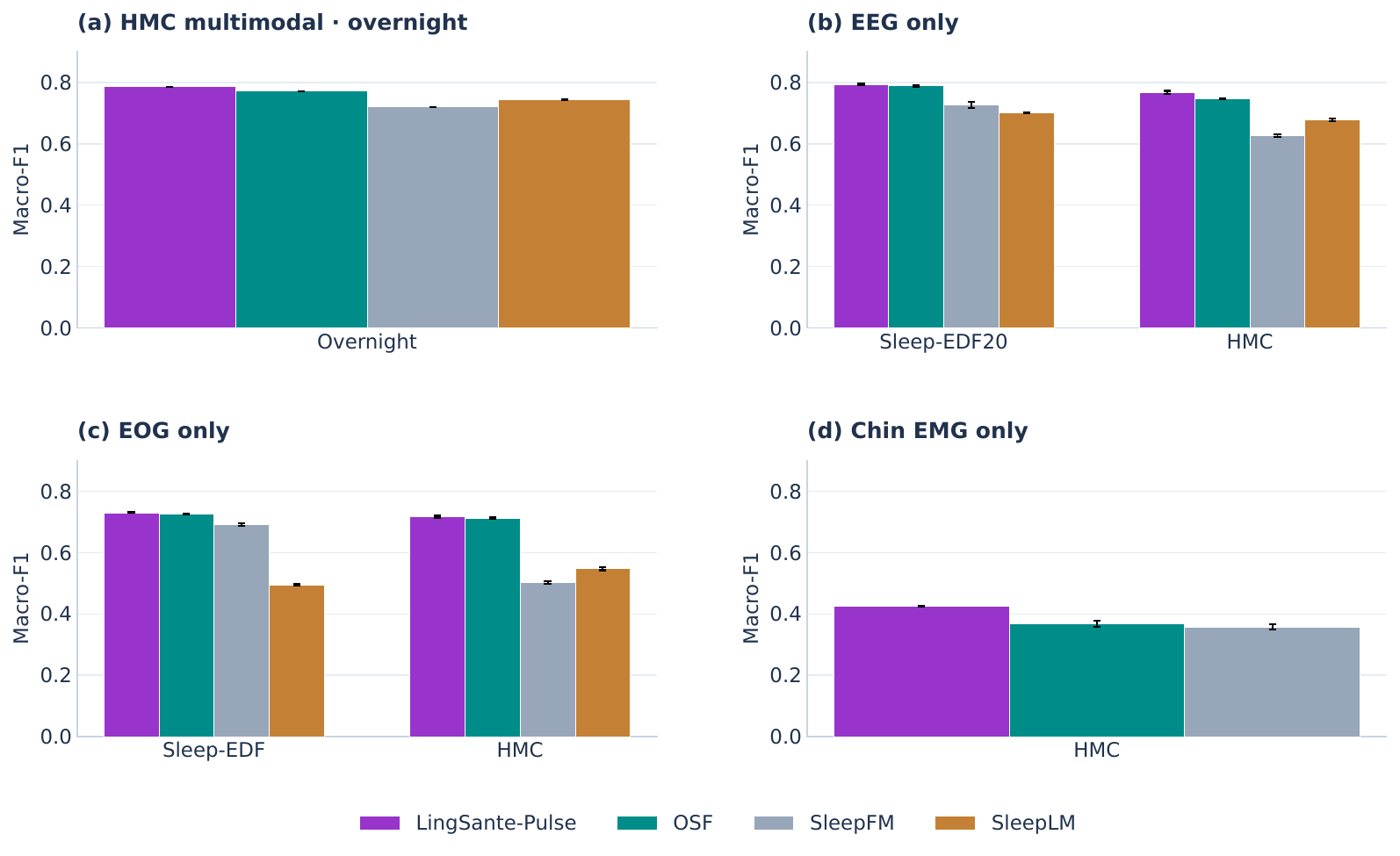}
\caption*{{\color{black}\textbf{Figure 7. Overnight and single-modality sleep staging.}} (a) Multimodal staging on the HMC dataset with overnight BiLSTM heads; its linear-staging results appear in \hyperref[fig:1]{Figure~1}. (b) EEG-only linear staging on the Sleep-EDF20 and HMC datasets. (c) EOG-only linear staging on the Sleep-EDF and HMC datasets. (d) Chin-EMG-only linear staging on the HMC dataset. All panels report Macro-F1. Linear-probe error bars show sample SDs across five head fits; overnight values use fixed fits. Appendix F.3 provides complete numerical results and scoring resolutions.}
\end{figure}

\textbf{Sleep-state information remains accessible without EEG input.} With EOG as the sole modality, LingSante-Pulse reaches Macro-F1 0.7158 on the HMC dataset and 0.7299 on the Sleep-EDF dataset. These results are competitive with the strongest external sleep model in these panels, OSF, at 0.7126 and 0.7252, respectively. On the HMC dataset, they also exceed SleepLM and SleepFM Clinical at 0.5470 and 0.5026. Chin-EMG-only staging reaches Macro-F1 0.4254, ahead of the evaluated external models. Together, these results show that the unified encoder exposes sleep-state information across ocular and muscle-tone signals as well as EEG.

The same pretrained weights consequently support several distinct sleep-assessment configurations: multimodal staging, four-channel EEG-only staging on the HMC dataset, and staging from EOG or chin EMG alone. EEG-only assessment also transfers to the Sleep-EDF20 dataset, where LingSante-Pulse reaches Macro-F1 0.7941 and leads the evaluated models. This combination of strong modality-specific features and straightforward multimodal composition supports applications with different available physiological observations.

\subsubsection{External single-lead apnea screening}\label{external-single-lead-apnea-screening}

The Apnea-ECG dataset tests whether the shared representation extends from sleep-stage classification to sleep-breathing assessment using a different physiological signal. A sequence head operating on frozen 10-second, single-lead ECG patch-mean features achieves minute-level screening AUROC 0.9183. With the same validation-based context selection, ECGFounder-1lead, xECG and ST-MEM achieve 0.8089, 0.7647 and 0.7305, respectively. LingSante-Pulse therefore exceeds these ECG-specialized encoders by 10.94--18.78 pp in AUROC on this benchmark (Appendix F.3.5).

Together with the staging results, this finding establishes reuse across both sensor configurations and sleep-related prediction targets. One frozen physiological encoder supports joint assessment from multiple signals, staging from individual modalities, and apnea screening from single-lead ECG. Section 5.4 examines how synchronized auxiliary relationships can improve representations subsequently evaluated through a single signal.

\section{Understanding LingSante-Pulse pretraining}\label{understanding-ling-pulse-pretraining}

Section 4 evaluates representation reuse across physiological and sleep tasks. Here, we examine how training objectives shape the information available in the encoder. We first use objective ablations to analyze the contribution of each pretraining component to physiological decoding performance. Controlled synthetic signals reveal which attributes remain readable. Online and independent-teacher evaluations examine the learning of signal-derived functions. A paired-alignment study then tests transfer through auxiliary physiological relationships.

\Needspace{8\baselineskip}
\subsection{How the pretraining objectives shape physiological representations}\label{how-the-pretraining-objectives-shape-physiological-representations}

LingSante-Pulse combines cross-modal alignment, structured random-feature prediction and masked waveform reconstruction. We examine how these components contribute to physiological state inference, diagnostic morphology, rhythm recovery and continuous temporal decoding through five ablations. The five variants comprise three objective ablations that individually omit cross-modal alignment, RFP or masked reconstruction; a fixed-teacher variant that reuses the same teacher parameters throughout pretraining; and a support-fitting variant that optimizes prediction error on the support set instead of held-out queries. The fixed-teacher variant retains episodic ridge adaptation; the support-set variant retains the ridge learner and changes the predictions used by the encoder's outer objective.

All ablated encoders retain the backbone architecture and pretraining data mixture. Downstream evaluation freezes each encoder and follows the corresponding task protocol. \hyperref[tab:5]{Table~5a} compares representative tasks using patch mean for window-level prediction and the unpooled patch sequence for trajectories from the DB8 dataset. EOG staging on the Sleep-EDF dataset tests physiological state information; subclass classification on the PTB-XL dataset and beat classification on the MIT-BIH dataset test diagnostic and localized waveform distinctions; heart-rate estimation on the PPG-DaLiA dataset tests pulse-rate recovery under motion; and the DB8 dataset tests continuous motor decoding. \hyperref[tab:5]{Table~5b} adds the corresponding motor-imagery comparison on the EEGMMIDB dataset.

\Needspace{26\baselineskip}

\phantomsection\label{tab:5}
{\color{black}\textbf{Table 5. Representative tasks illustrating the contributions of the pretraining objectives.}} (a) Physiological state inference, diagnostic morphology, beat classification, heart-rate recovery under motion and continuous trajectory decoding. Window-level tasks use patch mean; the DB8 dataset uses the patch sequence. (b) Matched patch-mean and CLS readouts on four-class motor imagery from the 64-channel EEGMMIDB dataset. Classification reports Macro-F1; the PPG-DaLiA dataset reports MAE in bpm, and the DB8 dataset reports MAE in benchmark glove-coordinate units. Means and sample SDs summarize repeated head fits; deterministic probes are shown as single values.

\textbf{(a) Primary downstream interfaces}

\begingroup
\small
\setlength{\tabcolsep}{3.5pt}\renewcommand{\arraystretch}{1.18}
\begin{tabular}{@{}p{3.7cm}*{5}{>{\raggedleft\arraybackslash}p{\dimexpr(\linewidth-3.7cm-10\tabcolsep)/5\relax}}@{}}
\toprule
\rowcolor{PulsePale}Configuration & \shortstack[r]{Sleep-EDF\\EOG F1 $\uparrow$} & \shortstack[r]{PTB-XL\\Subclass F1 $\uparrow$} & \shortstack[r]{MIT-BIH\\Beat F1 $\uparrow$} & \shortstack[r]{PPG-DaLiA\\HR MAE $\downarrow$} & \shortstack[r]{DB8\\MAE $\downarrow$}\\\midrule
Full & 0.7299{\scriptsize\,$\pm$\,0.0014} & 0.4113 & 0.4703{\scriptsize\,$\pm$\,0.0211} & 5.6843{\scriptsize\,$\pm$\,0.1351} & 14.3060\\
Without alignment & 0.6381{\scriptsize\,$\pm$\,0.0148} & 0.4153 & 0.4435{\scriptsize\,$\pm$\,0.0080} & 7.2640{\scriptsize\,$\pm$\,0.1846} & 14.6084\\
Without RFP & 0.6599{\scriptsize\,$\pm$\,0.0097} & 0.4007 & 0.4686{\scriptsize\,$\pm$\,0.0188} & 5.5792{\scriptsize\,$\pm$\,0.1386} & 14.7674\\
Without reconstruction & 0.6488{\scriptsize\,$\pm$\,0.0204} & 0.3394 & 0.4708{\scriptsize\,$\pm$\,0.0193} & 6.1541{\scriptsize\,$\pm$\,0.1440} & 14.4794\\
Fixed teacher & 0.7048{\scriptsize\,$\pm$\,0.0016} & 0.4078 & 0.4578{\scriptsize\,$\pm$\,0.0152} & 5.9552{\scriptsize\,$\pm$\,0.1531} & 14.5182\\
Support-set outer loss & 0.6780{\scriptsize\,$\pm$\,0.0087} & 0.4129 & 0.4821{\scriptsize\,$\pm$\,0.0121} & 5.7628{\scriptsize\,$\pm$\,0.1292} & 14.1968\\
\bottomrule
\end{tabular}
\endgroup

\medskip

\Needspace{9\baselineskip}
\textbf{(b) Motor-imagery results on the EEGMMIDB dataset}

\begingroup
\small
\setlength{\tabcolsep}{6pt}\renewcommand{\arraystretch}{1.18}
\begin{tabular}{@{}p{.50\linewidth}*{2}{>{\raggedleft\arraybackslash}p{\dimexpr.25\linewidth-2\tabcolsep\relax}}@{}}
\toprule
\rowcolor{PulsePale}Configuration & Patch mean F1 $\uparrow$ & CLS F1 $\uparrow$\\\midrule
Full & 0.5150{\scriptsize\,$\pm$\,0.0015} & 0.5301{\scriptsize\,$\pm$\,0.0035}\\
Without alignment & 0.5162{\scriptsize\,$\pm$\,0.0040} & 0.5426{\scriptsize\,$\pm$\,0.0015}\\
Without RFP & 0.4889{\scriptsize\,$\pm$\,0.0017} & 0.3804{\scriptsize\,$\pm$\,0.0013}\\
Without reconstruction & 0.4999{\scriptsize\,$\pm$\,0.0028} & 0.5439{\scriptsize\,$\pm$\,0.0017}\\
Fixed teacher & 0.5153{\scriptsize\,$\pm$\,0.0015} & 0.5547{\scriptsize\,$\pm$\,0.0014}\\
Support-set outer loss & 0.5013{\scriptsize\,$\pm$\,0.0027} & 0.5198{\scriptsize\,$\pm$\,0.0022}\\
\bottomrule
\end{tabular}
\endgroup

\medskip

\Needspace{7\baselineskip}
\textbf{Without alignment: weaker state inference and pulse-rate recovery.} Removing alignment reduces EOG-only sleep-staging Macro-F1 from 0.7299 to 0.6381 and increases heart-rate MAE on the PPG-DaLiA dataset from 5.6843 to 7.2640 bpm. These tasks require recovering physiological state or pulse timing from signals whose observations can be ambiguous or corrupted by motion. The losses are consistent with synchronized partner signals providing complementary training information that within-signal prediction and reconstruction alone do not supply. Section 5.4 examines the contribution of auxiliary correspondence under matched signal exposure.

\Needspace{7\baselineskip}
\textbf{Without RFP: poorer state recognition and continuous motor decoding.} Removing RFP reduces EOG sleep-staging Macro-F1 to 0.6599 and motor-imagery Macro-F1 on the EEGMMIDB dataset from 0.5150 to 0.4889, while trajectory MAE on the DB8 dataset rises from 14.3060 to 14.7674. Alignment emphasizes relationships shared with a partner signal, and reconstruction trains recovery of hidden samples. RFP additionally requires signal-derived functions to remain predictable across examples. Its removal therefore weakens an explicit source of supervision for broadly accessible signal attributes, consistent with the deficits in state inference and trajectory decoding.

\Needspace{7\baselineskip}
\textbf{Without reconstruction: loss of diagnostic morphology and temporal detail.} Removing reconstruction reduces diagnostic-subclass Macro-F1 on the PTB-XL dataset from 0.4113 to 0.3394 and EOG staging Macro-F1 to 0.6488. heart-rate MAE on the PPG-DaLiA dataset rises to 6.1541 bpm, and trajectory MAE on the DB8 dataset rises to 14.4794. Reconstruction directly supervises hidden waveform samples, including local shapes and their temporal arrangement. The resulting deficits are consistent with alignment and teacher-feature prediction providing less direct pressure to preserve the fine waveform detail needed by these tasks. Section 5.2 separates these information requirements using controlled signals.

\Needspace{7\baselineskip}
\textbf{Fixed teacher: weaker transfer across beat structure, pulse timing and trajectories.} Keeping the teacher fixed reduces beat-classification Macro-F1 on the MIT-BIH dataset from 0.4703 to 0.4578, raises MAE on the PPG-DaLiA dataset from 5.6843 to 5.9552 bpm, and raises MAE on the DB8 dataset from 14.3060 to 14.5182. EOG staging also decreases to 0.7048. Because episodic ridge adaptation is retained, the change restricts the measurement functions encountered during training. A persistent teacher can favor the attributes exposed by its particular initialization; refreshing broadens this functional supervision, consistent with better transfer across these distinct waveform requirements. Section 5.3 examines learning across independently sampled teachers.

\Needspace{7\baselineskip}
\textbf{Support-set outer loss: weaker generalization of state information.} Replacing query prediction error with support-set fitting error reduces EOG sleep-staging Macro-F1 from 0.7299 to 0.6780 and motor-imagery Macro-F1 on the EEGMMIDB dataset from 0.5150 to 0.5013. heart-rate MAE on the PPG-DaLiA dataset also rises to 5.7628 bpm. The ridge learner is unchanged, but fitting its support targets no longer directly rewards predictions on other examples. This provides a plausible explanation for the poorer state-sensitive transfer: information useful for fitting the current support set need not be equally useful on unseen waveforms.

\Needspace{10\baselineskip}
The variants thus lose performance under different information requirements. Full provides the most robust balance across the five representative primary tasks in \hyperref[tab:5]{Table~5a}: its largest relative shortfall from the best configuration is 2.45\%, compared with at least 6.74\% for every ablation. Here, relative shortfall is the difference from the best score divided by that score, with the direction reversed for MAE. Combining synchronized supervision, changing measurement functions, held-out prediction and waveform reconstruction avoids the larger task-specific deficits that appear when one component is removed or restricted. Section 5.2 examines these complementary information preferences with controlled temporal, spectral and rhythmic attributes.

\Needspace{8\baselineskip}
\subsection{Signal information preserved by different pretraining objectives}\label{signal-information-preserved-by-different-pretraining-objectives}

The within-signal objective determines which distinctions remain accessible in a frozen representation. We study this using synthetic waveforms with controlled amplitude envelopes, transient events, spectral mixtures and pulse timing. All training configurations share the waveform stream, encoder, optimizer and update budget. After pretraining, frozen ridge probes recover six predefined attributes. We report normalized mean squared error (NMSE), using target scales fitted on the probe-training split. Patch mean is the primary readout; CLS provides a parallel view of the global summary (\hyperref[fig:8]{Figure~8}; Appendix~\ref{b.2-matched-data-synthetic-study}).

\Needspace{8\baselineskip}
\subsubsection{View construction trades temporal context against rhythmic detail}\label{view-construction-trades-temporal-context-against-rhythmic-detail}

Temporal crops can preserve repeated rhythms while changing the meaning of a local event. A transient appearing early in a crop may occur anywhere in the original window. Training the crop and full window to agree therefore encourages the encoder to overlook distinctions that matter for global timing.

\begin{figure}[H]
\phantomsection\label{fig:8}
\centering
\includegraphics[width=\linewidth,height=.70\textheight,keepaspectratio]{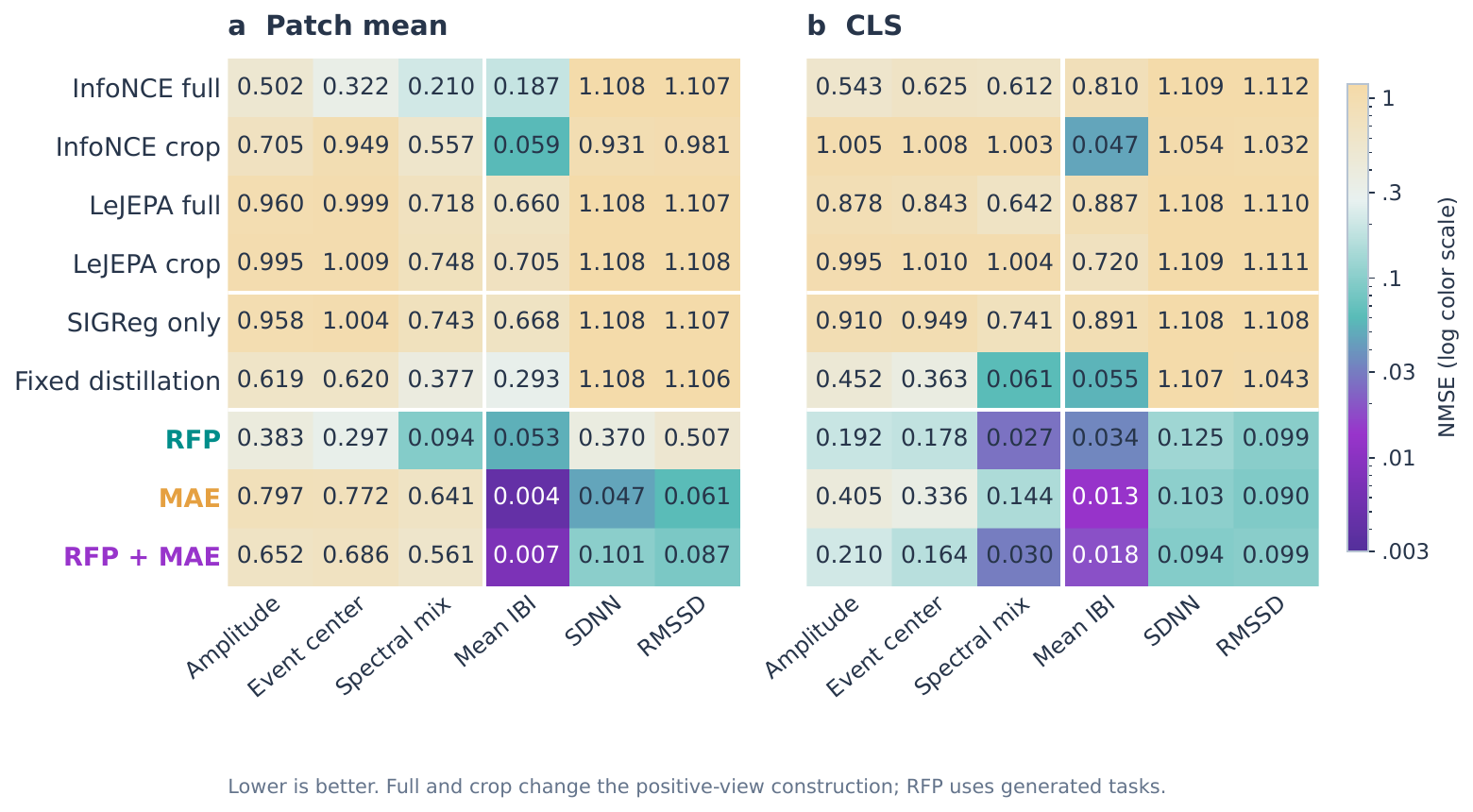}
\caption*{{\color{black}\textbf{Figure 8. View construction changes which signal attributes remain readable.}} Test NMSE for nine matched training configurations, evaluated with patch mean in (a) and CLS in (b). Each panel holds the readout fixed across all methods and tasks. Values are means over three training repetitions at observation noise 0.3; lower is better and color is logarithmic. The first four rows compare full/full and full/crop objectives; subsequent rows compare regularization, fixed-target distillation, RFP, reconstruction and their combination. \hyperref[tab:synthetic-task-errors]{Table~D1} gives means and SDs; \hyperref[fig:synthetic-readout-ranks]{Figure~D1} provides the corresponding ranks.}
\end{figure}

\hyperref[fig:8]{Figure~8} shows how this choice changes attribute recovery. With patch mean, InfoNCE crop recovers the mean inter-beat interval with NMSE 0.059, compared with 0.187 for InfoNCE full. The order reverses for event center: full-window noisy views give error 0.322, whereas cropping raises it to 0.949. Full views preserve global event position more effectively; cropped views favor the repeated pulse structure in this comparison. Neither view construction consistently preserves all six attributes across the InfoNCE and LeJEPA configurations.

RFP trains the predictability of signal-derived functions across examples. Its patch-mean errors are 0.383 for amplitude profile, 0.297 for event center, 0.094 for spectral mixture and 0.053 for mean inter-beat interval. It combines the low mean-interval error of cropped InfoNCE with readable global temporal and spectral attributes. This balance motivates a within-signal objective that exposes several types of signal structure without prescribing crop invariance.

\begin{figure}[H]
\phantomsection\label{fig:9}
\centering
\includegraphics[width=\linewidth,height=.70\textheight,keepaspectratio]{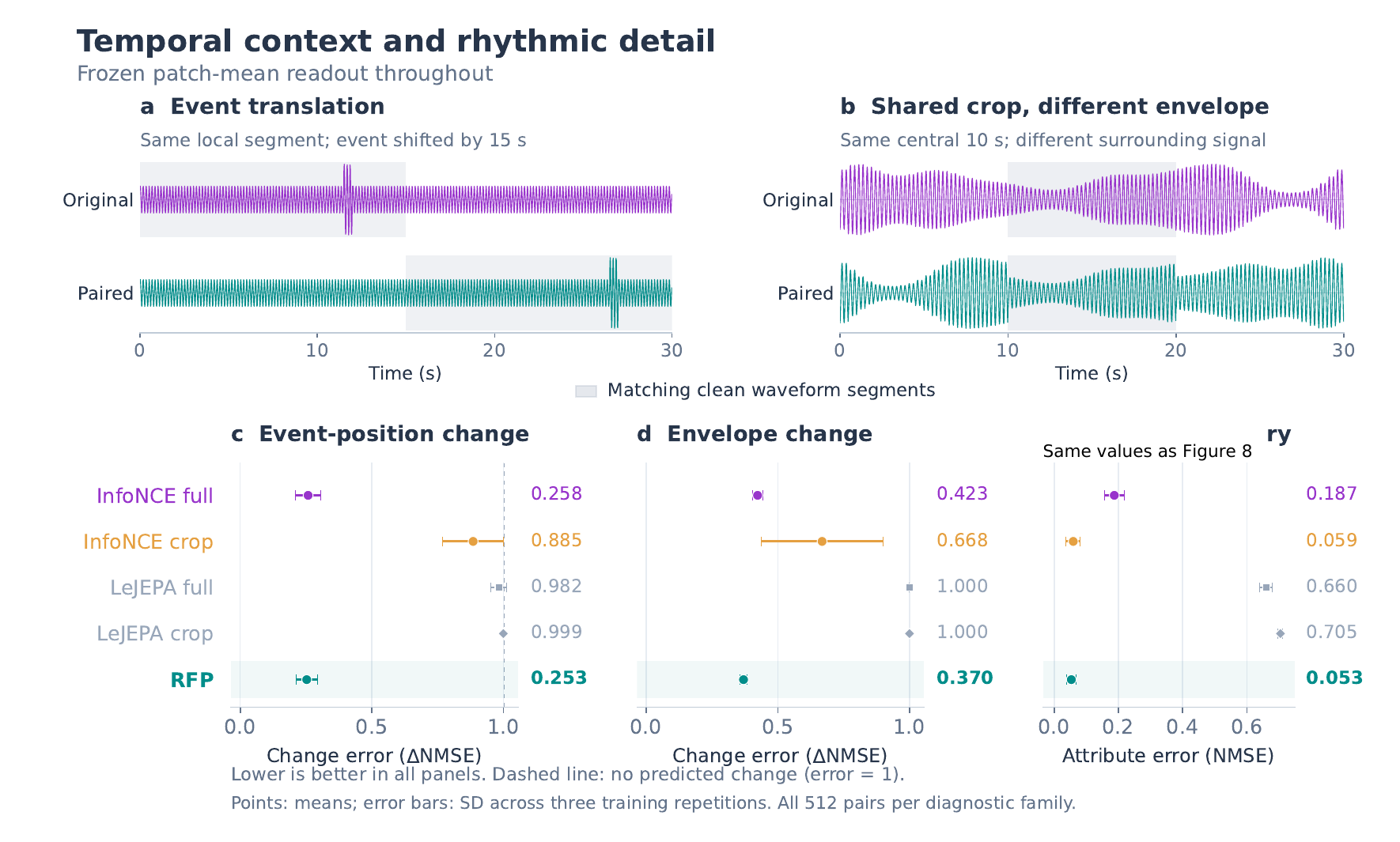}
\caption*{{\color{black}\textbf{Figure 9. Temporal context and rhythmic detail under different pretraining objectives.}} All panels use frozen patch-mean representations. (a,b) Clean event-translation and shared-crop waveform pairs; matching segments are shaded. (c,d) Original frozen ridge probes recover between-member changes. Change error is normalized by the squared true change; predicting no change gives 1 (dashed line). (e) Mean inter-beat interval (IBI) NMSE from \hyperref[fig:8]{Figure~8} connects temporal readability with rhythmic detail. Lower is better throughout. Points and bars show means and sample SDs across three training repetitions at observation noise 0.3, using all 512 pairs per diagnostic family. Appendices~\ref{b.2-matched-data-synthetic-study} and~\ref{b.7-counterfactual-representation-diagnostics} provide protocols and CLS diagnostics.}
\end{figure}

Counterfactual pairs clarify the temporal side of this trade-off (\hyperref[fig:9]{Figure~9a,b}). In the event-translation family, moving an event by 15 seconds changes its global position. Nevertheless, the first 15 seconds of one waveform exactly match the last 15 seconds of its paired waveform. Without the original crop offset, these identical local observations cannot distinguish the two global positions. A second diagnostic holds the central 10 seconds fixed while changing the surrounding amplitude envelope. It tests whether information outside shared local content remains readable from the complete window.

We evaluate all 512 test pairs in each diagnostic family using the original frozen patch-mean probes (\hyperref[fig:9]{Figure~9c,d}). Change error compares the predicted attribute difference with the true difference; an unchanged prediction gives error 1. InfoNCE full retains substantial temporal information: its event-position change error is 0.258, close to RFP at 0.253, and its envelope-change error is 0.423, compared with 0.370 for RFP. InfoNCE crop gives larger errors of 0.885 and 0.668, respectively. LeJEPA crop approaches 1 on both diagnostics. The counterfactuals thus reveal the temporal cost of crop-based agreement, while also showing that full-window InfoNCE preserves these distinctions well.

\hyperref[fig:9]{Figure~9e} places mean-interval recovery from \hyperref[fig:8]{Figure~8} alongside these temporal diagnostics. The two figures together expose the view-design trade-off: full-window InfoNCE preserves global temporal differences but recovers the repeated pulse interval less accurately; cropping improves interval recovery while weakening global temporal readability. RFP achieves low errors on both types of structure in this setting. Its benefit is the ability to retain these complementary attributes within the same representation.

\Needspace{8\baselineskip}
\subsubsection{From target matching to task-readable representations}\label{from-target-matching-to-task-readable-representations}

A structured random target provides measurements of the signal, while the learning objective determines how those measurements shape the representation. The fixed-target distillation configuration uses a persistent learned prediction head. RFP instead fits an episode-specific linear readout and trains its predictions on held-out query examples.

With patch mean, direct distillation gives amplitude-profile NMSE 0.619, compared with 0.383 for RFP. Its two interval-variability errors are 1.108 and 1.106, compared with 0.370 and 0.507. These targets are the standard deviation of inter-beat intervals (SDNN) and the root mean square of successive interval differences (RMSSD). The comparison highlights the role of the supervision recipe: fitting a target through a persistent head and exposing signal functions to an adapted linear learner produce different attribute preferences. The fixed-teacher RFP configurations, which retain episodic adaptation, are reported separately in \hyperref[tab:synthetic-task-errors]{Table~D1}.

\Needspace{8\baselineskip}
\subsubsection{Complementary prediction and reconstruction}\label{complementary-prediction-and-reconstruction}

Prediction and reconstruction emphasize different information. RFP gives the lowest patch-mean errors among the primary configurations for amplitude profile, event center and spectral mixture. MAE gives the lowest errors for mean inter-beat interval, SDNN and RMSSD. This pattern connects cross-example function prediction to broader signal summaries, and waveform reconstruction to precise pulse timing.

Joint training combines these preferences. Adding RFP to MAE reduces amplitude-profile, event-center and spectral-mixture errors by 18.2\%, 11.1\% and 12.5\%, respectively. Adding MAE to RFP reduces mean-interval error from 0.0533 to 0.0075. Its interval-variability errors also move toward those of reconstruction alone (\hyperref[fig:synthetic-complementarity]{Figure~D2}). The combination therefore improves the readability of global attributes over pure reconstruction while recovering much of the pulse-timing precision absent from RFP alone.

These controlled results explain why function prediction and waveform reconstruction can complement each other within the unified encoder. They also connect the view-design motivation to observable representation behavior: the objective shapes which temporal, spectral and rhythmic distinctions remain available to a frozen downstream learner.

\Needspace{8\baselineskip}
\subsection{Learning transferable signal structure from refreshed random teachers}\label{learning-transferable-signal-structure-from-refreshed-random-teachers}

RFP changes its target functions at every training update while continually refining the same encoder. Its effectiveness therefore depends on learning signal structure that remains useful across different teacher realizations. We examine this process from two perspectives: prediction of freshly sampled teachers during pretraining, and adaptation to independently sampled teachers after freezing the encoder.

\Needspace{8\baselineskip}
\subsubsection{Online prediction improves during pretraining}\label{prediction-improves-despite-continual-teacher-refreshing}

We track RFP prediction throughout the model's training trajectory. At each update, the newly sampled teacher produces the complete 576-dimensional target, including the channel-conditioned residual transformation and random mixing projection. The student observes masked waveforms, and an episode-specific ridge learner maps its CLS representations to the current teacher outputs. Each direction fits the learner on 512 support examples and evaluates its predictions on 512 query examples.

To compare prediction quality across changing targets, we define normalized query skill as

\[
\operatorname{skill}=1-\frac{L_Q}{B_Q},
\]

where \(L_Q\) is the query mean squared prediction error and \(B_Q\) is the error obtained by predicting every query target with the support-set target mean. Feature and target standardization use support examples only. A skill of zero corresponds to the error of this constant predictor, whereas a skill of one indicates perfect query prediction.

\begin{figure}[H]
\phantomsection\label{fig:10}
\centering
\includegraphics[width=\linewidth]{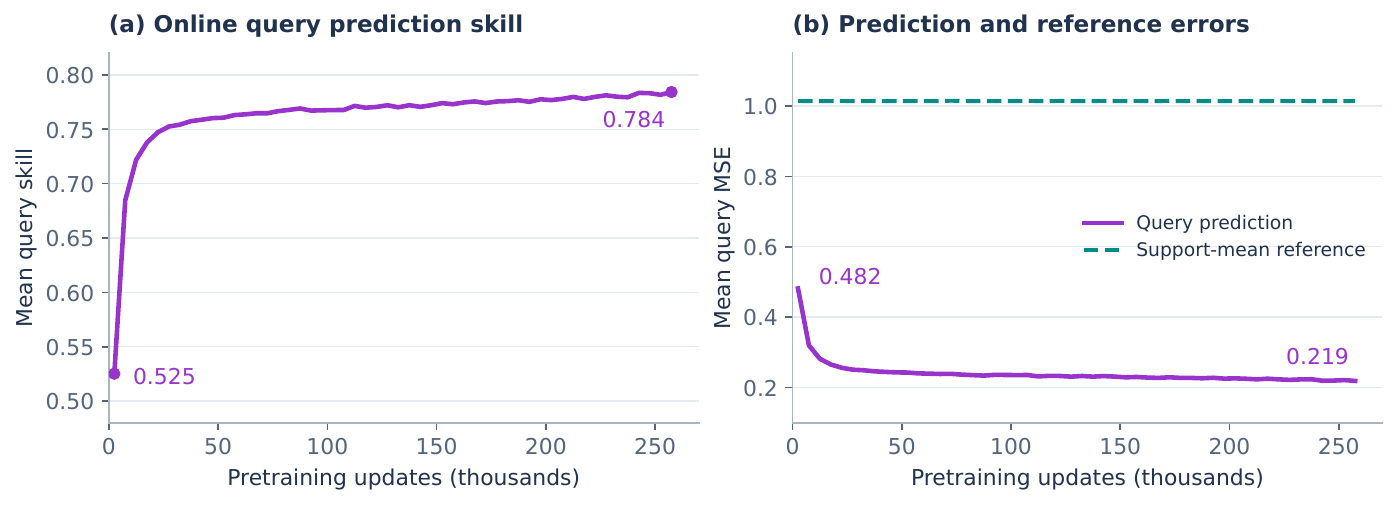}
\caption*{{\color{black}\textbf{Figure 10. Online prediction improves under refreshed random supervision.}} (a) Mean query skill during pretraining. (b) Mean query MSE and the support-mean reference error. Each point aggregates 5,000 consecutive updates, weighted by the number of eligible updates contributing waveform targets. The teacher is freshly sampled at each update; the episode-specific ridge learner is fitted on support examples and evaluated on held-out queries.}
\end{figure}

Prediction improves as training proceeds under continually refreshed teachers (\hyperref[fig:10]{Figure~10}). Mean query skill rises from 0.525 in the first 5,000 updates to 0.784 in the final 5,000 updates, while query MSE decreases from 0.482 to 0.219 against a stable support-mean reference error of approximately 1.014.

The changing targets therefore support cumulative learning. Although each update introduces a different measurement function, the encoder becomes increasingly effective at exposing information from which a support-fitted linear learner can predict that function on other recordings. The shared waveform, spectral and rhythmic structure across teacher realizations provides continuity beneath their changing output coordinates.

\Needspace{8\baselineskip}
\subsubsection{Frozen representations support previously unseen teacher realizations}\label{frozen-representations-support-previously-unseen-teacher-realizations}

The online trajectory measures learning under the changing training objective. We next evaluate whether the resulting representation supports teacher realizations independently sampled after training.

We evaluate frozen encoders from early, intermediate and late training using 128 new random teachers on a common panel of PPG windows from the BIDMC dataset. Each teacher uses the same structured construction as during pretraining. All encoder stages receive identical waveform windows, patch masks, teacher targets and support/query assignments. Support and query contain different patients. Only the episode-specific ridge learner is fitted; the encoder remains frozen throughout evaluation (Appendix~\ref{b.5-independent-generated-task-protocol}).

\begin{figure}[H]
\phantomsection\label{fig:11}
\centering
\includegraphics[width=\linewidth,height=.70\textheight,keepaspectratio]{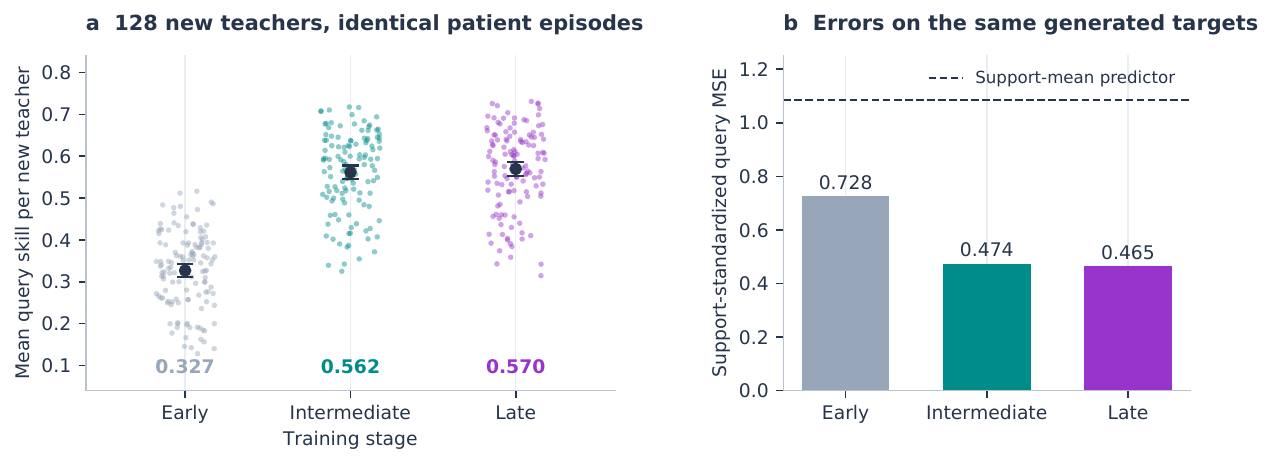}
\caption*{{\color{black}\textbf{Figure 11. Frozen representations adapt to new random teachers.}} (a) Each colored point is one teacher's mean skill over four fixed patient partitions and both directions. Black points and intervals show the teacher mean and normal-approximation 95\% interval for that mean, conditional on the fixed patient panel. (b) Mean query MSE and the common support-mean reference error. All 128 teachers, patient episodes and masks are shared across stages. Teacher variation does not estimate population-level patient uncertainty; Appendix~\ref{b.5-independent-generated-task-protocol} records the protocol.}
\end{figure}

Mean prediction skill increases from 0.327 at the early stage to 0.562 at the intermediate stage and 0.570 at the final stage (\hyperref[fig:11]{Figure~11}). Query MSE correspondingly decreases from 0.728 to 0.474 and 0.465. Most improvement occurs before the intermediate stage, followed by a smaller subsequent gain.

Because the evaluation functions and input episodes are shared across stages, this improvement reflects greater accessibility of signal information in the frozen representation. The encoder has learned features that support adaptation to new realizations of the teacher family, beyond the particular functions encountered during its updates.

Together, the online and frozen-encoder evaluations explain how refreshed random supervision can produce coherent learning. The ridge learner accommodates the current measurement function, while the encoder accumulates signal structure useful across functions and examples. This behavior supports the task-family interpretation in Section 3.6 and complements the downstream refreshing ablation in Section 5.1.

\Needspace{8\baselineskip}
\subsection{Transitive transfer through pairwise alignment}\label{transitive-transfer-through-pairwise-alignment}

Pairwise alignment connects synchronized observations available within individual recording collections. We demonstrate that relational supervision from an auxiliary signal pair can improve the representation of a third signal connected through a shared modality. We investigate this through an EEG-ECG-respiration study: auxiliary EEG-ECG observations are introduced during pretraining, while downstream evaluation uses respiration alone. EEG and respiration are never directly paired during training.

\Needspace{8\baselineskip}
\subsubsection{Separating auxiliary observations from auxiliary correspondence}\label{separating-auxiliary-observations-from-auxiliary-correspondence}

We train four unified encoders using synchronized EEG, ECG and thoracic-effort recordings from identical initial parameters and with the same total budget of 50,000 pair updates. Arm A receives only ECG-respiration pairs. Arms B, C and D allocate half their updates to ECG-respiration and half to auxiliary EEG-ECG observations.

The three auxiliary-data arms differ only in how the EEG-ECG alignment term is applied. Arm B aligns correctly synchronized pairs. Arm C disables their alignment while retaining the same waveform inputs and other objectives. Arm D aligns mismatched EEG-ECG observations by shuffling correspondence. Their ECG-respiration pairs remain correctly aligned. Across B, C and D, the complete input stream, waveform masks, teacher draws and non-alignment objectives are matched.

These comparisons distinguish two contributions. B versus C tests the effect of auxiliary alignment at fixed signal exposure. B versus D tests whether the correct physiological correspondence matters. Arm A provides a reference that spends the entire training budget on the relationship directly involving the evaluation signal.

After pretraining, each frozen encoder processes the same thoracic-effort recordings and is evaluated with an identical linear sleep-staging probe. No EEG or ECG is supplied at inference. Both CLS and patch mean are evaluated on the same held-out patient panel (\hyperref[tab:6]{Table~6}; Appendix E).

\Needspace{12\baselineskip}

\phantomsection\label{tab:6}
{\color{black}\textbf{Table 6. Auxiliary correspondence and respiration-only sleep staging.}} Test Macro-F1 (\%). The auxiliary relationship is EEG--ECG; every test input is thoracic effort. B/C/D share their waveform inputs, masks, teacher draws and non-alignment objectives.

{
\begin{longtable}[]{@{}>{\raggedright\arraybackslash}p{7mm}>{\centering\arraybackslash}p{24mm}>{\centering\arraybackslash}p{23mm}>{\raggedright\arraybackslash}p{\dimexpr\linewidth-84mm-10\tabcolsep\relax}>{\raggedleft\arraybackslash}p{12mm}>{\raggedleft\arraybackslash}p{18mm}@{}}
\toprule\noalign{}
\rowcolor{PulsePale}
Arm & \shortstack{ECG--respiration\\updates} & \shortstack{EEG--ECG\\updates} & EEG--ECG treatment & CLS & Patch mean \\
\midrule\noalign{}
\endhead
\bottomrule\noalign{}
\endlastfoot
A & 50,000 & 0 & ECG--respiration training only & 39.92 & 43.79 \\
B & 25,000 & 25,000 & Align correctly synchronized EEG--ECG pairs & \textbf{43.86} & \textbf{44.56} \\
C & 25,000 & 25,000 & Same EEG and ECG inputs; their alignment loss disabled & 39.65 & 44.00 \\
D & 25,000 & 25,000 & Align EEG with ECG from other patients & 39.63 & 44.03 \\
\end{longtable}
}

\Needspace{8\baselineskip}
\subsubsection{Correct auxiliary relationships improve respiration-only representations}\label{correct-auxiliary-relationships-improve-respiration-only-representations}

\begin{figure}[H]
\phantomsection\label{fig:12}
\centering
\includegraphics[width=\linewidth,height=.70\textheight,keepaspectratio]{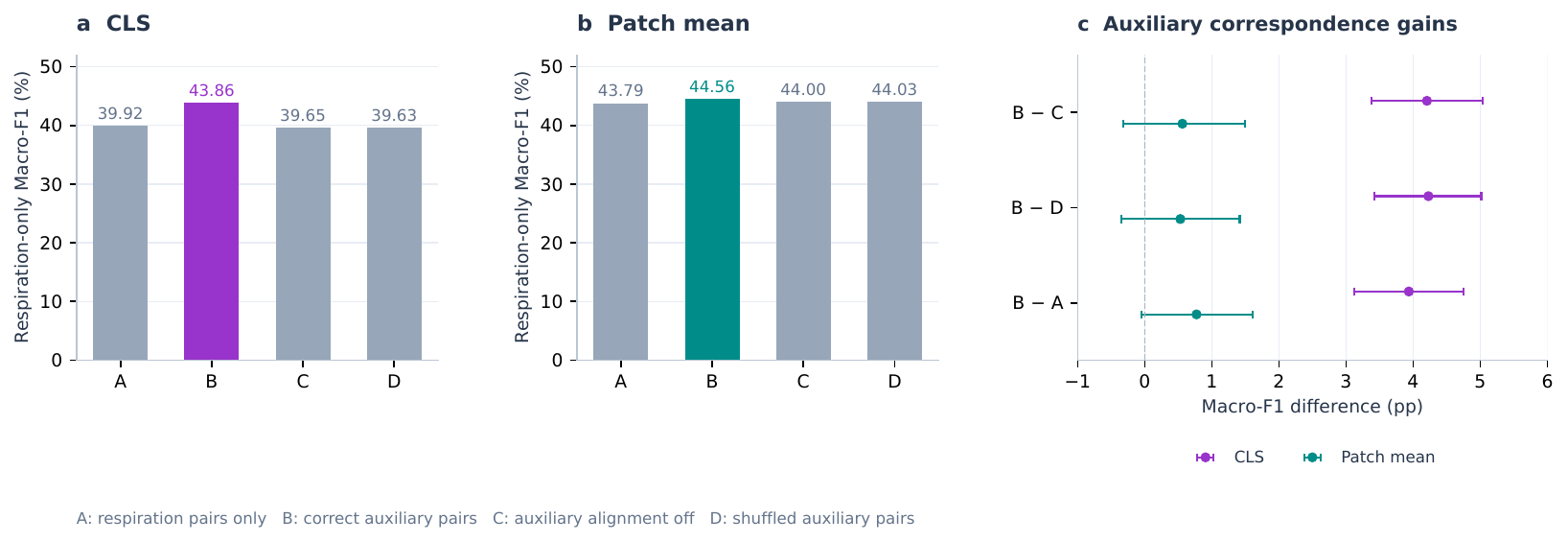}
\caption*{{\color{black}\textbf{Figure 12. Auxiliary correspondence across readouts.}} (a,b) Respiration-only test Macro-F1 for the four arms, separately for CLS and patch mean. (c) B−C, B−D and B−A differences with 95\% paired patient-bootstrap intervals from 2,000 resamples of 642 patients. The same patient draws are used across arms and readouts.}
\end{figure}

Correct auxiliary correspondence produces the highest point estimate under both readouts, with the clearest effect expressed through CLS (\hyperref[fig:12]{Figure~12}). Relative to the matched-input controls, B improves CLS Macro-F1 by 4.21 pp over C and 4.23 points over D. The corresponding paired patient-bootstrap 95\% intervals are 3.37-5.04 and 3.42-5.02 points.

Because C retains the same auxiliary observations, the B-C difference shows that merely exposing the encoder to EEG waveforms does not account for the improvement. The B-D comparison further identifies synchronized correspondence as the useful training signal: preserving the marginal EEG and ECG inputs while breaking their relationship does not reproduce the gain.

Arm B also exceeds A in respiration-only CLS performance, despite receiving half as many ECG-respiration updates. Thus, allocating part of the training budget to a correctly aligned auxiliary relationship can be more useful than spending every update on pairs that directly contain the downstream sensor.

With patch-mean features, auxiliary alignment improves Macro-F1 by 0.56 pp over C and 0.53 pp over D, compared with CLS gains of 4.21 and 4.23 pp, respectively. The paired 95\% intervals for the patch-mean gains are -0.32 to 1.49 and -0.35 to 1.42 pp. The stronger CLS response is consistent with alignment acting directly on its projected representation. Uniform patch pooling exposes a different summary of the contextualized signal, through which the auxiliary relationship has a more modest measured effect.

This study demonstrates a practical consequence of learning over partially overlapping physiological observations. Correct EEG--ECG alignment improves respiration-only CLS representations when ECG serves as the shared modality between the EEG--ECG and ECG--respiration training pairs, even though EEG and respiration are never directly paired. The auxiliary relationship supplies useful structure to the unified encoder that remains available when its partner signals are absent at deployment. This result shows how auxiliary correspondence in fragmented recording collections can improve single-sensor inference through a unified encoder.

\Needspace{8\baselineskip}
\section{Discussion and conclusion}\label{discussion-and-conclusion}

\subsection{Shared physiological features and task accessibility}\label{shared-physiological-features-and-task-accessibility}

LingSante-Pulse addresses heterogeneous physiological learning at two levels. Synchronized alignment learns relationships from the available signal pairs, while generated-task prediction and reconstruction encourage each individual signal to retain readable structure. The four-domain comparison shows that this organization can support a competitive shared encoder across physiological sources. Multimodal sleep staging demonstrates joint use of its channel representations, while single-modality results show reuse under restricted sensing.

The controlled analyses help interpret how this representation is formed. Random teachers provide useful supervision because their outputs remain structured functions of the input. The ridge head adapts to each function; the encoder learns across their distribution. Prediction skill improves over the training trajectory and extends to new teacher draws after freezing the encoder, supporting the accessibility of this shared signal structure. Counterfactual diagnostics show how crop-based alignment suppresses useful temporal distinctions, while generated-task prediction keeps them accessible to a frozen learner. The synthetic study also shows complementary attribute preferences for prediction and reconstruction. The auxiliary-alignment study further shows that correct synchronized correspondence benefits respiration-only CLS representations, with smaller changes in patch mean. Together, these analyses explain how the pretraining objectives preserve temporal information, support generalization to new teachers and improve single-signal representations through auxiliary alignment.

\subsection{Conclusion}\label{conclusion}

LingSante-Pulse organizes heterogeneous recordings through synchronized pairwise alignment and trains within-signal information with structured random-feature prediction and masked reconstruction. Refreshed teachers supply changing measurements from a persistent family of signal functions, while support/query adaptation trains their accessibility. The resulting shared encoder supports task-specific use across four signal domains, multimodal sleep staging, and deployment with individual sensors.

\phantomsection
\addcontentsline{toc}{section}{References}
\begingroup
\addtolength{\bibsep}{-0.8pt}
\bibliographystyle{unsrtnat}
\bibliography{references}
\endgroup

\clearpage
\appendix
\addtocontents{toc}{\protect\setcounter{tocdepth}{1}}
\section{Implementation details}\label{appendix-a-implementation-details}

This section follows the implementation from waveform preparation and channel-wise encoding to the alignment, random-feature prediction and reconstruction objectives, ending with the complete pretraining update.

\subsection{Pretraining configuration}\label{a.1-pretraining-configuration}

\begin{figure}[H]
\phantomsection\label{fig:A1}
\centering
\includegraphics[width=\linewidth,height=.70\textheight,keepaspectratio]{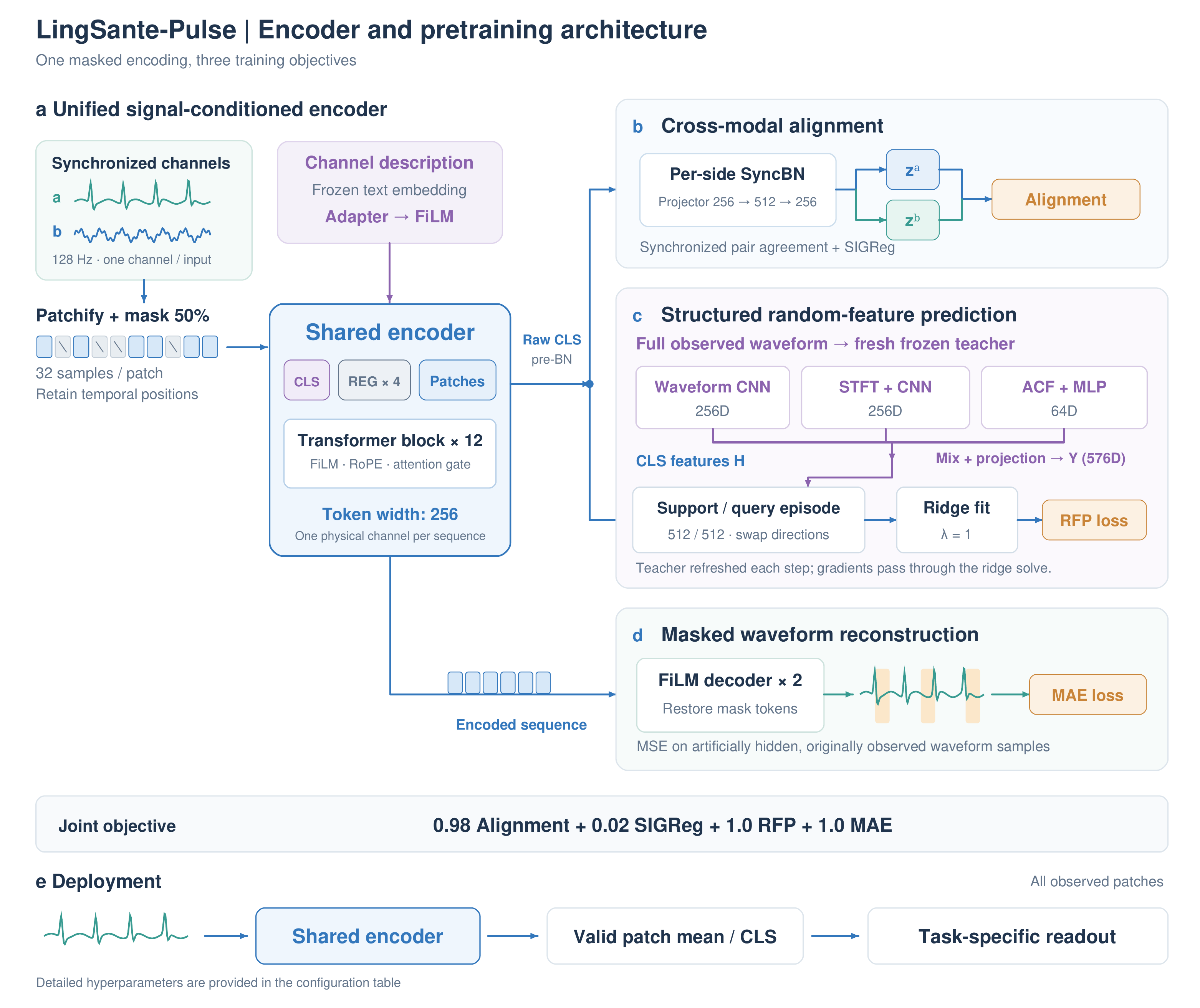}
\caption*{{\color{black}\textbf{Figure A1. Encoder and pretraining architecture.}} Each synchronized signal channel is encoded independently at 128 Hz using 32-sample patches, one CLS token and four register tokens. Alignment and RFP use the CLS representation; reconstruction uses the encoded sequence. All three objectives share one masked encoder pass per waveform channel. \hyperref[tab:A1]{Table~A1} lists the configuration in the same component order.}
\end{figure}

\Needspace{12\baselineskip}
\phantomsection\label{tab:A1}
{\color{black}\textbf{Table A1. Pretraining configuration.}} Components follow panels (a)--(e) of \hyperref[fig:A1]{Figure~A1}; Sections A.2--A.7 describe their implementation.

{
\begin{longtable}{@{}>{\raggedright\arraybackslash}p{\dimexpr.28\linewidth-\tabcolsep\relax}>{\raggedright\arraybackslash}p{\dimexpr.72\linewidth-\tabcolsep\relax}@{}}
\toprule
\rowcolor{PulsePale}Component & Configuration\\\midrule\endhead
\bottomrule\endlastfoot
\rowcolor{PulsePale}\multicolumn{2}{@{}l@{}}{\textbf{(a) Signal input and shared encoder}}\\
Input unit & One physical channel per encoder input\\
Sampling and patches & 128 Hz; 32 samples per patch, 0.25 s; physical-time RoPE\\
Windows & Waveform windows 30 s; SpO2 pairs 60 s; diagnostic ECG up to 10 s with observed-support masks\\
Normalization & Window mean and standard deviation over the observed samples of each physical channel\\
Low-rate numerical signals & Fixed scales; zero-order hold; all patches retained; alignment and SIGReg\\
Masking & Artificially hide 50\% of observed waveform patches; preserve physical positions; one masked student forward per side\\
Channel conditioning & Frozen 384D BGE channel descriptions; shared adapter; identity-initialized FiLM in all blocks\\
Encoder & 12 Transformer blocks; width 256; 8 heads; SwiGLU hidden width 1024\\
Prefix tokens and attention & One CLS and four learned registers; sigmoid attention-output gate\\
\rowcolor{PulsePale}\multicolumn{2}{@{}l@{}}{\textbf{(b) Cross-modal alignment}}\\
Projector & Per-side global SyncBN without learned affine parameters; 256--512--256 GELU projector\\
Alignment & Mean squared distance between synchronized projected CLS embeddings\\
\rowcolor{PulsePale}\multicolumn{2}{@{}l@{}}{\textbf{(c) Structured random-feature prediction}}\\
Teacher & New random teacher at each update; waveform/time-frequency/autocorrelation branches of 256/256/64 coordinates; fixed across support and query examples\\
Teacher target mixing & Channel-conditioned residual of width 32, \(\alpha=1\); random 576×576 projection\\
RFP episode and loss & 576D target MSE; 1024 examples per episode; support 512/query 512, both directions; support-set normalization; ridge penalty 1\\
\rowcolor{PulsePale}\multicolumn{2}{@{}l@{}}{\textbf{(d) Masked waveform reconstruction and joint objective}}\\
Decoder & Two FiLM-conditioned blocks; width 256; 8 heads; SwiGLU hidden 512; four registers and attention gate; output 32 samples/patch\\
MAE & MSE on artificially hidden, originally observed normalized waveform samples; per-window then eligible-side mean\\
Objective & 0.98 alignment + 0.02 SIGReg + 1 RFP + 1 MAE\\
\rowcolor{PulsePale}\multicolumn{2}{@{}l@{}}{\textbf{(e) Deployment}}\\
Deployment & All observed patches; per-channel CLS or valid patch mean; fixed-order channel-vector concatenation for multichannel tasks; discard teacher, decoder, ridge learner, SyncBN and projector\\
\end{longtable}
}

\subsection{Waveform preparation}\label{a.3-full-input-construction}

\textbf{Waveform signals.} Waveform channels are resampled to 128 Hz with antialiasing. For monitoring recordings, each 30-second window is standardized separately for each physical channel using only its observed samples. For a window with at least one observed sample, let $x_n$ be sample $n$ and let $o_n$ be one for an observed sample and zero otherwise:

\begingroup\small
\[
\mu_o=\frac{\sum_n o_n x_n}{\sum_n o_n},\qquad
\sigma_o=\left(\frac{\sum_n o_n(x_n-\mu_o)^2}{\sum_n o_n}\right)^{1/2},\qquad
\widetilde x_n=\begin{cases}(x_n-\mu_o)/\max(\sigma_o,10^{-6}),&o_n=1,\\0,&o_n=0\end{cases}.
\]
\endgroup

Here, $\mu_o$ and $\sigma_o$ are the mean and population standard deviation over observed samples. Missing samples and padding do not contribute to these statistics; $10^{-6}$ is the standard-deviation floor.

\Needspace{5\baselineskip}
\textbf{Low-rate numerical signals.} Physiological measurements recorded at low sampling rates, including oxygen saturation and vital-sign trends, are mapped to the common time grid by zero-order hold and scaled using fixed channel-specific constants. This preserves their physiological levels alongside the standardized waveform inputs.

\textbf{Sample masks and patch tokenization.} The sample-present mask identifies positions within the recorded sequence, separating them from padding; the observation mask identifies samples with available measurements. Every 32 samples form a patch, corresponding to 0.25 seconds at 128 Hz. One shared linear projection embeds the waveform values. A separate projection embeds the concatenated observation and sample-present masks, and its output is added to the waveform embedding. Entirely unobserved patches use a learned mask token; padding is excluded from attention.

\subsection{Signal-conditioned encoder}\label{a.4-full-conditional-encoder}

The encoder prepends one learned CLS and four register tokens to each single-channel waveform sequence and applies 12 Transformer blocks with rotary position encoding, pre-layer normalization, and SwiGLU feed-forward layers. The encoder uses width 256, eight attention heads, and a SwiGLU hidden width of 1024. Final LayerNorm followed by CLS selection produces \(h\).

A frozen text encoder produces 384-dimensional channel-description vectors, which a shared linear adapter and LayerNorm map to the model width. Descriptions specify the physiological modality and acquisition lead. Each block maps this condition to four vectors that modulate the attention and feed-forward inputs:

\[
\widetilde v_{\ell,m}=(1+\gamma_{\ell,m})\odot\operatorname{LN}(v_{\ell,m})+\beta_{\ell,m}.
\]

Here, \(v_{\ell,m}\) is the input to an attention or feed-forward sublayer in block \(\ell\) for channel \(m\). The channel-dependent vectors \(\gamma_{\ell,m},\beta_{\ell,m}\in\mathbb R^{256}\) scale and shift the normalized features. \(\operatorname{LN}\) denotes layer normalization, and \(\odot\) denotes element-wise multiplication. Separate modulation heads are used for the two sublayers.

Each attention block also applies a learned sigmoid gate to its attention output before the output projection. Gate weights and biases start at zero, giving initial sigmoid value 0.5. The modulation heads start at zero, so the FiLM transformation initially leaves the normalized features unchanged. All signals share the patch encoder, Transformer and condition adapter. Conditioning uses FiLM \citep{ref10} across signal types.

\textbf{Masked encoder and reconstruction decoder.} We select visible patches only from positions with observed source samples. Ragged visible-token batches are padded with invalid slots, which attention excludes. The remaining patches retain their original rotary coordinates; CLS and the four registers retain their prefix positions. The final normalized encoded tensor is shared by all objectives.

The decoder projects encoded tokens to width 256, restores the complete patch grid using a learned mask token, and applies two signal-conditioned Transformer blocks. Its own channel-description adapter and FiLM heads are trained independently of the encoder's conditioning heads and start with identity modulation. Final LayerNorm and a linear map predict 32 samples per patch. Reconstruction targets are the normalized waveform samples at artificially hidden, originally observed positions. Both encoder and decoder receive reconstruction gradients. The decoder is absent at inference.

\subsection{Cross-modal alignment and SIGReg}\label{a.8-exact-alignment-and-sigreg}

Synchronized waveform channels are encoded with 50\% artificial patch masking before alignment.

The CLS representation first passes through SyncBatchNorm without learned affine parameters, followed by a shared \(256\rightarrow512\rightarrow256\) projector with GELU. Normalization statistics are computed separately for each signal side over the global batch of 1024 examples; the same module maintains shared running statistics across sides.

Let \(z_i^a,z_i^b\in\mathbb R^K\) be the projected embeddings of synchronized pair \(i\), with \(K=256\) and batch size \(B=1024\). Their midpoint is \(c_i=(z_i^a+z_i^b)/2\). The alignment loss averages the squared deviations from this midpoint:

\[
\mathcal L_{\mathrm{align}}=\frac{1}{2BK}\sum_{i=1}^B
\left(\|z_i^a-c_i\|_2^2+\|z_i^b-c_i\|_2^2\right)
=\frac{1}{4BK}\sum_i\|z_i^a-z_i^b\|_2^2.
\]

The equivalent pairwise form on the right gives the normalization used in training. Gradients propagate through both projected embeddings.

We use SIGReg from LeJEPA \citep{ref01} to regularize each signal's projected batch. Let \(Z=\{z_i\}_{i=1}^B\), let \(w_s\in\mathbb R^K\) be a random unit direction, and let \(g(t)=e^{-t^2/2}\) be the characteristic function of a standard Gaussian. The regularizer compares the empirical cosine and sine averages along each direction with this reference:

\[
\mathcal R(Z)=\frac{2n_{\mathrm{ref}}}{S}\sum_{s=1}^S\int_0^{T}g(t)\left[\left(\frac1B\sum_i\cos(tw_s^\top z_i)-g(t)\right)^2+\left(\frac1B\sum_i\sin(tw_s^\top z_i)\right)^2\right]dt.
\]

Here, \(S=256\) is the number of sampled directions, \(T=3\) is the integration limit, and \(n_{\mathrm{ref}}=320\) sets the loss scale. The integral is approximated using 17 trapezoidal integration points on \([0,3]\). Empirical averages use the global batch separately for each signal side, and the two penalties are averaged as \(\mathcal L_{\mathrm{reg}}=[\mathcal R(Z^a)+\mathcal R(Z^b)]/2\).

\subsection{Structured random teacher}\label{a.6-structured-teacher-details}

\textbf{Teacher realization and inputs.} A new randomly initialized teacher is drawn for each training update and held fixed across all support and query examples in that episode. The teacher receives the normalized waveform before artificial patch masking and produces fixed targets for the episode; its parameters receive no gradients.

\textbf{Branch architecture.}

{
\begin{longtable}[]{@{}
  >{\raggedright\arraybackslash}p{(\linewidth - 1pt - 4\tabcolsep) * \real{0.3333}}
  >{\raggedright\arraybackslash}p{(\linewidth - 1pt - 4\tabcolsep) * \real{0.3333}}
  >{\raggedright\arraybackslash}p{(\linewidth - 1pt - 4\tabcolsep) * \real{0.3333}}@{}}
\toprule\noalign{}
\rowcolor{PulsePale}
\begin{minipage}[b]{\linewidth}\raggedright
Branch
\end{minipage} & \begin{minipage}[b]{\linewidth}\raggedright
Transformation and random network
\end{minipage} & \begin{minipage}[b]{\linewidth}\raggedright
Pooling and target width
\end{minipage} \\
\midrule\noalign{}
\endhead
\bottomrule\noalign{}
\endlastfoot
Waveform & Three Conv1D/GELU stages, channels \(1\to16\to16\to8\), kernel 9 and stride 2 & Coverage-aware pooling into 32 ordered temporal bins; \(8\times32=256\) \\
Time-frequency & Log-STFT with 2, 4, and 8 second Hann windows, 1 second hop, FFT size 1024; centered log magnitudes interpolated to 96 log-spaced frequencies from 0.2 to 31.5 Hz; random Conv2D/GELU & Four output maps pooled to 16 frequency by 4 time bins; \(4\times16\times4=256\) \\
Autocorrelation & Observed-pair-corrected, zero-lag-normalized autocorrelation at sample lags 2, 4, \ldots, 512 at 128 Hz, covering up to 4 seconds; random \(256\to128\to64\) MLP & 64 coordinates \\
\end{longtable}
}

\textbf{Channel-conditioned target mixing.} The three branch outputs are concatenated as \(u\in\mathbb R^{576}\). For a channel with identity \(c\), a randomly initialized residual transform of width 32 is followed by a random projection in the same 576-dimensional space. For coordinate \(k\), this gives

\[
r_k=\sum_{\ell=1}^{32}b_{\omega,c,\ell}\,\mathrm{GELU}(a_{\omega,c,\ell}\mathrm{LN}(u)_k),
\qquad T_{\omega,c}(x)=R_{\omega,c}(u+r).
\]

Here, \(\omega\) identifies the sampled teacher; \(a_{\omega,c,\ell}\) and \(b_{\omega,c,\ell}\) are its channel-dependent coefficients, \(\mathrm{LN}(u)\) normalizes the concatenated branch features, and \(R_{\omega,c}\in\mathbb R^{576\times576}\) is the random projection. These weights are redrawn at each update. The output \(T_{\omega,c}(x)\) combines measurements of waveform shape and temporal position, spectral organization and lag-dependent rhythm. Teacher prediction is applied to waveform channels; low-rate numerical signals contribute to alignment and SIGReg.

\subsection{Ridge adaptation and loss aggregation}\label{a.7-ridge-implementation-and-exact-weighting}

For each eligible waveform side, an RFP episode contains 1,024 examples, divided into support and query sets \(S\) and \(Q\) with \(n_S=n_Q=512\). Let \(H_S\in\mathbb R^{n_S\times256}\) and \(H_Q\in\mathbb R^{n_Q\times256}\) contain the CLS features, and let \(Y_S\in\mathbb R^{n_S\times576}\) and \(Y_Q\in\mathbb R^{n_Q\times576}\) contain the teacher targets. Each feature and target coordinate is standardized using its support-set mean and population standard deviation, with variance floor \(10^{-8}\); tildes denote the standardized matrices.

The episode-specific matrix \(W\in\mathbb R^{256\times576}\) maps the standardized CLS features to the current teacher targets. It is fitted on the support set by ridge regression:

\[
W^*=\arg\min_W\frac1{n_S}\|\widetilde H_SW-\widetilde Y_S\|_F^2+\lambda_r\|W\|_F^2,
\qquad\lambda_r=1.
\]

The first term measures support prediction error, and the second penalizes the squared Frobenius norm of the mapping. The implementation solves the 256-by-256 primal linear system for all 576 target coordinates together. The query predictions are \(\widehat Y_Q=\widetilde H_QW^*\), equivalently written in dual form as

\[
\widehat Y_Q=\widetilde H_Q\widetilde H_S^\top
(\widetilde H_S\widetilde H_S^\top+n_S\lambda_rI)^{-1}\widetilde Y_S.
\]

The sets are then exchanged, and both the standardization statistics and ridge mapping are refitted. For waveform side \(m\), \(\widehat Y_{Q,m}\) denotes predictions obtained by fitting on \(S\), while \(\widehat Y_{S,m}\) denotes predictions obtained by fitting on \(Q\). Averaging squared error over target coordinates, query examples, the two directions and the eligible sides \(\mathcal A\) gives

\[
\mathcal L_{\mathrm{RFP}}=
\frac1{|\mathcal A|}\sum_{m\in\mathcal A}\frac12\left[
\frac{\|\widehat Y_{Q,m}-\widetilde Y_{Q,m}\|_F^2}{576n_Q}
+\frac{\|\widehat Y_{S,m}-\widetilde Y_{S,m}\|_F^2}{576n_S}\right].
\]

Each direction uses targets standardized with its own support statistics. Gradients pass through the support and query features, standardization and ridge solve; teacher targets remain fixed within the episode.

\subsection{Joint pretraining algorithm}\label{a.9-joint-pretraining-algorithm}

Algorithm~\ref{alg:ling-pulse-pretraining} summarizes one pretraining update, from synchronized channel sampling and a shared masked encoding to the three objective terms and the encoder update.

\begin{algorithm}[H]
\caption{Joint Pretraining of LingSante-Pulse}
\label{alg:ling-pulse-pretraining}
\footnotesize
\begin{algorithmic}[1]
\Require Recording datasets \(\mathcal D\), signal conditions \(\{c_m\}\), teacher distribution \(p(\omega)\), updates \(T\), ridge penalty \(\lambda_r\), mask ratio \(\rho\), weights \(\lambda_s,\beta,\gamma\)
\Ensure Pretrained shared encoder \(F_\theta\)
\State Initialize encoder \(\theta\), projector \(\phi\), decoder \(\psi\), and optimizer
\For{\(t=1,\ldots,T\)}
  \State Sample a source and synchronized paired minibatches \((X^a,X^b)\) across the batch
  \State Let \(\mathcal A\subseteq\{a,b\}\) contain waveform sides; set \(\mathcal L_{\mathrm{MAE}}\gets0\)
  \For{\(m\in\{a,b\}\)}
    \State Sample observed-patch mask \(M^m\) at ratio \(\rho\) for \(m\in\mathcal A\); otherwise keep all patches
    \State Encode once: \(U^m\gets F_\theta(X^m_{\mathrm{visible}};c_m)\), preserving original positions
    \State Set \(H^m\gets U^m_{\mathrm{CLS}}\); project \(Z^m\gets P_\phi(\operatorname{SyncBN}(H^m))\)
    \If{\(m\in\mathcal A\)}
      \State Decode \(\widehat X^m\gets D_\psi(U^m,M^m;c_m)\); accumulate hidden observed-sample MSE
    \EndIf
  \EndFor
  \State Average reconstruction over eligible sides to obtain \(\mathcal L_{\mathrm{MAE}}\)
  \State Compute projected \(\mathcal L_{\mathrm{align}}\) and global per-side \(\mathcal L_{\mathrm{reg}}\)
  \State Draw full frozen channel-conditioned \(T_{\omega_t}\), including residual/projection, shared across the batch and sides; set \(\mathcal L_{\mathrm{RFP}}\gets0\)
  \For{\(m\in\mathcal A\)}
    \State Compute targets \(Y^m\gets T_{\omega_t}(X^m)\) from complete observed windows
    \State Gather \(H^m,Y^m\) across the batch; split the global episode into disjoint equal sets \(I,J\)
    \For{\((S,Q)\in\{(I,J),(J,I)\}\)}
      \State \((\widetilde H_S,\widetilde H_Q)\gets\operatorname{Standardize}_S(H_S^m,H_Q^m)\)
      \State \((\widetilde Y_S,\widetilde Y_Q)\gets\operatorname{Standardize}_S(Y_S^m,Y_Q^m)\)
      \State \(G_S\gets\widetilde H_S^\top\widetilde H_S+|S|\lambda_r I\)
      \State \(W_S\gets\operatorname{Solve}(G_S,\widetilde H_S^\top\widetilde Y_S)\); \(\widehat Y_Q\gets\widetilde H_QW_S\)
      \State \(\displaystyle\ell_m^{S\to Q}\gets\frac{\|\widehat Y_Q-\widetilde Y_Q\|_F^2}{576|Q|}\)
    \EndFor
    \State \(\displaystyle\mathcal L_{\mathrm{RFP}}\gets\mathcal L_{\mathrm{RFP}}+\frac{\ell_m^{I\to J}+\ell_m^{J\to I}}{2|\mathcal A|}\)
  \EndFor
  \State \(\mathcal L\gets(1-\lambda_s)\mathcal L_{\mathrm{align}}+\lambda_s\mathcal L_{\mathrm{reg}}+\beta\mathcal L_{\mathrm{RFP}}+\gamma\mathcal L_{\mathrm{MAE}}\)
  \State Backpropagate through encoder, decoder, differentiable gathering, standardization, and ridge solves
  \State Update \(\theta,\phi,\psi\) using the configured optimizer
\EndFor
\State \Return \(F_\theta\)
\end{algorithmic}
\vspace{0.3em}
\footnotesize
\(\operatorname{Standardize}_S\) fits statistics on \(S\) only. Teacher targets receive no gradients.
RFP and MAE are zero when no waveform side is eligible.
\end{algorithm}

\section{Alignment identities and proofs}\label{app:alignment-proofs}

This section supports the pairwise alignment objective in Section~\ref{pairwise-cross-modal-alignment}. We first relate pair sampling to alignment over a complete observation set, then extend the identities to observed subsets and collections of signal pairs linked by shared modalities.

\subsection{Complete observations and pair sampling}\label{c.1-complete-observations-and-pair-sampling}

Let \(z_1,\ldots,z_M\in\mathbb R^K\) be projected channel embeddings, with \(z_m=P_\phi(\operatorname{SyncBN}(h_{m,\mathrm{CLS}}))\), \(M\geq2\), and \(\bar z=M^{-1}\sum_m z_m\). Define the leave-one-out mean \(\bar z_{-m}=(M-1)^{-1}\sum_{j\ne m}z_j\), the complete pair sum \(L_{\mathrm{pair}}=\sum_{m<j}\|z_m-z_j\|_2^2\), and \(L_{\mathrm{LOO}}=\sum_m\|z_m-\bar z_{-m}\|_2^2\). Expanding the squared distances gives

\[
\sum_{m<j}\|z_m-z_j\|_2^2
=M\sum_m\|z_m\|_2^2-\left\|\sum_m z_m\right\|_2^2
=M\sum_m\|z_m-\bar z\|_2^2.
\]

Also, \(z_m-\bar z_{-m}=M(z_m-\bar z)/(M-1)\). Substitution yields

\[
\sum_m\|z_m-\bar z_{-m}\|_2^2
=\frac{M^2}{(M-1)^2}\sum_m\|z_m-\bar z\|_2^2
=\frac{M}{(M-1)^2}\sum_{m<j}\|z_m-z_j\|_2^2.
\]

Taking expectations over complete observations preserves both identities. If \(P=\binom M2\) and an unordered pair \((a,b)\) is sampled uniformly, then \(\mathbb E[P\|z_a-z_b\|_2^2]=L_{\mathrm{pair}}\). For nonuniform probabilities \(p_{ab}>0\), \(\|z_a-z_b\|_2^2/p_{ab}\) is an unbiased estimator of the same unnormalized complete sum. Without importance correction, the sampler instead defines a weighted pair objective. A pair with zero sampling probability cannot be included in this unbiased estimator. Under independent view sampling, the same statements apply to the joint distribution of per-modality views.

The constants depend on loss conventions. If both losses are averaged over their respective numbers of terms, then

\[
\frac1M L_{\mathrm{LOO}}
=\frac{M}{2(M-1)}\left(\frac{1}{\binom M2}L_{\mathrm{pair}}\right).
\]

A variable number of observed modalities changes this factor across observations. Pair and LOO losses must therefore be reweighted consistently before comparing corpus-level objectives. Proportional alignment terms have proportional gradients under the stated differentiation convention, but different sampling schemes can yield different stochastic training trajectories. Matching the full regularized objective also requires preserving the alignment-to-regularization ratio.

\subsection{Channel subsets and pair centers}\label{c.2-subset-consistency-and-the-implemented-view-center}

Let \(S\) be a nonempty subset of the observed channels and let \(\alpha_j\geq0\) satisfy \(\sum_{j\in S}\alpha_j=1\). For the pooled embedding \(z_S=\sum_{j\in S}\alpha_jz_j\), convexity of the squared Euclidean norm gives

\[
\left\|z_m-\sum_{j\in S}\alpha_jz_j\right\|_2^2
=\left\|\sum_{j\in S}\alpha_j(z_m-z_j)\right\|_2^2
\leq\sum_{j\in S}\alpha_j\|z_m-z_j\|_2^2.
\]

Thus, squared alignment error to a pooled subset is bounded by the corresponding weighted pairwise errors. For an \(L_g\)-Lipschitz prediction head \(f\), the prediction difference satisfies \(\|f(z_m)-f(z_S)\|_2^2\leq L_g^2\sum_{j\in S}\alpha_j\|z_m-z_j\|_2^2\). The result concerns convex pooling of the same observations, before any additional nonlinear pooling transformation or output normalization. Such transformations require their own stability assumptions. It does not compare task risks without assumptions connecting prediction error, labels, and the reference input.

For one synchronized pair \(i\), set \(u_1=z_i^a\), \(u_2=z_i^b\) and \(N=2\). Let \(q\in\{1,\ldots,N\}\) index these projected embeddings, let \(c\in\mathbb R^K\) be a center in the same space, and let \(\widetilde u=N^{-1}\sum_q u_q\). The centered cross term vanishes because \(\sum_q(u_q-\widetilde u)=0\), giving

\[
\sum_q\|u_q-c\|_2^2
=\sum_q\|u_q-\widetilde u\|_2^2
+N\|\widetilde u-c\|_2^2.
\]

The complete-pair identity applied to the \(N\) views gives \(\sum_q\|u_q-\widetilde u\|_2^2=N^{-1}\sum_{q<s}\|u_q-u_s\|_2^2\). Division by \(NK\) gives the view-center decomposition; averaging over batch items yields the corresponding alignment term. Setting \(c=\widetilde u\) gives \((NK)^{-1}\sum_q\|u_q-c\|_2^2=\|z_i^a-z_i^b\|_2^2/(4K)\). Averaging over the batch gives the alignment loss in Section 3.3 and Appendix A.4.

\subsection{Consistency on an incomplete observation graph}\label{c.3-consistency-on-an-incomplete-observation-graph}

For a common set of \(M\) synchronized embeddings, consider an undirected connected graph with positive edge weights \(w_{mj}\), counted once per unordered edge. Let \(\mathbf Z\in\mathbb R^{M\times K}\) stack the embeddings and \(L_{\mathcal G}\) be the unnormalized weighted graph Laplacian. Its alignment energy, summed over the observed edges \(\mathcal E\), is

\[
E_{\mathcal G}=\sum_{(m,j)\in\mathcal E}w_{mj}\|z_m-z_j\|_2^2
=\operatorname{tr}(\mathbf Z^\top L_{\mathcal G}\mathbf Z).
\]

Let \(\lambda_2(L_{\mathcal G})\) and \(\lambda_{\max}(L_{\mathcal G})\) denote the second-smallest and largest Laplacian eigenvalues. A connected graph has \(\lambda_2(L_{\mathcal G})>0\), and its Laplacian annihilates constant vectors. Therefore, the Rayleigh quotient bound on the centered columns of \(\mathbf Z\) gives

\[
\lambda_2(L_{\mathcal G})\sum_m\|z_m-\bar z\|_2^2
\leq E_{\mathcal G}
\leq\lambda_{\max}(L_{\mathcal G})\sum_m\|z_m-\bar z\|_2^2.
\]

Thus, energy on connected observed edges bounds complete dispersion for this common set of embeddings. Connectivity permits a common zero-energy consensus; the nonzero-error bound depends on edge weights and spectral connectivity. It neither makes arbitrary incomplete graphs equivalent to a complete graph nor makes their optimization trajectories identical. A disconnected graph provides no corresponding global lower bound.

The identities above apply to synchronized observations on the stated graph. Section 5.4 evaluates auxiliary-alignment transfer using controlled EEG--ECG and ECG--respiration training pairs with ECG as their shared modality.

\section{Proofs for random-feature prediction}\label{app:rfp-proofs}

This section develops the analysis in Section~\ref{theoretical-analysis-of-random-feature-prediction}. The task-kernel identity and coverage bound characterize the expected prediction objective and its relation to downstream targets. The linear example, temporal-separation argument and fresh-task analysis explain the roles of task diversity, information preservation and held-out prediction, respectively.

\subsection{Expected task-kernel identity}\label{d.1-expected-task-kernel-identity}

We first derive Proposition 1, expressing the mean query prediction loss across teacher realizations through a common task kernel.

Fix the encoder parameters \(\theta\), support and query inputs \(X_S,X_Q\), student patch masks and prediction direction, as in Section 3.6.1. Write \(n=|S|\) and \(q=|Q|\) for the support and query sizes. Student support-only centering and scaling determine \(\widetilde H_S,\widetilde H_Q\), so \(A_\theta\) is independent of a newly sampled teacher. For each joint mixed target coordinate \(j\), define the normalized channel-conditioned target function

\[
g_{\omega,c,j,S}(x)=
\frac{T_{\omega,c,j}(x)-\mu_{\omega,c,j,S}}
{\max(\sigma_{\omega,c,j,S},10^{-4})}.
\]

Here, \(\omega\) denotes the sampled teacher weights, \(c\) the physical channel identity and \(j\) an output coordinate; \(\mu_{\omega,c,j,S}\) and \(\sigma_{\omega,c,j,S}\) are its support-target mean and standard deviation. The teacher realization and uniformly sampled coordinate jointly define the scalar task \(\xi\) of Section 3.6.1. The denominator implements the variance floor \(10^{-8}\). The same fitted statistics are used for support and query. Choosing \(j\) uniformly among the 576 mixed coordinates makes the expected scalar squared error equal to the coordinate-averaged MSE. The channel identity \(c\) is fixed by the episode; the argument does not require independent output coordinates. The positive denominator floor preserves finite target second moments whenever the raw teacher outputs on the fixed episode have finite second moments, since its reciprocal is bounded by \(10^4\).

Write \(g=g_{\xi,S}=[g_S^\top,g_Q^\top]^\top\in\mathbb R^{n+q}\), with support targets followed by query targets. The matrix \(A_\theta\in\mathbb R^{q\times n}\) maps support targets to query predictions, and \(M_\theta=[-A_\theta,I_q]\) maps the stacked targets to prediction residuals. The mean squared query error is

\[
\ell_Q(\theta,\xi)=\frac1q\|A_\theta g_S-g_Q\|_2^2
=\frac1q\operatorname{tr}(M_\theta gg^\top M_\theta^\top).
\]

Finite second moments permit expectation through this finite trace, giving

\[
\mathbb E_\xi\,\ell_Q(\theta,\xi)
=\frac1q\operatorname{tr}
(M_\theta\mathbb E_\xi[gg^\top]M_\theta^\top).
\]

The expectation \(K_{T,S}=\mathbb E_\xi[gg^\top]\) is the task kernel, proving Proposition 1. Partitioning it into support and query blocks \(K_{SS},K_{SQ},K_{QS},K_{QQ}\) gives

\[
\frac1q\operatorname{tr}
(K_{QQ}-2A_\theta K_{SQ}+A_\theta K_{SS}A_\theta^\top).
\]

Both directions refit their own statistics and therefore have their own \(A_\theta\) and \(K_{T,S}\). Averaging the identities proves the bidirectional result. Signal sides may share the same teacher realization; linearity of expectation still permits averaging their individual losses.

The kernel is a second moment over normalized functions, not necessarily a mean-zero covariance. Its support dependence cannot generally be removed: replacing it with the raw teacher-output kernel changes the objective. No Gaussian-process limit or independent-coordinate approximation is used.

\subsection{Conditional task-coverage bound}\label{d.2-conditional-task-coverage-bound}

Starting from this kernel identity, we prove Proposition 2 and give a function-space condition for the teacher family to cover downstream targets.

If \(C K_{T,S}-K_{D,S}\succeq0\), then

\[
M_\theta(C K_{T,S}-K_{D,S})M_\theta^\top\succeq0.
\]

Let \(d_S\in\mathbb R^{n+q}\) stack a downstream task's support and query targets in the fixed evaluation scale used in Section 3.6.2. Then \(K_{D,S}=\mathbb E_d[d_Sd_S^\top]\), and its mean query error is \(\ell_D=q^{-1}\|M_\theta d_S\|_2^2\). Taking the trace and averaging over downstream and generated tasks on the fixed episode gives

\[
\mathbb E_d[\ell_D\mid\text{episode}]
=\frac1q\operatorname{tr}(M_\theta K_{D,S}M_\theta^\top)
\leq\frac Cq\operatorname{tr}(M_\theta K_{T,S}M_\theta^\top)
=C\,\mathbb E_\xi[\ell_Q\mid\text{episode}].
\]

Averaging over \(\nu\) proves Proposition 2 in the fixed downstream target scale.

For approximate coverage, let \(\epsilon\ge0\) and let \(I_{n+q}\) be the identity on the stacked target coordinates. Replace the left-hand difference with \(C K_{T,S}+\epsilon I_{n+q}-K_{D,S}\). The added term is

\[
\frac{\epsilon}{q}\operatorname{tr}(M_\theta M_\theta^\top)
=\epsilon\left(1+\frac{\|A_\theta\|_F^2}{q}\right).
\]

Averaging gives the approximate-coverage bound

\[
\mathcal R_D(\theta)\le C\mathcal R_T(\theta)
+\epsilon\,\mathbb E\!\left[1+\frac{\|A_\theta\|_F^2}{q}\right].
\]

\textbf{A sufficient function-space condition.} For a fixed support set, let \(k_{T,S}\) be the kernel defined by the same normalized random functions over inputs. If a downstream target function \(v_S\) belongs to its RKHS with norm at most \(B\), then its values on any finite input episode satisfy \(v_Sv_S^\top\preceq B^2K_{T,S}\). Indeed, for any coefficient vector \(u\), the reproducing property and Cauchy--Schwarz give

\[
(u^\top v_S)^2
=\left\langle v_S,\sum_i u_i k_{T,S}(x_i,\cdot)\right\rangle^2
\le B^2 u^\top K_{T,S}u.
\]

A distribution of such targets inherits the same bound when the norm bound is uniform. This is a sufficient coverage condition; membership and a useful norm bound are not established for every physiological endpoint. The targets and normalization used in this condition must match the evaluated risk.

\subsection{Finite-support linear illustration}\label{d.3-finite-support-linear-illustration}

This example makes task diversity explicit: its query risk depends on task-relevant signal directions omitted by the representation.

Assume \(x\in\mathbb R^d\) follows \(N(0,I_d)\). Each task has target \(y=w^\top x\), with weights \(w\in\mathbb R^d\) independent of the inputs and second-moment matrix \(\Sigma_w=\mathbb E[ww^\top]\). Let \(U\in\mathbb R^{d\times r}\) satisfy \(U^\top U=I_r\), and write \(P=UU^\top\) for the projection onto the represented subspace. For a fixed task,

\[
y=(U^\top x)^\top U^\top w+\varepsilon,
\qquad
\varepsilon=x^\top(I-P)w.
\]

Gaussian orthogonality makes \(\varepsilon\) independent of the represented features, with variance \(\rho_w=\|(I-P)w\|_2^2\). This residual is caused by omitted signal directions; no external label noise is assumed.

Let \(X_S\in\mathbb R^{n\times d}\) stack the support inputs, set \(Z=X_SU\), and fit ordinary least squares. With support residual vector \(\varepsilon_S\), \(\widehat\beta-U^\top w=(Z^\top Z)^{-1}Z^\top\varepsilon_S\). Conditional on \(Z,w\), the residuals have zero mean and covariance \(\rho_w I_n\), giving

\[
\mathbb E[\|\widehat\beta-U^\top w\|_2^2\mid Z,w]
=\rho_w\operatorname{tr}((Z^\top Z)^{-1}).
\]

For an independent query \(x_q\), let \(v_q=U^\top x_q\). Then \(\mathbb E[v_qv_q^\top]=I_r\), and \(v_q\) is independent of the zero-mean omitted-direction residual \(\varepsilon_q\). The prediction-error cross term therefore vanishes, giving query risk \(\|\widehat\beta-U^\top w\|_2^2+\rho_w\) conditional on the fitted head and \(w\). Since \(Z\) has independent standard Gaussian rows and \(n>r+1\),

\[
\mathbb E[(Z^\top Z)^{-1}]
=\frac{I_r}{n-r-1}.
\]

Averaging over input samples and using \(\mathbb E_w[\rho_w]=\operatorname{tr}((I-P)\mathbb E_w[ww^\top])\) gives

\[
\mathcal R_Q(U)
=\left(1+\frac{r}{n-r-1}\right)
\mathbb E_w[w^\top(I-P)w]
=\left(1+\frac{r}{n-r-1}\right)
\operatorname{tr}[(I-P)\Sigma_w].
\]

This establishes the finite-support linear illustration. For fixed rank, maximizing \(\operatorname{tr}(U^\top \Sigma_w U)\) selects leading eigendirections; the remaining risk is the prefactor times the sum of omitted eigenvalues.

For the three-dimensional example, take \(\Sigma_w=\operatorname{diag}(1,1,0)\) and \(r=2\). A fixed task \(w=e_1\) is perfectly representable by either \(U=[e_1,e_2]\) or \(U=[e_1,e_3]\). Under the task family with second moment \(\Sigma_w\), the first has zero risk and the second has risk \(1+2/(n-3)\). Thus, a fixed-task optimum need not be an optimum for the generated family.

If a fixed vector-valued teacher already spans both relevant directions, the distinction disappears. If \(\Sigma_w\) is isotropic and rank is fixed, every orthogonal rank-\(r\) subspace has the same average risk. The example therefore establishes a possible mechanism rather than universal superiority of refreshing.

This model omits support-fitted normalization and uses OLS. It does not replace the positive-ridge characterization in Proposition 1. It also does not alone prove superiority over a support-error training objective at fixed rank.

\subsection{Temporal invariance and readable target separation}\label{d.4-temporal-invariance-and-readable-target-separation}

We next relate invariance to loss of target separation, providing the mathematical motivation for the controlled temporal-context diagnostics in Appendix~\ref{b.7-counterfactual-representation-diagnostics}.

Suppose complete signals \(x,x'\) have the same local observation \(c\), and an aligned representation \(r\) satisfies \(\|r(x)-r(c)\|\le\epsilon\) and \(\|r(x')-r(c)\|\le\epsilon\). The triangle inequality gives \(\|r(x)-r(x')\|\le2\epsilon\). A linear head with \(\|w\|\le L\) has prediction separation at most \(2L\epsilon\).

If the absolute difference between the true scalar targets is \(\Delta\), the mean squared error on these two observations is at least

\[
\frac14(\Delta-2L\epsilon)_+^2.
\]

To see this, let errors be \(e_1,e_2\). Their difference has magnitude at least \((\Delta-2L\epsilon)_+\), and \((e_1^2+e_2^2)/2\ge(e_1-e_2)^2/4\).

This argument concerns the representation where invariance is imposed. A projector can suppress a distinction that remains in raw encoder features. The matched-data toy study evaluates frozen CLS and patch-mean encoder readouts; Appendix~\ref{b.2-matched-data-synthetic-study} specifies its full/full and full/crop comparison recipes, and Appendix~\ref{b.7-counterfactual-representation-diagnostics} tests recovery of the resulting temporal distinctions.

\subsection{Fresh tasks and held-out prediction}\label{d.5-fresh-tasks-and-held-out-prediction}

Finally, we distinguish sampling new prediction functions from fitting one fixed teacher, and query prediction from fitting support targets.

For a fixed student/episode, independent fresh teachers sample the expected loss in Proposition 1. If differentiation and expectation can be interchanged, the sampled gradient is unbiased for its expectation. The implementation's random-teacher sampling scheme is modeled by independent draws for this statement; individual output coordinates are not assumed to be independent draws.

For a globally fixed teacher \(\omega_0\), let \(G_{\omega_0,S}\in\mathbb R^{(n+q)\times576}\) stack its standardized targets across support and query examples. The corresponding empirical task kernel is

\[
\widehat K_{\omega_0,S}
=\frac1{576}G_{\omega_0,S}G_{\omega_0,S}^\top.
\]

A sufficient fixed family can cover the same useful directions as the population kernel. Refreshing samples more functions, but finite optimization, task variance, and teacher coverage determine its empirical benefit. The evolving training trajectory is not a kernel estimate evaluated at one fixed student.

Support-error supervision uses a different residual operator from query supervision. In the unregularized overparameterized limit, a full-row-rank support feature matrix can interpolate arbitrary targets; after centering, spanning the \(n-1\) dimensional centered target subspace gives the analogous example. Such interpolation need not imply correct prediction outside support. Positive ridge in the actual configuration prevents importing this literal zero-error argument unchanged. Related theory on train--validation splitting establishes benefits under explicit linear-subspace assumptions, rather than a universal ordering of split and non-split learning \citep{ref15}.

\section{Controlled signal and random-teacher analyses}\label{app:controlled}

This section addresses two questions from Sections 5.2 and 5.3: which signal attributes different objectives preserve, and whether learning accumulates across changing teachers. Controlled signals and paired temporal diagnostics address the first question; online prediction and independent-teacher evaluations address the second.

\subsection{Controlled signals: setup and complete results}\label{b.2-matched-data-synthetic-study}

The controlled study in Section 5.2 compares recovery of known waveform attributes under matched signal exposure and model capacity. The following sections define the signals, training objectives and evaluation measures.

\subsubsection{Signal construction and independent recovery tasks}\label{b.2.1-signal-construction-and-independent-recovery-tasks}

Each waveform spans 30 seconds at 64 Hz (1,920 samples). Its generator uses 60 half-second bins, while the encoder uses 120 patches of 16 samples. Four families enter the training stream with probabilities 1/6, 1/6, 1/3 and 1/3 for amplitude, event location, spectral mixture and rhythm. Each latent group defines two counterfactual members; training samples one member per group. Probe fit/validation/test retain both members and keep each group within one split.

Amplitude examples modulate a 4/6/8/10-Hz carrier by a smooth envelope and reverse the bin order in the paired member. Their targets are the 60-bin relative log-power profile and its 59 successive differences. Event examples increase local amplitude for 0.5, 1 or 2 seconds and translate the event by 15 seconds; targets are normalized center, duration and a 60-bin event mask. Spectral examples mix low frequencies of 2/4 Hz with high frequencies of 10/12/14 Hz while preserving each bin's total power; targets recover the 60-bin high-frequency fraction and its 59 changes. Rhythm examples contain 25--36 Gaussian pulses. Reordering the same latent intervals preserves their mean and population SD but changes successive-interval differences; the targets are Mean IBI, SDNN and RMSSD computed directly from those continuous intervals, with SDNN calculated as the population standard deviation.

Two additional diagnostic families are evaluated outside the training stream. Motif-order examples exchange motifs with matching local power and Fourier magnitude; shared-crop examples retain identical waveforms from 10 to 20 seconds while changing the outside envelope. The complete panel has 14 tasks: 13 regressions and one motif-order classification. \hyperref[fig:8]{Figure~8} and the supplementary rank summary use the same six predefined regression targets: amplitude profile, event center, spectral fraction, Mean IBI, SDNN and RMSSD.

Each family contains 512/256/512 latent groups for probe fit/validation/test, giving 1,024/512/1,024 waveforms. Input and target scales are fitted on the fit split only. Independent frozen ridge probes select one of nine regularization values from \(10^{-6}\) to \(10^2\) on validation, with larger regularization breaking exact ties. NMSE averages squared residuals normalized by each target coordinate's fit SD; the motif-order binary task retains a separate classification error. SD across training runs reflects three training repetitions on one common evaluation panel.

\subsubsection{Student architecture and training objectives}\label{b.2.2-shared-training-contract-and-objectives}

Students use six Transformer layers, width 64, four heads, SwiGLU hidden width 256 and patch size 16, with one CLS token, one auxiliary register token with constant inputs, and patch tokens. Complete inputs have 122 tokens; 50\% masking physically selects 60 patches plus the two prefix tokens and preserves original RoPE positions. The MAE decoder has two layers, width 64 and hidden width 128. Frozen patch mean averages all 120 contextualized patch tokens, excluding the two prefix tokens; evaluation uses the complete observed input and discards training-only heads.

InfoNCE and LeJEPA use full/full or full/random-15-second-crop observations with a 64--128--64 projector. Crop starts align to patches and the crop is encoded without its original full-window offset. InfoNCE has temperature 0.2; LeJEPA weights alignment and SIGReg by 38/39 and 1/39. SIGReg-only retains the regularizer without alignment. Fixed-teacher direct distillation uses a learned linear prediction head and SIGReg of 0.025 per view. MAE reconstructs 50\% masked patches. Unmasked RFP predicts structured random features through CLS with a differentiable support-fitted ridge learner. The joint objective is \(\mathcal L_{\mathrm{RFP}}+\mathcal L_{\mathrm{MAE}}\), with unit weights, one masked student encoding and no cross-view alignment or SIGReg term.

The structured teacher combines raw-waveform Conv1D, multiscale log-STFT Conv2D and normalized-autocorrelation MLP branches with output widths 256/256/64. A frozen teacher is sampled each update; the three training repetitions have independent teacher sequences. Each direction of the RFP episode contains 512 support and 128 query examples. A 128-example core swaps two 64-example halves between directions, supplemented in each direction by 448 support and 64 query examples. Support-fitted feature/target standardization precedes ridge regularization \(n_S\alpha I\), where \(\alpha=1\). The loss averages both directions and the three branches, differentiating through the support features, normalization and ridge solve. Teacher targets are held fixed during the student update. Auxiliary fixed-teacher and support-error configurations are also evaluated.

\subsubsection{Complete scores and aggregate comparisons}\label{b.2.3-fixed-field-rank-and-complete-primary-task-scores}

\begin{figure}[H]
\phantomsection\label{fig:synthetic-readout-ranks}
\centering
\includegraphics[width=\linewidth,height=.70\textheight,keepaspectratio]{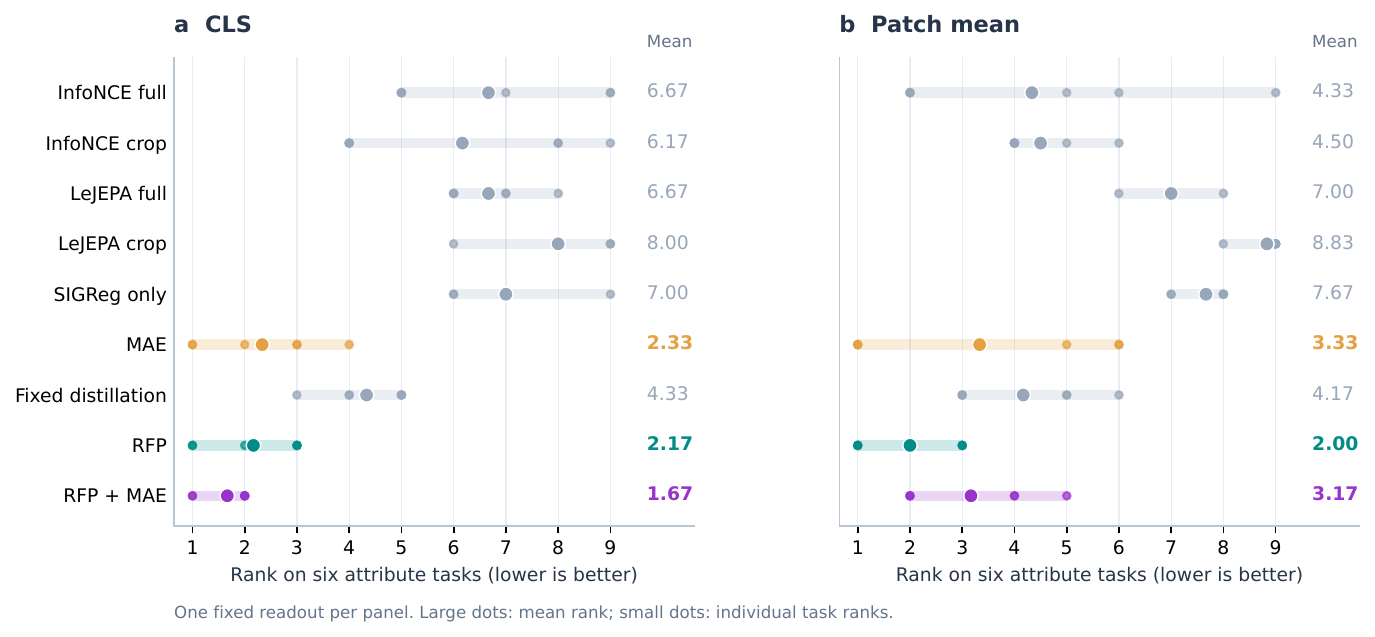}
\caption*{{\color{black}\textbf{Figure D1. Task ranks under two fixed readouts.}} Both panels contain the same nine configurations and six tasks as \hyperref[fig:8]{Figure~8}. Large dots are equal-weight mean ranks and small dots individual task ranks. CLS and patch mean are ranked separately.}
\end{figure}

\hyperref[fig:8]{Figure~8} compares the same nine configurations using each readout. \hyperref[fig:synthetic-readout-ranks]{Figure~D1} ranks their unrounded mean test NMSE separately for CLS and patch mean; ties receive average ranks and the six task ranks are weighted equally. Auxiliary teacher/objective variants appear in \hyperref[tab:synthetic-task-errors]{Table~D1} outside these nine-configuration fields.

\clearpage
\phantomsection\label{tab:synthetic-task-errors}
{\color{black}\textbf{Table D1. Primary synthetic tasks from the matched-data study.}} Primary-test NMSE at observation noise 0.3, mean ± sample SD over three training repetitions. CLS and patch mean are reported separately with unchanged task order. The first block contains the nine primary configurations; the final three rows compare fixed-teacher and support-error objectives. The native 576D random teacher is a separate reference.

\textbf{CLS.}

\begingroup\footnotesize\setlength{\tabcolsep}{4pt}\renewcommand{\arraystretch}{1.18}

\begin{longtable}{@{}p{3.2cm}rrrrrr@{}}\toprule

\rowcolor{PulsePale}Method & Amplitude & Event center & Spectral mix & Mean IBI & SDNN & RMSSD\\\midrule\endhead

Untrained encoder & \shortstack[r]{0.5761\\[-1pt]{\scriptsize$\pm$\,0.0369}} & \shortstack[r]{0.5871\\[-1pt]{\scriptsize$\pm$\,0.0944}} & \shortstack[r]{0.3102\\[-1pt]{\scriptsize$\pm$\,0.0530}} & \shortstack[r]{0.5452\\[-1pt]{\scriptsize$\pm$\,0.0671}} & \shortstack[r]{1.1084\\[-1pt]{\scriptsize$\pm$\,0.0001}} & \shortstack[r]{1.1099\\[-1pt]{\scriptsize$\pm$\,0.0033}}\\

InfoNCE full & \shortstack[r]{0.5428\\[-1pt]{\scriptsize$\pm$\,0.0367}} & \shortstack[r]{0.6247\\[-1pt]{\scriptsize$\pm$\,0.0420}} & \shortstack[r]{0.6120\\[-1pt]{\scriptsize$\pm$\,0.0169}} & \shortstack[r]{0.8096\\[-1pt]{\scriptsize$\pm$\,0.0315}} & \shortstack[r]{1.1090\\[-1pt]{\scriptsize$\pm$\,0.0011}} & \shortstack[r]{1.1123\\[-1pt]{\scriptsize$\pm$\,0.0049}}\\

InfoNCE crop & \shortstack[r]{1.0051\\[-1pt]{\scriptsize$\pm$\,0.0181}} & \shortstack[r]{1.0083\\[-1pt]{\scriptsize$\pm$\,0.0001}} & \shortstack[r]{1.0033\\[-1pt]{\scriptsize$\pm$\,0.0001}} & \shortstack[r]{0.0470\\[-1pt]{\scriptsize$\pm$\,0.0041}} & \shortstack[r]{1.0538\\[-1pt]{\scriptsize$\pm$\,0.0562}} & \shortstack[r]{1.0320\\[-1pt]{\scriptsize$\pm$\,0.1026}}\\

LeJEPA full & \shortstack[r]{0.8779\\[-1pt]{\scriptsize$\pm$\,0.0594}} & \shortstack[r]{0.8433\\[-1pt]{\scriptsize$\pm$\,0.1110}} & \shortstack[r]{0.6423\\[-1pt]{\scriptsize$\pm$\,0.1443}} & \shortstack[r]{0.8873\\[-1pt]{\scriptsize$\pm$\,0.0061}} & \shortstack[r]{1.1082\\[-1pt]{\scriptsize$\pm$\,0.0004}} & \shortstack[r]{1.1095\\[-1pt]{\scriptsize$\pm$\,0.0016}}\\

LeJEPA crop & \shortstack[r]{0.9946\\[-1pt]{\scriptsize$\pm$\,0.0002}} & \shortstack[r]{1.0104\\[-1pt]{\scriptsize$\pm$\,0.0020}} & \shortstack[r]{1.0036\\[-1pt]{\scriptsize$\pm$\,0.0006}} & \shortstack[r]{0.7198\\[-1pt]{\scriptsize$\pm$\,0.0108}} & \shortstack[r]{1.1086\\[-1pt]{\scriptsize$\pm$\,0.0026}} & \shortstack[r]{1.1113\\[-1pt]{\scriptsize$\pm$\,0.0034}}\\

SIGReg only & \shortstack[r]{0.9099\\[-1pt]{\scriptsize$\pm$\,0.0581}} & \shortstack[r]{0.9489\\[-1pt]{\scriptsize$\pm$\,0.0924}} & \shortstack[r]{0.7406\\[-1pt]{\scriptsize$\pm$\,0.0540}} & \shortstack[r]{0.8910\\[-1pt]{\scriptsize$\pm$\,0.0042}} & \shortstack[r]{1.1079\\[-1pt]{\scriptsize$\pm$\,0.0003}} & \shortstack[r]{1.1083\\[-1pt]{\scriptsize$\pm$\,0.0003}}\\

MAE, 50\% mask & \shortstack[r]{0.4052\\[-1pt]{\scriptsize$\pm$\,0.0454}} & \shortstack[r]{0.3360\\[-1pt]{\scriptsize$\pm$\,0.0423}} & \shortstack[r]{0.1443\\[-1pt]{\scriptsize$\pm$\,0.0452}} & \shortstack[r]{0.0134\\[-1pt]{\scriptsize$\pm$\,0.0009}} & \shortstack[r]{0.1032\\[-1pt]{\scriptsize$\pm$\,0.0224}} & \shortstack[r]{0.0898\\[-1pt]{\scriptsize$\pm$\,0.0146}}\\

Fixed-teacher distillation & \shortstack[r]{0.4521\\[-1pt]{\scriptsize$\pm$\,0.0477}} & \shortstack[r]{0.3628\\[-1pt]{\scriptsize$\pm$\,0.1178}} & \shortstack[r]{0.0606\\[-1pt]{\scriptsize$\pm$\,0.0018}} & \shortstack[r]{0.0547\\[-1pt]{\scriptsize$\pm$\,0.0223}} & \shortstack[r]{1.1067\\[-1pt]{\scriptsize$\pm$\,0.0016}} & \shortstack[r]{1.0431\\[-1pt]{\scriptsize$\pm$\,0.0919}}\\

RFP, unmasked & \shortstack[r]{0.1918\\[-1pt]{\scriptsize$\pm$\,0.0052}} & \shortstack[r]{0.1777\\[-1pt]{\scriptsize$\pm$\,0.0026}} & \shortstack[r]{0.0266\\[-1pt]{\scriptsize$\pm$\,0.0027}} & \shortstack[r]{0.0340\\[-1pt]{\scriptsize$\pm$\,0.0049}} & \shortstack[r]{0.1246\\[-1pt]{\scriptsize$\pm$\,0.0096}} & \shortstack[r]{0.0991\\[-1pt]{\scriptsize$\pm$\,0.0038}}\\

RFP + MAE & \shortstack[r]{0.2097\\[-1pt]{\scriptsize$\pm$\,0.0137}} & \shortstack[r]{0.1640\\[-1pt]{\scriptsize$\pm$\,0.0156}} & \shortstack[r]{0.0300\\[-1pt]{\scriptsize$\pm$\,0.0043}} & \shortstack[r]{0.0185\\[-1pt]{\scriptsize$\pm$\,0.0025}} & \shortstack[r]{0.0943\\[-1pt]{\scriptsize$\pm$\,0.0126}} & \shortstack[r]{0.0989\\[-1pt]{\scriptsize$\pm$\,0.0155}}\\

Fixed RFP, unmasked & \shortstack[r]{0.2341\\[-1pt]{\scriptsize$\pm$\,0.0300}} & \shortstack[r]{0.1791\\[-1pt]{\scriptsize$\pm$\,0.0380}} & \shortstack[r]{0.0269\\[-1pt]{\scriptsize$\pm$\,0.0006}} & \shortstack[r]{0.0273\\[-1pt]{\scriptsize$\pm$\,0.0007}} & \shortstack[r]{0.1416\\[-1pt]{\scriptsize$\pm$\,0.0266}} & \shortstack[r]{0.1088\\[-1pt]{\scriptsize$\pm$\,0.0397}}\\

MAE + fixed RFP & \shortstack[r]{0.2355\\[-1pt]{\scriptsize$\pm$\,0.0336}} & \shortstack[r]{0.1576\\[-1pt]{\scriptsize$\pm$\,0.0499}} & \shortstack[r]{0.0238\\[-1pt]{\scriptsize$\pm$\,0.0047}} & \shortstack[r]{0.0117\\[-1pt]{\scriptsize$\pm$\,0.0010}} & \shortstack[r]{0.0920\\[-1pt]{\scriptsize$\pm$\,0.0202}} & \shortstack[r]{0.0876\\[-1pt]{\scriptsize$\pm$\,0.0129}}\\

MAE + support-loss RFP & \shortstack[r]{0.1982\\[-1pt]{\scriptsize$\pm$\,0.0055}} & \shortstack[r]{0.1873\\[-1pt]{\scriptsize$\pm$\,0.0064}} & \shortstack[r]{0.0297\\[-1pt]{\scriptsize$\pm$\,0.0012}} & \shortstack[r]{0.0163\\[-1pt]{\scriptsize$\pm$\,0.0023}} & \shortstack[r]{0.0792\\[-1pt]{\scriptsize$\pm$\,0.0145}} & \shortstack[r]{0.0844\\[-1pt]{\scriptsize$\pm$\,0.0108}}\\

\bottomrule\end{longtable}\endgroup

\clearpage
\textbf{Patch mean.}

\begingroup\footnotesize\setlength{\tabcolsep}{4pt}\renewcommand{\arraystretch}{1.18}

\begin{longtable}{@{}p{3.2cm}rrrrrr@{}}\toprule

\rowcolor{PulsePale}Method & Amplitude & Event center & Spectral mix & Mean IBI & SDNN & RMSSD\\\midrule\endhead

Untrained encoder & \shortstack[r]{0.6381\\[-1pt]{\scriptsize$\pm$\,0.0261}} & \shortstack[r]{0.5530\\[-1pt]{\scriptsize$\pm$\,0.0887}} & \shortstack[r]{0.4132\\[-1pt]{\scriptsize$\pm$\,0.0703}} & \shortstack[r]{0.4796\\[-1pt]{\scriptsize$\pm$\,0.0140}} & \shortstack[r]{1.1085\\[-1pt]{\scriptsize$\pm$\,0.0000}} & \shortstack[r]{1.1085\\[-1pt]{\scriptsize$\pm$\,0.0001}}\\

InfoNCE full & \shortstack[r]{0.5021\\[-1pt]{\scriptsize$\pm$\,0.0305}} & \shortstack[r]{0.3221\\[-1pt]{\scriptsize$\pm$\,0.0797}} & \shortstack[r]{0.2105\\[-1pt]{\scriptsize$\pm$\,0.0209}} & \shortstack[r]{0.1871\\[-1pt]{\scriptsize$\pm$\,0.0311}} & \shortstack[r]{1.1081\\[-1pt]{\scriptsize$\pm$\,0.0000}} & \shortstack[r]{1.1071\\[-1pt]{\scriptsize$\pm$\,0.0002}}\\

InfoNCE crop & \shortstack[r]{0.7054\\[-1pt]{\scriptsize$\pm$\,0.1173}} & \shortstack[r]{0.9486\\[-1pt]{\scriptsize$\pm$\,0.0861}} & \shortstack[r]{0.5571\\[-1pt]{\scriptsize$\pm$\,0.1437}} & \shortstack[r]{0.0592\\[-1pt]{\scriptsize$\pm$\,0.0219}} & \shortstack[r]{0.9305\\[-1pt]{\scriptsize$\pm$\,0.0211}} & \shortstack[r]{0.9815\\[-1pt]{\scriptsize$\pm$\,0.0168}}\\

LeJEPA full & \shortstack[r]{0.9596\\[-1pt]{\scriptsize$\pm$\,0.0544}} & \shortstack[r]{0.9991\\[-1pt]{\scriptsize$\pm$\,0.0155}} & \shortstack[r]{0.7179\\[-1pt]{\scriptsize$\pm$\,0.0579}} & \shortstack[r]{0.6600\\[-1pt]{\scriptsize$\pm$\,0.0200}} & \shortstack[r]{1.1081\\[-1pt]{\scriptsize$\pm$\,0.0001}} & \shortstack[r]{1.1074\\[-1pt]{\scriptsize$\pm$\,0.0002}}\\

LeJEPA crop & \shortstack[r]{0.9953\\[-1pt]{\scriptsize$\pm$\,0.0011}} & \shortstack[r]{1.0095\\[-1pt]{\scriptsize$\pm$\,0.0025}} & \shortstack[r]{0.7485\\[-1pt]{\scriptsize$\pm$\,0.0209}} & \shortstack[r]{0.7046\\[-1pt]{\scriptsize$\pm$\,0.0075}} & \shortstack[r]{1.1081\\[-1pt]{\scriptsize$\pm$\,0.0001}} & \shortstack[r]{1.1076\\[-1pt]{\scriptsize$\pm$\,0.0000}}\\

SIGReg only & \shortstack[r]{0.9581\\[-1pt]{\scriptsize$\pm$\,0.0531}} & \shortstack[r]{1.0040\\[-1pt]{\scriptsize$\pm$\,0.0071}} & \shortstack[r]{0.7430\\[-1pt]{\scriptsize$\pm$\,0.0289}} & \shortstack[r]{0.6679\\[-1pt]{\scriptsize$\pm$\,0.0174}} & \shortstack[r]{1.1081\\[-1pt]{\scriptsize$\pm$\,0.0002}} & \shortstack[r]{1.1075\\[-1pt]{\scriptsize$\pm$\,0.0002}}\\

MAE, 50\% mask & \shortstack[r]{0.7966\\[-1pt]{\scriptsize$\pm$\,0.0438}} & \shortstack[r]{0.7721\\[-1pt]{\scriptsize$\pm$\,0.1183}} & \shortstack[r]{0.6411\\[-1pt]{\scriptsize$\pm$\,0.0458}} & \shortstack[r]{0.0043\\[-1pt]{\scriptsize$\pm$\,0.0003}} & \shortstack[r]{0.0474\\[-1pt]{\scriptsize$\pm$\,0.0055}} & \shortstack[r]{0.0608\\[-1pt]{\scriptsize$\pm$\,0.0098}}\\

Fixed-teacher distillation & \shortstack[r]{0.6193\\[-1pt]{\scriptsize$\pm$\,0.0472}} & \shortstack[r]{0.6196\\[-1pt]{\scriptsize$\pm$\,0.0532}} & \shortstack[r]{0.3769\\[-1pt]{\scriptsize$\pm$\,0.0400}} & \shortstack[r]{0.2929\\[-1pt]{\scriptsize$\pm$\,0.0578}} & \shortstack[r]{1.1080\\[-1pt]{\scriptsize$\pm$\,0.0002}} & \shortstack[r]{1.1064\\[-1pt]{\scriptsize$\pm$\,0.0013}}\\

RFP, unmasked & \shortstack[r]{0.3832\\[-1pt]{\scriptsize$\pm$\,0.0395}} & \shortstack[r]{0.2965\\[-1pt]{\scriptsize$\pm$\,0.0442}} & \shortstack[r]{0.0938\\[-1pt]{\scriptsize$\pm$\,0.0205}} & \shortstack[r]{0.0533\\[-1pt]{\scriptsize$\pm$\,0.0138}} & \shortstack[r]{0.3704\\[-1pt]{\scriptsize$\pm$\,0.0904}} & \shortstack[r]{0.5071\\[-1pt]{\scriptsize$\pm$\,0.0424}}\\

RFP + MAE & \shortstack[r]{0.6517\\[-1pt]{\scriptsize$\pm$\,0.0544}} & \shortstack[r]{0.6863\\[-1pt]{\scriptsize$\pm$\,0.1290}} & \shortstack[r]{0.5608\\[-1pt]{\scriptsize$\pm$\,0.0804}} & \shortstack[r]{0.0075\\[-1pt]{\scriptsize$\pm$\,0.0019}} & \shortstack[r]{0.1006\\[-1pt]{\scriptsize$\pm$\,0.0208}} & \shortstack[r]{0.0872\\[-1pt]{\scriptsize$\pm$\,0.0102}}\\

Fixed RFP, unmasked & \shortstack[r]{0.4108\\[-1pt]{\scriptsize$\pm$\,0.0410}} & \shortstack[r]{0.2756\\[-1pt]{\scriptsize$\pm$\,0.0084}} & \shortstack[r]{0.0857\\[-1pt]{\scriptsize$\pm$\,0.0095}} & \shortstack[r]{0.0489\\[-1pt]{\scriptsize$\pm$\,0.0049}} & \shortstack[r]{0.3797\\[-1pt]{\scriptsize$\pm$\,0.0286}} & \shortstack[r]{0.5057\\[-1pt]{\scriptsize$\pm$\,0.0943}}\\

MAE + fixed RFP & \shortstack[r]{0.6564\\[-1pt]{\scriptsize$\pm$\,0.0700}} & \shortstack[r]{0.6976\\[-1pt]{\scriptsize$\pm$\,0.0631}} & \shortstack[r]{0.5502\\[-1pt]{\scriptsize$\pm$\,0.0913}} & \shortstack[r]{0.0068\\[-1pt]{\scriptsize$\pm$\,0.0015}} & \shortstack[r]{0.0689\\[-1pt]{\scriptsize$\pm$\,0.0101}} & \shortstack[r]{0.0826\\[-1pt]{\scriptsize$\pm$\,0.0186}}\\

MAE + support-loss RFP & \shortstack[r]{0.6400\\[-1pt]{\scriptsize$\pm$\,0.0807}} & \shortstack[r]{0.7181\\[-1pt]{\scriptsize$\pm$\,0.0576}} & \shortstack[r]{0.5876\\[-1pt]{\scriptsize$\pm$\,0.0504}} & \shortstack[r]{0.0060\\[-1pt]{\scriptsize$\pm$\,0.0002}} & \shortstack[r]{0.0770\\[-1pt]{\scriptsize$\pm$\,0.0074}} & \shortstack[r]{0.0808\\[-1pt]{\scriptsize$\pm$\,0.0054}}\\

\bottomrule\end{longtable}\endgroup

\textbf{Native teacher reference.}

\begingroup\footnotesize\setlength{\tabcolsep}{4pt}\renewcommand{\arraystretch}{1.18}

\begin{longtable}{@{}p{3.2cm}rrrrrr@{}}\toprule

\rowcolor{PulsePale}Method & Amplitude & Event center & Spectral mix & Mean IBI & SDNN & RMSSD\\\midrule\endhead

Native random teacher (576D) & \shortstack[r]{0.2311\\[-1pt]{\scriptsize$\pm$\,0.0783}} & \shortstack[r]{0.1694\\[-1pt]{\scriptsize$\pm$\,0.0406}} & \shortstack[r]{0.0336\\[-1pt]{\scriptsize$\pm$\,0.0053}} & \shortstack[r]{0.0361\\[-1pt]{\scriptsize$\pm$\,0.0025}} & \shortstack[r]{0.1982\\[-1pt]{\scriptsize$\pm$\,0.0326}} & \shortstack[r]{0.1009\\[-1pt]{\scriptsize$\pm$\,0.0348}}\\

\bottomrule\end{longtable}\endgroup

\Needspace{7\baselineskip}
\paragraph{Aggregate comparisons and objective complementarity.}\label{b.6-objective-and-readout-comparisons}

\begin{figure}[H]
\phantomsection\label{fig:synthetic-complementarity}
\centering
\includegraphics[width=\linewidth,height=.70\textheight,keepaspectratio]{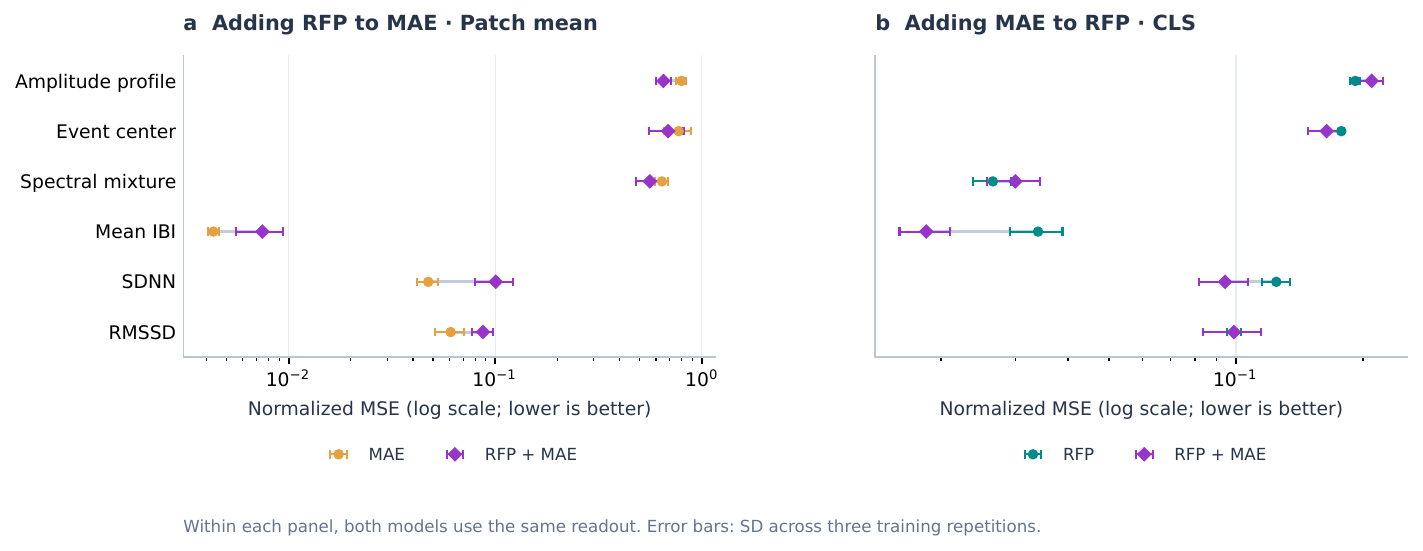}
\caption*{{\color{black}\textbf{Figure D2. Complementary attributes under fixed readouts.}} (a) Adding RFP to MAE, evaluated with patch mean. (b) Adding MAE to RFP, evaluated with CLS. Points and error bars show mean NMSE and sample SD across three training repetitions. MAE-containing models use 50\% visible patches; pure RFP uses complete inputs.}
\end{figure}

The two readouts expose different responses to the same training objectives. \hyperref[fig:synthetic-readout-ranks]{Figure~D1} summarizes the six-task ranks with the comparison field held fixed. For CLS, joint training has mean rank 1.67 and RFP 2.17; for patch mean, RFP has mean rank 2.00 and joint training 3.17. \hyperref[fig:synthetic-complementarity]{Figure~D2} then compares the joint objective with MAE through patch mean and with RFP through CLS. These within-readout contrasts complement the full error matrix in \hyperref[fig:8]{Figure~8}.

\subsection{Temporal-context diagnostics with controlled waveform pairs}\label{b.7-counterfactual-representation-diagnostics}

These diagnostics test whether a representation retains temporal distinctions that can be invisible within a cropped view. We vary event position or the surrounding waveform envelope while holding a local segment unchanged, then ask whether the original frozen probes recover the corresponding attribute changes from complete-window representations. This directly examines the temporal-information mechanism studied in Section~\ref{view-construction-trades-temporal-context-against-rhythmic-detail}.

\textbf{Controlled waveform pairs.} The held-out event-translation and shared-crop panels each contain 512 latent groups with two members; all members of a group remain in the same data partition. Translating an event by 15 seconds makes the first 15 seconds of one clean waveform identical to the last 15 seconds of its counterpart. Shared-crop pairs preserve the clean 10--20-second segment and reorder only its surroundings. This 10-second construction tests local-view ambiguity; contrastive pretraining itself uses random 15-second crops.

\textbf{Fixed representations and probes.} Encoders receive complete windows without an original-crop offset. We retain the original frozen Ridge probes and evaluate observation noise 0 and 0.3; the main figure reports 0.3. Pair cosine is the cosine similarity between the two members' representations. Same-frequency controls pair each original with a different latent group, using the same assignment for every model and readout.

To test whether the changed attributes remain readable, we compare the frozen probe's predicted change with the true target change. For pair \(i\), superscripts \(0\) and \(1\) identify its two members; \(y_i\) and \(\widehat y_i\) denote the true and predicted targets. Let \(s_y\) be the target standard deviation fitted on the probe-training split. Define \(\Delta y_i=(y_i^{(1)}-y_i^{(0)})/s_y\) and \(\Delta\widehat y_i=(\widehat y_i^{(1)}-\widehat y_i^{(0)})/s_y\), with coordinate-wise division for vector targets. The change error is

\[
\Delta\mathrm{NMSE}
=\frac{\sum_i\|\Delta\widehat y_i-\Delta y_i\|_2^2}
{\sum_i\|\Delta y_i\|_2^2}.
\]

An unchanged prediction gives error 1. Event-location analysis uses event center; shared-crop analysis uses the 60-bin relative log-power profile. The similarity and change-error summaries below average all 512 pairs within a trained model, then report means and sample SDs across three training repetitions. Together, feature similarity and change error describe sensitivity to these controlled waveform changes.

\Needspace{17\baselineskip}

\phantomsection\label{tab:event-translation}
{\color{black}\textbf{Table D2. Event translation diagnostics.}} Complete-window features and original frozen probes at observation noise 0.3. Cosine columns use CLS; change errors show both readouts. Means ± sample SD across three training repetitions.

{
\begin{longtable}[]{@{}
  >{\raggedright\arraybackslash}p{(\linewidth - 1pt - 8\tabcolsep) * \real{0.1579}}
  >{\raggedleft\arraybackslash}p{(\linewidth - 1pt - 8\tabcolsep) * \real{0.2105}}
  >{\raggedleft\arraybackslash}p{(\linewidth - 1pt - 8\tabcolsep) * \real{0.2105}}
  >{\raggedleft\arraybackslash}p{(\linewidth - 1pt - 8\tabcolsep) * \real{0.2105}}
  >{\raggedleft\arraybackslash}p{(\linewidth - 1pt - 8\tabcolsep) * \real{0.2105}}@{}}
\toprule\noalign{}
\rowcolor{PulsePale}
\begin{minipage}[b]{\linewidth}\raggedright
Objective
\end{minipage} & \begin{minipage}[b]{\linewidth}\raggedleft
Pair cosine
\end{minipage} & \begin{minipage}[b]{\linewidth}\raggedleft
Control cosine
\end{minipage} & \begin{minipage}[b]{\linewidth}\raggedleft
CLS change error
\end{minipage} & \begin{minipage}[b]{\linewidth}\raggedleft
Patch-mean change error
\end{minipage} \\
\midrule\noalign{}
\endhead
\bottomrule\noalign{}
\endlastfoot
InfoNCE full & 0.6394 ± 0.0253 & 0.0690 ± 0.0073 & 0.4806 ± 0.0371 & 0.2578 ± 0.0487 \\
InfoNCE crop & 0.9970 ± 0.0001 & 0.1610 ± 0.0192 & 1.0000 ± 0.0000 & 0.8845 ± 0.1166 \\
LeJEPA full & 0.8282 ± 0.0970 & 0.1042 ± 0.0172 & 0.7553 ± 0.1506 & 0.9823 ± 0.0307 \\
LeJEPA crop & 0.9942 ± 0.0003 & 0.0580 ± 0.0049 & 0.9998 ± 0.0003 & 0.9991 ± 0.0015 \\
RFP & 0.9603 ± 0.0011 & 0.8657 ± 0.0085 & 0.1539 ± 0.0065 & 0.2526 ± 0.0404 \\
\end{longtable}
}

\Needspace{17\baselineskip}

\phantomsection\label{tab:shared-crop}
{\color{black}\textbf{Table D3. Shared crop diagnostics.}} Complete-window features and original frozen probes at observation noise 0.3. Cosine columns use CLS; change errors show both readouts. Means ± sample SD across three training repetitions.

{
\begin{longtable}[]{@{}
  >{\raggedright\arraybackslash}p{(\linewidth - 1pt - 8\tabcolsep) * \real{0.1579}}
  >{\raggedleft\arraybackslash}p{(\linewidth - 1pt - 8\tabcolsep) * \real{0.2105}}
  >{\raggedleft\arraybackslash}p{(\linewidth - 1pt - 8\tabcolsep) * \real{0.2105}}
  >{\raggedleft\arraybackslash}p{(\linewidth - 1pt - 8\tabcolsep) * \real{0.2105}}
  >{\raggedleft\arraybackslash}p{(\linewidth - 1pt - 8\tabcolsep) * \real{0.2105}}@{}}
\toprule\noalign{}
\rowcolor{PulsePale}
\begin{minipage}[b]{\linewidth}\raggedright
Objective
\end{minipage} & \begin{minipage}[b]{\linewidth}\raggedleft
Pair cosine
\end{minipage} & \begin{minipage}[b]{\linewidth}\raggedleft
Control cosine
\end{minipage} & \begin{minipage}[b]{\linewidth}\raggedleft
CLS change error
\end{minipage} & \begin{minipage}[b]{\linewidth}\raggedleft
Patch-mean change error
\end{minipage} \\
\midrule\noalign{}
\endhead
\bottomrule\noalign{}
\endlastfoot
InfoNCE full & 0.4533 ± 0.0112 & 0.0601 ± 0.0027 & 0.5289 ± 0.0239 & 0.4231 ± 0.0197 \\
InfoNCE crop & 0.9984 ± 0.0002 & 0.1772 ± 0.0315 & 1.0000 ± 0.0000 & 0.6680 ± 0.2312 \\
LeJEPA full & 0.5152 ± 0.1940 & 0.0777 ± 0.0097 & 0.7416 ± 0.2096 & 1.0000 ± 0.0000 \\
LeJEPA crop & 0.9940 ± 0.0003 & 0.0602 ± 0.0087 & 1.0000 ± 0.0000 & 1.0000 ± 0.0000 \\
RFP & 0.9462 ± 0.0019 & 0.8246 ± 0.0073 & 0.1744 ± 0.0054 & 0.3696 ± 0.0122 \\
\end{longtable}
}
The cropped-alignment models give nearly identical CLS features for paired waveforms, yet their change errors remain near the no-change reference of 1. RFP reduces CLS change error to 0.1539 for event translation and 0.1744 for shared-crop pairs. Feature similarity alone therefore does not establish that temporal information has been lost: the probe must also test whether the changed attribute remains recoverable. These diagnostics explain the temporal-context side of the context--rhythm tradeoff in Section~\ref{view-construction-trades-temporal-context-against-rhythmic-detail}.

\subsection{Online random-teacher prediction}\label{b.4-random-teacher-prediction-metrics}

The online analysis asks whether the student becomes better at predicting freshly sampled signal functions. For each eligible training update, ridge regression fits the mixed 576-dimensional teacher targets on the support set and is evaluated on the query set. The reference predictor assigns the support-target mean to every query example. Skill is one minus the ratio of query MSE to reference MSE, computed within each update before averaging.

Successive 5,000-update bins weight eligible intervals by their number of updates. Updates without a waveform target are excluded. The trajectory in \hyperref[fig:10]{Figure~10} shows increasing query skill and decreasing query MSE against an approximately stable support-mean reference error.

\subsection{Independent random-teacher evaluation}\label{b.5-independent-generated-task-protocol}

The evaluation in Section 5.3.2 tests whether learning across changing teachers produces frozen representations that support new teacher realizations. All stages use the same signal panel and newly sampled teachers.

The BIDMC dataset panel contains 320 distinct windows from 40 patients. Support and query share no patient within an episode. The four patient partitions reuse the same fixed windows, and each is evaluated in both directions. A common patch mask retains 50\% of the input patches at every stage. Early, intermediate and late training stages are evaluated on the same diagnostic panel. All evaluation teachers are newly sampled independently of the training teachers. Each teacher uses the original raw/time-frequency/autocorrelation construction followed by the channel-conditioned residual and random 576-dimensional mixing projection.

For each direction, feature and target means and population variances are fitted on the support only, with variance floor \(10^{-8}\). The dual ridge solve adds \(n_S\lambda I\), with \(\lambda=1\). All task metrics are computed in float64 after frozen float32 feature extraction. We report per-direction query MSE, reference MSE and skill; we first average over the eight fixed directions, then over 128 teachers. The teacher SD and its standard error condition on the fixed patients, masks and partitions. The episodes reuse patients across partitions and are not independent patient replications.

\Needspace{12\baselineskip}

\phantomsection\label{tab:independent-teacher}
{\color{black}\textbf{Table D4. Independent random-teacher evaluation.}} The same 128 generated teachers are used at every stage.

{
\begin{longtable}[]{@{}
  >{\raggedleft\arraybackslash}p{(\linewidth - 1pt - 10\tabcolsep) * \real{0.1667}}
  >{\raggedleft\arraybackslash}p{(\linewidth - 1pt - 10\tabcolsep) * \real{0.1667}}
  >{\raggedleft\arraybackslash}p{(\linewidth - 1pt - 10\tabcolsep) * \real{0.1667}}
  >{\raggedleft\arraybackslash}p{(\linewidth - 1pt - 10\tabcolsep) * \real{0.1667}}
  >{\raggedleft\arraybackslash}p{(\linewidth - 1pt - 10\tabcolsep) * \real{0.1667}}
  >{\raggedleft\arraybackslash}p{(\linewidth - 1pt - 10\tabcolsep) * \real{0.1667}}@{}}
\toprule\noalign{}
\rowcolor{PulsePale}
\begin{minipage}[b]{\linewidth}\raggedleft
Training stage
\end{minipage} & \begin{minipage}[b]{\linewidth}\raggedleft
Mean query MSE
\end{minipage} & \begin{minipage}[b]{\linewidth}\raggedleft
Mean reference MSE
\end{minipage} & \begin{minipage}[b]{\linewidth}\raggedleft
Mean skill
\end{minipage} & \begin{minipage}[b]{\linewidth}\raggedleft
Teacher SD
\end{minipage} & \begin{minipage}[b]{\linewidth}\raggedleft
Teacher-mean SE
\end{minipage} \\
\midrule\noalign{}
\endhead
\bottomrule\noalign{}
\endlastfoot
Early & 0.7276 & 1.0853 & 0.3270 & 0.0934 & 0.0083 \\
Intermediate & 0.4741 & 1.0853 & 0.5616 & 0.0945 & 0.0083 \\
Late & 0.4647 & 1.0853 & 0.5697 & 0.0980 & 0.0087 \\
\end{longtable}
}

\section{Controlled cross-modal bridge experiment}\label{app:bridge}

This section supports the bridge experiment in Section 5.4. We define the four training groups and their controlled differences, specify the common respiration-only evaluation, and report the complete scores and paired contrasts.

\subsection{Pairing structure and controlled training groups}\label{e.2.1-relationships-populations-and-matched-exposure}

We train four shared-encoder models using ECG--thoracic-effort and EEG--ECG pairs. Each physical signal is encoded separately using shared parameters and its modality condition. The study uses no direct EEG--thoracic-effort training pair and no joint three-signal encoder input.

The pretraining data use participant-grouped train/validation/test splits in an approximately \(8:1:1\) ratio. Sleep-stage labels enter only the frozen downstream probe. Inputs are 30-second, 64-Hz windows (1,920 samples), normalized over observed samples with a standard-deviation floor of \(10^{-6}\); missing samples are zeroed and their masks retained.

{
\begin{longtable}[]{@{}lrrrr@{}}
\toprule\noalign{}
\rowcolor{PulsePale}
Training allocation or policy & A & B & C & D \\
\midrule\noalign{}
\endhead
\bottomrule\noalign{}
\endlastfoot
ECG--thoracic-effort updates & 50,000 & 25,000 & 25,000 & 25,000 \\
EEG--ECG updates & 0 & 25,000 & 25,000 & 25,000 \\
Auxiliary alignment & Absent & Correct & Disabled & Shuffled \\
ECG--respiration alignment & Correct & Correct & Correct & Correct \\
Total pair updates & 50,000 & 50,000 & 50,000 & 50,000 \\
\end{longtable}
}

Each update contains 1,024 synchronized pairs. B, C and D receive the same patients, windows, observation masks, artificial masks and teacher functions; their auxiliary alignment policy is the controlled difference.

\textbf{Matched objectives and auxiliary correspondence.} All four models share the encoder and common objectives: projected alignment, SIGReg, RFP and masked reconstruction. Only the auxiliary alignment policy varies among B, C and D; C sets its EEG--ECG alignment coefficient to zero without rescaling the remaining terms.

D changes only the EEG--ECG alignment loss: it cyclically permutes the ECG partners within each batch so that every auxiliary EEG window is paired with a different patient. Original waveform inputs, MAE targets, RFP targets and SIGReg inputs are retained; the ECG--respiration steps remain correctly aligned. Thus B−C compares adding the auxiliary projected-alignment term at fixed input exposure, and B−D compares correct and incorrect auxiliary correspondence.

\subsection{Respiration-only evaluation and statistical analysis}\label{e.2.2-objectives-frozen-evaluation-and-results}

\textbf{Identical respiration-only panels and probes.} Fit/validation/test contain 256/128/642 patients, with 32 windows each, totaling 8,192/4,096/20,544 windows. All groups use the same patients, recordings, window positions, labels and sample ordering. No test patient is excluded by panel sampling. Complete observed thoracic-effort inputs are encoded without an artificial evaluation mask. Final-layer-normalized CLS and the mean of valid observed patch tokens provide the two frozen readouts.

Every encoder/readout uses the same standardized, class-balanced linear classification probe, fitted only on the probe-training panel. Macro-F1 averages five class F1 values from the confusion matrix pooled across test windows. Both readouts appear in \hyperref[tab:6]{Table~6} and \hyperref[tab:E2]{Table~E2}.

\textbf{Paired uncertainty.} We resample the same 642 test patients 2,000 times, preserving all windows for each sampled patient and using identical patient draws across arms. Each draw recomputes Macro-F1 from pooled confusion matrices. \hyperref[tab:E1]{Table~E1} reports percentile intervals for all six paired differences.

\Needspace{21\baselineskip}
\subsection{Complete results and paired contrasts}\label{app:bridge-results}

The tables report the group-wise staging scores and the paired differences used to assess auxiliary correspondence in Section 5.4.

\Needspace{12\baselineskip}

\phantomsection\label{tab:E1}
{\color{black}\textbf{Table E1. Paired bridge contrasts and patient-bootstrap intervals.}} Differences and 95\% intervals are in pp, calculated from the unrounded scores. Resampling unit: patient, with all sampled windows retained. These intervals describe test-patient sampling for the fixed fitted encoders and probes.

{
\begin{longtable}[]{@{}llrl@{}}
\toprule\noalign{}
\rowcolor{PulsePale}
Contrast & Readout & Difference (pp) & 95\% interval (pp) \\
\midrule\noalign{}
\endhead
\bottomrule\noalign{}
\endlastfoot
B−C & CLS & +4.21 & {[}3.37, 5.04{]} \\
B−C & Patch mean & +0.56 & {[}-0.32, 1.49{]} \\
B−D & CLS & +4.23 & {[}3.42, 5.02{]} \\
B−D & Patch mean & +0.53 & {[}-0.35, 1.42{]} \\
B−A & CLS & +3.94 & {[}3.13, 4.76{]} \\
B−A & Patch mean & +0.77 & {[}-0.05, 1.61{]} \\
\end{longtable}
}

\Needspace{12\baselineskip}

\phantomsection\label{tab:E2}
{\color{black}\textbf{Table E2. Bridge readout comparison.}} Test Macro-F1 (\%). The main text uses patch mean; CLS is secondary. All six paired contrasts and their uncertainty remain in \hyperref[tab:E1]{Table~E1}.

{
\begin{longtable}[]{@{}lrr@{}}
\toprule\noalign{}
\rowcolor{PulsePale}
Arm & CLS (\%) & Patch mean (\%) \\
\midrule\noalign{}
\endhead
\bottomrule\noalign{}
\endlastfoot
A & 39.92 & 43.79 \\
B & 43.86 & 44.56 \\
C & 39.65 & 44.00 \\
D & 39.63 & 44.03 \\
\end{longtable}
}

\clearpage
\section{Downstream evaluation and detailed results}\label{app:downstream}

Shared protocols and complete task results supplement Section 4. Physiological signal transfer is followed by multimodal and single-modality sleep assessment.

\subsection{Shared evaluation protocols}\label{e.1-baselines-adaptation-ranking-and-uncertainty}

\subsubsection{Baselines and adaptation protocols}\label{e.1.1-baselines-and-adaptation-protocols}

\textbf{Pretrained baselines.} \hyperref[tab:4]{Table~4} groups the evaluated foundation models by signal domain. Each model uses its published sampling rate, channel requirements and representation interface; the task-specific tables below identify the models compared for each input and endpoint.

\textbf{Readout and supervision.} The primary representation evaluation uses frozen encoders and linear probes, with patch mean for window-level tasks and CLS results reported separately, as defined in Section 3.2. Sleep sequence tasks use task-specific temporal heads, while continuous EMG decoding retains patch features. Feature scaling, hyperparameter selection, early stopping, label budgets and temporal-head settings use training and validation data. Repeated fits report means and sample SDs over head/split repeats; fixed overnight fits and DB8 trajectory fits are reported as single values.

\textbf{Input access.} Each comparison fixes physical observation duration, available signals/leads, and label partitions. 

\textbf{Input duration and temporal resolution.} The backbone accepts variable token counts. For the encoder, the collator pads to the batch maximum rounded to a 32-sample patch boundary and retains separate sample-present and observed masks. Each downstream adapter specifies its window duration, fixed or batch-padded tensor policy, and valid-patch pooling rule. Global CLS and patch mean are extracted before projection; temporal tasks retain their declared sequence-to-decoder alignment.

The encoder is pretrained on a 128-Hz waveform grid with antialiasing and ordered groups of 32 samples. The DB8 evaluation applies this model directly to the published 2,000-Hz arrays without temporal resampling. Native-rate baselines retain their own input bandwidth, lead availability and normalization. 

\subsubsection{Rank definition and aggregation}\label{e.1.2-rank-definition-and-aggregation}

Each ranking panel fixes a signal domain, task set and baseline set shared by both readouts. CLS and patch mean are ranked separately; trajectory tasks use patch-sequence features and their primary metric.

For modality \(m\) and protocol \(p\), fix a common model set \(\mathcal M_{m,p}\), a dataset set \(\mathcal D_{m,p}\), and primary tasks \(\mathcal T_d\) in each dataset. Scores are oriented so that a better score receives a lower rank, with average ranks assigned to exact ties. The dataset-balanced mean rank is:

\[
\overline r_{m,p}(a)
=
\frac1{|\mathcal D_{m,p}|}
\sum_{d\in\mathcal D_{m,p}}
\frac1{|\mathcal T_d|}
\sum_{t\in\mathcal T_d}
r_{m,p,d,t}(a).
\]

Here, \(a\in\mathcal M_{m,p}\) denotes the evaluated model, and \(r_{m,p,d,t}(a)\) is its rank on primary task \(t\) in dataset \(d\). The inner average weights tasks equally within a dataset; the outer average weights datasets equally.

Repeated fits or folds are summarized before ranking, and each label contributes one primary metric. Predefined groups consolidate highly overlapping tasks. Primary panels use common completed task/model matrices; model-specific partial coverage is reported with its denominator.

\subsection{Detailed results for physiological signal transfer}\label{app:complete}

Results are organized by PPG, ECG, EEG and EMG, followed by dataset-balanced rank summaries.

\begingroup\footnotesize

\subsubsection{PPG}

Participants are grouped within each data partition. The BIDMC dataset uses 10-s windows for heart-rate estimation and 30-s windows for respiratory-rate and oxygen-saturation estimation. The PPG-DaLiA dataset uses 8-s wrist blood-volume-pulse (BVP) windows, and the WESAD dataset uses 10-s BVP windows. Models retain their native sampling rates and the stated channel mappings.

\begin{longtable}{@{}p{1.15cm}p{8.0cm}p{4.5cm}@{}}\toprule\rowcolor{PulsePale}ID & Dataset / endpoint & Metric\\\midrule\endhead

P01a & BIDMC / PPG HR & MAE (bpm) $\downarrow$\\

P01b & BIDMC / PPG monitor RR & MAE (breaths/min) $\downarrow$\\

P01c & BIDMC / PPG oxygen saturation & MAE (pp) $\downarrow$\\

P01d & PPG-DaLiA / Wrist PPG HR & MAE (bpm) $\downarrow$\\

P01e & WESAD / PPG four-state classification & Macro-F1 $\uparrow$\\

P01f & BUT-PPG / Signal quality & Macro-F1 $\uparrow$\\

P01g & BUT-PPG / SBP & MAE (mmHg) $\downarrow$\\

\bottomrule\end{longtable}

\begingroup\footnotesize\setlength{\tabcolsep}{4pt}\renewcommand{\arraystretch}{1.15}

\begin{longtable}{@{}p{5.1cm}rrrr@{}}\toprule

\rowcolor{PulsePale}Method & P01a & P01b & P01c & P01d\\\midrule\endhead

\rowcolor{PulsePale}
LingSante-Pulse CLS & 3.0486\,$\pm$\,0.3724 & 2.4385\,$\pm$\,0.0477 & 2.6088\,$\pm$\,0.0395 & 6.1251\,$\pm$\,0.1715\\
\rowcolor{PulsePale}
LingSante-Pulse patch mean & 2.2108\,$\pm$\,0.3415 & 2.3838\,$\pm$\,0.0427 & 2.6592\,$\pm$\,0.0789 & 5.6843\,$\pm$\,0.1351\\
AnyPPG & 2.2517\,$\pm$\,0.3043 & 2.5248\,$\pm$\,0.1100 & 2.6609\,$\pm$\,0.0431 & 7.5924\,$\pm$\,0.1711\\
PaPaGei-P & 5.1165\,$\pm$\,0.3611 & 2.4778\,$\pm$\,0.0429 & 2.4919\,$\pm$\,0.1206 & 12.0720\,$\pm$\,0.2994\\
PaPaGei-S & 5.8433\,$\pm$\,0.2263 & 2.4870\,$\pm$\,0.0194 & 2.5749\,$\pm$\,0.0864 & 14.2051\,$\pm$\,0.3289\\
PulsePPG & 3.2198\,$\pm$\,0.1380 & 2.4346\,$\pm$\,0.0303 & 2.5246\,$\pm$\,0.0729 & 10.2717\,$\pm$\,0.2440\\
SIGMA-PPG & 4.7019\,$\pm$\,0.3501 & 2.4580\,$\pm$\,0.0190 & 2.6042\,$\pm$\,0.0249 & 11.7553\,$\pm$\,0.2738\\
MOMENT-small & 2.5515\,$\pm$\,0.1873 & 2.4610\,$\pm$\,0.0127 & 2.7921\,$\pm$\,0.1785 & 8.4638\,$\pm$\,0.1861\\
SPOTR-Base & 5.0036\,$\pm$\,0.2783 & 2.3545\,$\pm$\,0.0378 & 2.5144\,$\pm$\,0.0392 & 11.6058\,$\pm$\,0.2730\\
DNA-PPG & 5.4323\,$\pm$\,4.4710 & 3.3503\,$\pm$\,0.9730 & 3.0029\,$\pm$\,0.6176 & 8.1482\,$\pm$\,0.2052\\
\bottomrule\end{longtable}\endgroup

\begingroup\footnotesize\setlength{\tabcolsep}{4pt}\renewcommand{\arraystretch}{1.15}

\begin{longtable}{@{}p{5.1cm}rrr@{}}\toprule

\rowcolor{PulsePale}Method & P01e & P01f & P01g\\\midrule\endhead

\rowcolor{PulsePale}
LingSante-Pulse CLS & 0.5482\,$\pm$\,0.0053 & 0.7584\,$\pm$\,0.0077 & 12.3836\,$\pm$\,0.2745\\
\rowcolor{PulsePale}
LingSante-Pulse patch mean & 0.5887\,$\pm$\,0.0067 & 0.7735\,$\pm$\,0.0009 & 12.2939\,$\pm$\,0.3116\\
AnyPPG & 0.5321\,$\pm$\,0.0043 & 0.6506\,$\pm$\,0.0119 & 12.5955\,$\pm$\,0.3250\\
PaPaGei-P & 0.3969\,$\pm$\,0.0097 & 0.5369\,$\pm$\,0.0088 & 12.4237\,$\pm$\,0.2381\\
PaPaGei-S & 0.3809\,$\pm$\,0.0092 & 0.6548\,$\pm$\,0.0029 & 12.1721\,$\pm$\,0.1601\\
PulsePPG & 0.4637\,$\pm$\,0.0031 & 0.7385\,$\pm$\,0.0084 & 12.5411\,$\pm$\,0.3916\\
SIGMA-PPG & 0.4526\,$\pm$\,0.0064 & 0.5652\,$\pm$\,0.0051 & 12.4238\,$\pm$\,0.2462\\
MOMENT-small & 0.4904\,$\pm$\,0.0093 & 0.7241\,$\pm$\,0.0043 & 12.3662\,$\pm$\,0.2335\\
SPOTR-Base & 0.4244\,$\pm$\,0.0047 & 0.6179\,$\pm$\,0.0056 & 12.4192\,$\pm$\,0.2582\\
DNA-PPG & 0.4941\,$\pm$\,0.0082 & 0.7303\,$\pm$\,0.0092 & 12.7695\,$\pm$\,0.3741\\
\bottomrule\end{longtable}\endgroup

\clearpage

The WildPPG dataset uses 93,961 valid 8-s windows from 16 participants under fixed repeated subject splits. For the WESAD dataset, valence and arousal targets are the self-reported ratings for each experimental condition, assigned to the PPG windows from that condition.

\begin{longtable}{@{}p{1.15cm}p{8.0cm}p{4.5cm}@{}}\toprule\rowcolor{PulsePale}ID & Dataset / endpoint & Metric\\\midrule\endhead

P02a & PPG-DaLiA / Nine-class activity & Macro-F1 $\uparrow$\\

P02b & WESAD / Valence & MAE (SAM units) $\downarrow$\\

P02c & WESAD / Arousal & MAE (SAM units) $\downarrow$\\

P02d & WildPPG / Wrist PPG HR & MAE (bpm) $\downarrow$\\

P02e & BUT-PPG / Glucose, quality filtered & MAE (label units) $\downarrow$\\

P02f & BUT-PPG / SpO2, quality filtered & MAE (pp) $\downarrow$\\

\bottomrule\end{longtable}

\begingroup\footnotesize\setlength{\tabcolsep}{4pt}\renewcommand{\arraystretch}{1.15}

\begin{longtable}{@{}p{5.1cm}rrrr@{}}\toprule

\rowcolor{PulsePale}Method & P02a & P02b & P02c & P02d\\\midrule\endhead

\rowcolor{PulsePale}
LingSante-Pulse CLS & 0.3577\,$\pm$\,0.0068 & 1.0394\,$\pm$\,0.0344 & 1.3698\,$\pm$\,0.0785 & 10.6295\,$\pm$\,0.1745\\
\rowcolor{PulsePale}
LingSante-Pulse patch mean & 0.3875\,$\pm$\,0.0052 & 0.9786\,$\pm$\,0.0318 & 1.3157\,$\pm$\,0.0279 & 10.2875\,$\pm$\,0.1771\\
AnyPPG & 0.3357\,$\pm$\,0.0063 & 0.9482\,$\pm$\,0.0129 & 1.2647\,$\pm$\,0.0312 & 10.6786\,$\pm$\,0.0651\\
PaPaGei-P & 0.2605\,$\pm$\,0.0028 & 1.1329\,$\pm$\,0.0479 & 1.5795\,$\pm$\,0.0339 & 12.5152\,$\pm$\,0.1067\\
PaPaGei-S & 0.2190\,$\pm$\,0.0016 & 1.1985\,$\pm$\,0.0472 & 1.4928\,$\pm$\,0.0513 & 13.2045\,$\pm$\,0.1661\\
PulsePPG & 0.2863\,$\pm$\,0.0034 & 1.0741\,$\pm$\,0.0181 & 1.4579\,$\pm$\,0.0485 & 12.3005\,$\pm$\,0.0736\\
SIGMA-PPG & 0.2667\,$\pm$\,0.0047 & 1.1133\,$\pm$\,0.0319 & 1.5461\,$\pm$\,0.0842 & 13.3572\,$\pm$\,0.0453\\
MOMENT-small & 0.3365\,$\pm$\,0.0057 & 1.0755\,$\pm$\,0.0759 & 1.5466\,$\pm$\,0.0523 & 11.7683\,$\pm$\,0.1477\\
SPOTR-Base & 0.2817\,$\pm$\,0.0037 & 1.2094\,$\pm$\,0.0262 & 1.7282\,$\pm$\,0.0616 & 13.5169\,$\pm$\,0.1377\\
DNA-PPG & 0.3161\,$\pm$\,0.0056 & 1.6135\,$\pm$\,0.5777 & 2.1321\,$\pm$\,0.4021 & 11.4498\,$\pm$\,0.2224\\
\bottomrule\end{longtable}\endgroup

\begingroup\footnotesize\setlength{\tabcolsep}{4pt}\renewcommand{\arraystretch}{1.15}

\begin{longtable}{@{}p{5.1cm}rr@{}}\toprule

\rowcolor{PulsePale}Method & P02e & P02f\\\midrule\endhead

\rowcolor{PulsePale}
LingSante-Pulse CLS & 1.1131\,$\pm$\,0.0276 & 1.3660\,$\pm$\,0.0096\\
\rowcolor{PulsePale}
LingSante-Pulse patch mean & 1.1356\,$\pm$\,0.0390 & 1.3717\,$\pm$\,0.0054\\
AnyPPG & 1.1589\,$\pm$\,0.0264 & 1.3297\,$\pm$\,0.0108\\
PaPaGei-P & 1.1809\,$\pm$\,0.0325 & 1.3361\,$\pm$\,0.0093\\
PaPaGei-S & 1.1360\,$\pm$\,0.0303 & 1.3341\,$\pm$\,0.0149\\
PulsePPG & 1.1495\,$\pm$\,0.0483 & 1.3422\,$\pm$\,0.0128\\
SIGMA-PPG & 1.1705\,$\pm$\,0.0230 & 1.3373\,$\pm$\,0.0186\\
MOMENT-small & 1.1327\,$\pm$\,0.0122 & 1.3547\,$\pm$\,0.0140\\
SPOTR-Base & 1.1570\,$\pm$\,0.0339 & 1.3316\,$\pm$\,0.0217\\
DNA-PPG & 1.1001\,$\pm$\,0.0396 & 1.3756\,$\pm$\,0.0278\\
\bottomrule\end{longtable}\endgroup

\clearpage

\paragraph{PPG-BP dataset.}

The PPG-BP dataset is evaluated on heart rate, blood pressure, age and BMI estimation, together with diabetes and hypertension classification. Three 2.1-s recordings are aggregated per participant before scoring. Source filtering and participant splits remain fixed. Label units are retained; raw errors across targets are not averaged.

\begin{longtable}{@{}p{1.15cm}p{8.0cm}p{4.5cm}@{}}\toprule\rowcolor{PulsePale}ID & Dataset / endpoint & Metric\\\midrule\endhead

P03a & PPG-BP / HR & MAE (bpm) $\downarrow$\\

P03b & PPG-BP / SBP & MAE (mmHg) $\downarrow$\\

P03c & PPG-BP / DBP & MAE (mmHg) $\downarrow$\\

P03d & PPG-BP / Age & MAE (years) $\downarrow$\\

P03e & PPG-BP / BMI & MAE (kg/m2) $\downarrow$\\

P03f & PPG-BP / Diabetes & Macro-F1 $\uparrow$\\

P03g & PPG-BP / Four-class hypertension & Macro-F1 $\uparrow$\\

\bottomrule\end{longtable}

\begingroup\footnotesize\setlength{\tabcolsep}{4pt}\renewcommand{\arraystretch}{1.15}

\begin{longtable}{@{}p{5.1cm}rrrr@{}}\toprule

\rowcolor{PulsePale}Method & P03a & P03b & P03c & P03d\\\midrule\endhead

\rowcolor{PulsePale}
LingSante-Pulse CLS & 3.0793\,$\pm$\,0.0395 & 13.7687\,$\pm$\,0.2988 & 8.5075\,$\pm$\,0.1088 & 10.6939\,$\pm$\,0.2700\\
\rowcolor{PulsePale}
LingSante-Pulse patch mean & 3.1644\,$\pm$\,0.0218 & 13.8070\,$\pm$\,0.3055 & 8.0675\,$\pm$\,0.1583 & 10.4428\,$\pm$\,0.1610\\
AnyPPG & 3.2677\,$\pm$\,0.0703 & 13.3658\,$\pm$\,0.2755 & 8.2563\,$\pm$\,0.0633 & 10.6827\,$\pm$\,0.0809\\
PaPaGei-P & 5.2451\,$\pm$\,0.0620 & 14.5196\,$\pm$\,0.3260 & 8.6112\,$\pm$\,0.1294 & 11.4031\,$\pm$\,0.1386\\
PaPaGei-S & 4.9161\,$\pm$\,0.1470 & 14.9826\,$\pm$\,0.0697 & 8.7700\,$\pm$\,0.0848 & 11.3974\,$\pm$\,0.2744\\
PulsePPG & 3.7090\,$\pm$\,0.1066 & 13.5247\,$\pm$\,0.2685 & 8.5242\,$\pm$\,0.2546 & 10.3728\,$\pm$\,0.2454\\
SIGMA-PPG & 4.2509\,$\pm$\,0.1084 & 15.0866\,$\pm$\,0.2717 & 8.6465\,$\pm$\,0.0405 & 11.5741\,$\pm$\,0.1985\\
MOMENT-small & 3.5423\,$\pm$\,0.0373 & 13.0429\,$\pm$\,0.2189 & 8.4074\,$\pm$\,0.0268 & 10.3813\,$\pm$\,0.2299\\
\bottomrule\end{longtable}\endgroup

\begingroup\footnotesize\setlength{\tabcolsep}{4pt}\renewcommand{\arraystretch}{1.15}

\begin{longtable}{@{}p{5.1cm}rrr@{}}\toprule

\rowcolor{PulsePale}Method & P03e & P03f & P03g\\\midrule\endhead

\rowcolor{PulsePale}
LingSante-Pulse CLS & 3.0547\,$\pm$\,0.0323 & 0.5012\,$\pm$\,0.0252 & 0.3991\,$\pm$\,0.0215\\
\rowcolor{PulsePale}
LingSante-Pulse patch mean & 3.0641\,$\pm$\,0.0346 & 0.5150\,$\pm$\,0.0214 & 0.3890\,$\pm$\,0.0270\\
AnyPPG & 2.9857\,$\pm$\,0.0374 & 0.5703\,$\pm$\,0.0201 & 0.4342\,$\pm$\,0.0115\\
PaPaGei-P & 2.9902\,$\pm$\,0.0197 & 0.4744\,$\pm$\,0.0398 & 0.3732\,$\pm$\,0.0203\\
PaPaGei-S & 3.0051\,$\pm$\,0.0534 & 0.5283\,$\pm$\,0.0313 & 0.3524\,$\pm$\,0.0260\\
PulsePPG & 2.9956\,$\pm$\,0.0234 & 0.5347\,$\pm$\,0.0311 & 0.3934\,$\pm$\,0.0250\\
SIGMA-PPG & 3.0450\,$\pm$\,0.0413 & 0.5602\,$\pm$\,0.0114 & 0.3679\,$\pm$\,0.0137\\
MOMENT-small & 3.0108\,$\pm$\,0.0456 & 0.5279\,$\pm$\,0.0314 & 0.3990\,$\pm$\,0.0121\\
\bottomrule\end{longtable}\endgroup

\clearpage

The following binary tasks evaluate normal versus abnormal blood pressure, cerebral infarction and cerebrovascular disease. They are reported separately and excluded from the PPG rank summary in Section~\ref{app:ranks}.

\begin{longtable}{@{}p{1.15cm}p{8.0cm}p{4.5cm}@{}}\toprule\rowcolor{PulsePale}ID & Dataset / endpoint & Metric\\\midrule\endhead

P04a & PPG-BP / Normal / non-normal BP & Macro-F1 $\uparrow$\\

P04b & PPG-BP / Cerebral infarction & Macro-F1 $\uparrow$\\

P04c & PPG-BP / Cerebrovascular disease & Macro-F1 $\uparrow$\\

\bottomrule\end{longtable}

\begingroup\footnotesize\setlength{\tabcolsep}{4pt}\renewcommand{\arraystretch}{1.15}

\begin{longtable}{@{}p{5.1cm}rrr@{}}\toprule

\rowcolor{PulsePale}Method & P04a & P04b & P04c\\\midrule\endhead

\rowcolor{PulsePale}
LingSante-Pulse CLS & 0.6772\,$\pm$\,0.0309 & 0.4654\,$\pm$\,0.0196 & 0.5222\,$\pm$\,0.0328\\
\rowcolor{PulsePale}
LingSante-Pulse patch mean & 0.7113\,$\pm$\,0.0084 & 0.4861\,$\pm$\,0.0076 & 0.5230\,$\pm$\,0.0316\\
AnyPPG & 0.7092\,$\pm$\,0.0234 & 0.4992\,$\pm$\,0.0300 & 0.5024\,$\pm$\,0.0258\\
PaPaGei-P & 0.6495\,$\pm$\,0.0203 & 0.5065\,$\pm$\,0.0197 & 0.5222\,$\pm$\,0.0209\\
PaPaGei-S & 0.6481\,$\pm$\,0.0319 & 0.4977\,$\pm$\,0.0212 & 0.5211\,$\pm$\,0.0284\\
PulsePPG & 0.6787\,$\pm$\,0.0281 & 0.5349\,$\pm$\,0.0257 & 0.5300\,$\pm$\,0.0237\\
SIGMA-PPG & 0.6170\,$\pm$\,0.0255 & 0.5138\,$\pm$\,0.0341 & 0.5196\,$\pm$\,0.0249\\
MOMENT-small & 0.6779\,$\pm$\,0.0046 & 0.4942\,$\pm$\,0.0204 & 0.4916\,$\pm$\,0.0365\\
\bottomrule\end{longtable}\endgroup

\paragraph{Symptom screening, emotion and person identification.}

The ECSMP dataset evaluates screening based on SDS scores; the endpoint is not a clinical diagnosis. The 35-class RealWorldPPG \citep{ref64} task is closed-set identity recognition with record-separated splits and shared identities; it is excluded from unseen-subject aggregation.

\begin{longtable}{@{}p{1.15cm}p{8.0cm}p{4.5cm}@{}}\toprule\rowcolor{PulsePale}ID & Dataset / endpoint & Metric\\\midrule\endhead

P05a & ECSMP / SDS screening & Macro-F1 $\uparrow$\\

P05b & ECSMP / Emotion & Macro-F1 $\uparrow$\\

P05c & RealWorldPPG / Closed-set identity & Macro-F1 $\uparrow$\\

\bottomrule\end{longtable}

\begingroup\footnotesize\setlength{\tabcolsep}{4pt}\renewcommand{\arraystretch}{1.15}

\begin{longtable}{@{}p{5.1cm}rrr@{}}\toprule

\rowcolor{PulsePale}Method & P05a & P05b & P05c\\\midrule\endhead

\rowcolor{PulsePale}
LingSante-Pulse CLS & 0.5167\,$\pm$\,0.0349 & 0.0601\,$\pm$\,0.0175 & 0.9531\,$\pm$\,0.0023\\
\rowcolor{PulsePale}
LingSante-Pulse patch mean & 0.4925\,$\pm$\,0.0295 & 0.0631\,$\pm$\,0.0183 & 0.9610\,$\pm$\,0.0012\\
AnyPPG & 0.4469\,$\pm$\,0.0523 & 0.1003\,$\pm$\,0.0047 & 0.8408\,$\pm$\,0.0036\\
PaPaGei-P & 0.4306\,$\pm$\,0.0180 & 0.0795\,$\pm$\,0.0048 & 0.4117\,$\pm$\,0.0106\\
PaPaGei-S & 0.4341\,$\pm$\,0.0323 & 0.0625\,$\pm$\,0.0108 & 0.4168\,$\pm$\,0.0131\\
PulsePPG & 0.3609\,$\pm$\,0.0243 & 0.1015\,$\pm$\,0.0098 & 0.9153\,$\pm$\,0.0029\\
SIGMA-PPG & 0.3794\,$\pm$\,0.0292 & 0.0884\,$\pm$\,0.0192 & 0.4743\,$\pm$\,0.0061\\
MOMENT-small & 0.4284\,$\pm$\,0.0357 & 0.0791\,$\pm$\,0.0105 & 0.9654\,$\pm$\,0.0014\\
\bottomrule\end{longtable}\endgroup

\clearpage

\subsubsection{ECG}

The MIT-BIH dataset uses the 1-s, four-class panel with 43 patient groups (records 201/202 grouped). The PTB-XL dataset uses official folds 1--8/9/10 and the fixed five head fits. The European ST-T dataset recordings are grouped by patient. The CPSC2018 dataset uses the stated lead mapping and six-second windows.

\begin{longtable}{@{}p{1.15cm}p{8.0cm}p{4.5cm}@{}}\toprule\rowcolor{PulsePale}ID & Dataset / endpoint & Metric\\\midrule\endhead

E01a & MIT-BIH / Four-class beat classification & Macro-F1 $\uparrow$\\

E01b & PTB-XL / Diagnostic superclasses & Macro-F1 $\uparrow$\\

E01c & PTB-XL / Diagnostic subclasses & Macro-F1 $\uparrow$\\

E01d & PTB-XL / Rhythm labels & Macro-F1 $\uparrow$\\

E01e & CPSC2018 / Nine-label ECG classification & Macro-F1 $\uparrow$\\

E01f & European ST-T / ST/T presence & Macro-F1 $\uparrow$\\

\bottomrule\end{longtable}

\begingroup\footnotesize\setlength{\tabcolsep}{4pt}\renewcommand{\arraystretch}{1.15}

\begin{longtable}{@{}p{5.1cm}rrrr@{}}\toprule

\rowcolor{PulsePale}Method & E01a & E01b & E01c & E01d\\\midrule\endhead

\rowcolor{PulsePale}
LingSante-Pulse CLS & 0.4540\,$\pm$\,0.0186 & 0.6342\,$\pm$\,0.0000 & 0.3420\,$\pm$\,0.0002 & 0.5231\,$\pm$\,0.0000\\
\rowcolor{PulsePale}
LingSante-Pulse patch mean & 0.4703\,$\pm$\,0.0211 & 0.6865\,$\pm$\,0.0000 & 0.4113\,$\pm$\,0.0000 & 0.5890\,$\pm$\,0.0000\\
ECG-FM pretrained & 0.3775\,$\pm$\,0.0200 & 0.6729\,$\pm$\,0.0000 & 0.3759\,$\pm$\,0.0000 & 0.4431\,$\pm$\,0.0000\\
NormWear & 0.3376\,$\pm$\,0.0052 & 0.6552\,$\pm$\,0.0000 & 0.3713\,$\pm$\,0.0000 & 0.4193\,$\pm$\,0.0000\\
PanLUNA & 0.4149\,$\pm$\,0.0112 & 0.6826\,$\pm$\,0.0000 & 0.3623\,$\pm$\,0.0000 & 0.2933\,$\pm$\,0.0000\\
MOMENT-small & 0.3834\,$\pm$\,0.0156 & 0.6748\,$\pm$\,0.0000 & 0.4179\,$\pm$\,0.0000 & 0.4764\,$\pm$\,0.0000\\
xECG & 0.4627\,$\pm$\,0.0102 & 0.7268\,$\pm$\,0.0000 & 0.4414\,$\pm$\,0.0000 & 0.5247\,$\pm$\,0.0000\\
ST-MEM & 0.4619\,$\pm$\,0.0097 & 0.7047\,$\pm$\,0.0000 & 0.3898\,$\pm$\,0.0000 & 0.4853\,$\pm$\,0.0000\\
ECG-JEPA Kim multi-block & 0.4365\,$\pm$\,0.0303 & 0.7117\,$\pm$\,0.0000 & 0.4095\,$\pm$\,0.0000 & 0.4871\,$\pm$\,0.0000\\
ECG-CPC HEEDB & 0.3592\,$\pm$\,0.0174 & 0.6658\,$\pm$\,0.0000 & 0.3701\,$\pm$\,0.0000 & 0.4335\,$\pm$\,0.0000\\
SPOTR-Base & 0.3950\,$\pm$\,0.0126 & 0.6726\,$\pm$\,0.0000 & 0.3614\,$\pm$\,0.0000 & 0.2997\,$\pm$\,0.0000\\
HuBERT-ECG SSL Base & 0.3862\,$\pm$\,0.0117 & 0.6602\,$\pm$\,0.0000 & 0.3580\,$\pm$\,0.0000 & 0.3640\,$\pm$\,0.0000\\
ECGFounder-12lead & 0.4453\,$\pm$\,0.0178 & 0.7150\,$\pm$\,0.0000 & 0.4479\,$\pm$\,0.0000 & 0.5772\,$\pm$\,0.0000\\
ECGFounder-1lead & 0.4399\,$\pm$\,0.0112 & 0.6159\,$\pm$\,0.0000 & 0.3149\,$\pm$\,0.0000 & 0.5660\,$\pm$\,0.0000\\
ECG-LFM & 0.4410\,$\pm$\,0.0097 & 0.6742\,$\pm$\,0.0000 & 0.3679\,$\pm$\,0.0000 & 0.3736\,$\pm$\,0.0000\\
\bottomrule\end{longtable}\endgroup

\begingroup\footnotesize\setlength{\tabcolsep}{4pt}\renewcommand{\arraystretch}{1.15}

\begin{longtable}{@{}p{8.6cm}r@{}}\toprule
\rowcolor{PulsePale}Method & E01e\\\midrule\endhead
\rowcolor{PulsePale}
LingSante-Pulse CLS & 0.6646\,$\pm$\,0.0026\\
\rowcolor{PulsePale}
LingSante-Pulse patch mean & 0.7504\,$\pm$\,0.0043\\
NormWear & 0.6994\,$\pm$\,0.0020\\
PanLUNA & 0.6257\,$\pm$\,0.0035\\
MOMENT-small & 0.7379\,$\pm$\,0.0026\\
xECG & 0.7551\,$\pm$\,0.0006\\
ST-MEM & 0.7377\,$\pm$\,0.0019\\
ECG-JEPA Kim multi-block & 0.7450\,$\pm$\,0.0026\\
ECG-CPC HEEDB & 0.6674\,$\pm$\,0.0076\\
SPOTR-Base & 0.6358\,$\pm$\,0.0020\\
HuBERT-ECG SSL Base & 0.6283\,$\pm$\,0.0047\\
ECGFounder-12lead & 0.7831\,$\pm$\,0.0026\\
ECGFounder-1lead & 0.6988\,$\pm$\,0.0025\\
ECG-LFM & 0.6605\,$\pm$\,0.0026\\
\bottomrule\end{longtable}

\begin{longtable}{@{}p{8.6cm}r@{}}\toprule
\rowcolor{PulsePale}Method & E01f\\\midrule\endhead
\rowcolor{PulsePale}
LingSante-Pulse CLS & 0.3174\,$\pm$\,0.0235\\
\rowcolor{PulsePale}
LingSante-Pulse patch mean & 0.3188\,$\pm$\,0.0161\\
ECG-FM pretrained & 0.3326\,$\pm$\,0.0126\\
NormWear & 0.3041\,$\pm$\,0.0129\\
PanLUNA & 0.3163\,$\pm$\,0.0088\\
MOMENT-small & 0.3200\,$\pm$\,0.0167\\
xECG & 0.2861\,$\pm$\,0.0172\\
ST-MEM & 0.3176\,$\pm$\,0.0199\\
ECG-JEPA Kim multi-block & 0.3374\,$\pm$\,0.0149\\
ECG-CPC HEEDB & 0.2601\,$\pm$\,0.0252\\
SPOTR-Base & 0.3278\,$\pm$\,0.0056\\
HuBERT-ECG SSL Base & 0.2862\,$\pm$\,0.0158\\
ECGFounder-12lead & 0.2911\,$\pm$\,0.0167\\
ECGFounder-1lead & 0.2934\,$\pm$\,0.0065\\
ECG-LFM & 0.3684\,$\pm$\,0.0150\\
\bottomrule\end{longtable}\endgroup

\clearpage

\subsubsection{EEG}

The EEGMMIDB dataset variants form one dataset group; the EEGMAT dataset variants form one group. The BCIC-IV2a dataset uses grouped cross-validation. The Mumtaz dataset recordings are deduplicated and grouped by participant.

\begin{longtable}{@{}p{1.15cm}p{8.0cm}p{4.5cm}@{}}\toprule\rowcolor{PulsePale}ID & Dataset / endpoint & Metric\\\midrule\endhead

B01a & EEGMMIDB / Three-channel MI, two classes & Macro-F1 $\uparrow$\\

B01b & EEGMMIDB / 64-channel MI, four classes & Macro-F1 $\uparrow$\\

B01c & BCIC-IV2a / 22-channel MI, four classes & Macro-F1 $\uparrow$\\

B01d & EEGMAT / Three-channel rest / arithmetic & Macro-F1 $\uparrow$\\

B01e & EEGMAT / 20-channel rest / arithmetic & Macro-F1 $\uparrow$\\

B01f & Mumtaz / Depression state & Macro-F1 $\uparrow$\\

B01g & BCIC2020 \citep{ref73} Speech / Five-class imagined speech & Macro-F1 $\uparrow$\\

\bottomrule\end{longtable}

\begingroup\footnotesize\setlength{\tabcolsep}{4pt}\renewcommand{\arraystretch}{1.15}

\begin{longtable}{@{}p{5.1cm}rrrr@{}}\toprule

\rowcolor{PulsePale}Method & B01a & B01b & B01c & B01d\\\midrule\endhead

\rowcolor{PulsePale}
LingSante-Pulse CLS & 0.6793\,$\pm$\,0.0060 & 0.5301\,$\pm$\,0.0035 & 0.4652\,$\pm$\,0.0095 & 0.6196\,$\pm$\,0.0111\\
\rowcolor{PulsePale}
LingSante-Pulse patch mean & 0.6613\,$\pm$\,0.0018 & 0.5150\,$\pm$\,0.0015 & 0.4360\,$\pm$\,0.0086 & 0.6186\,$\pm$\,0.0105\\
CBraMod & 0.6000\,$\pm$\,0.0008 & 0.4428\,$\pm$\,0.0006 & 0.4340\,$\pm$\,0.0047 & 0.5987\,$\pm$\,0.0061\\
NormWear & 0.6652\,$\pm$\,0.0065 & 0.5124\,$\pm$\,0.0018 & 0.4615\,$\pm$\,0.0075 & 0.5886\,$\pm$\,0.0072\\
PanLUNA & 0.5697\,$\pm$\,0.0044 & 0.3704\,$\pm$\,0.0036 & 0.3354\,$\pm$\,0.0123 & 0.5746\,$\pm$\,0.0158\\
REVE-base & 0.6083\,$\pm$\,0.0023 & 0.3864\,$\pm$\,0.0021 & 0.4156\,$\pm$\,0.0104 & 0.5621\,$\pm$\,0.0124\\
CodeBrain EEGSSM & 0.6592\,$\pm$\,0.0061 & 0.5013\,$\pm$\,0.0029 & 0.4775\,$\pm$\,0.0092 & 0.5897\,$\pm$\,0.0120\\
ST-EEGFormer-base & 0.5300\,$\pm$\,0.0042 & 0.3577\,$\pm$\,0.0033 & 0.3908\,$\pm$\,0.0099 & 0.5794\,$\pm$\,0.0140\\
LaBraM-base & 0.5348\,$\pm$\,0.0049 & 0.2816\,$\pm$\,0.0025 & 0.3228\,$\pm$\,0.0046 & 0.5827\,$\pm$\,0.0122\\
EEGPT (Braindecode conversion) & 0.5412\,$\pm$\,0.0042 & 0.3157\,$\pm$\,0.0048 & 0.3322\,$\pm$\,0.0025 & 0.5497\,$\pm$\,0.0056\\
BIOT & 0.5176\,$\pm$\,0.0037 & 0.2749\,$\pm$\,0.0019 & 0.3007\,$\pm$\,0.0047 & 0.5637\,$\pm$\,0.0104\\
\bottomrule\end{longtable}\endgroup

\begingroup\footnotesize\setlength{\tabcolsep}{4pt}\renewcommand{\arraystretch}{1.15}

\begin{longtable}{@{}p{5.1cm}rrr@{}}\toprule

\rowcolor{PulsePale}Method & B01e & B01f & B01g\\\midrule\endhead

\rowcolor{PulsePale}
LingSante-Pulse CLS & 0.6983\,$\pm$\,0.0116 & 0.8826\,$\pm$\,0.0071 & 0.1998\,$\pm$\,0.0066\\
\rowcolor{PulsePale}
LingSante-Pulse patch mean & 0.6933\,$\pm$\,0.0118 & 0.8786\,$\pm$\,0.0106 & 0.1945\,$\pm$\,0.0037\\
CBraMod & 0.6780\,$\pm$\,0.0143 & 0.8739\,$\pm$\,0.0116 & 0.2031\,$\pm$\,0.0022\\
NormWear & 0.6177\,$\pm$\,0.0133 & 0.8610\,$\pm$\,0.0038 & 0.1912\,$\pm$\,0.0056\\
PanLUNA & 0.5861\,$\pm$\,0.0135 & 0.8740\,$\pm$\,0.0045 & 0.1971\,$\pm$\,0.0043\\
REVE-base & 0.6668\,$\pm$\,0.0048 & 0.8730\,$\pm$\,0.0085 & 0.2118\,$\pm$\,0.0036\\
CodeBrain EEGSSM & 0.6582\,$\pm$\,0.0119 & 0.8671\,$\pm$\,0.0077 & 0.1922\,$\pm$\,0.0030\\
ST-EEGFormer-base & 0.6971\,$\pm$\,0.0170 & 0.8857\,$\pm$\,0.0080 & 0.2016\,$\pm$\,0.0044\\
LaBraM-base & 0.6286\,$\pm$\,0.0128 & 0.8657\,$\pm$\,0.0074 & 0.2068\,$\pm$\,0.0045\\
EEGPT (Braindecode conversion) & 0.5509\,$\pm$\,0.0126 & 0.8328\,$\pm$\,0.0041 & 0.1998\,$\pm$\,0.0051\\
BIOT & 0.5888\,$\pm$\,0.0113 & 0.8759\,$\pm$\,0.0036 & 0.1909\,$\pm$\,0.0024\\
\bottomrule\end{longtable}\endgroup

\clearpage

\subsubsection{EMG}

The DB5 dataset: 10 participants, 53 classes including rest; repetitions 1/3/4/6 train, 2 validation, 5 test. The WESAD dataset: 15 participants, 4,419 10-s chest-EMG windows, subject folds. PhysioWave-EMG Small has a fixed eight-channel encoder; its single-channel WESAD dataset cell is not applicable.

\begin{longtable}{@{}p{1.15cm}p{8.0cm}p{4.5cm}@{}}\toprule\rowcolor{PulsePale}ID & Dataset / endpoint & Metric\\\midrule\endhead

M01a & NinaPro DB5 / 53-class gesture & Macro-F1 $\uparrow$\\

M01b & WESAD / Chest EMG four-state classification & Macro-F1 $\uparrow$\\

\bottomrule\end{longtable}

\begingroup\footnotesize\setlength{\tabcolsep}{4pt}\renewcommand{\arraystretch}{1.15}

\begin{longtable}{@{}p{8.6cm}r@{}}\toprule
\rowcolor{PulsePale}Method & M01a\\\midrule\endhead
\rowcolor{PulsePale}
LingSante-Pulse CLS & 0.0747\,$\pm$\,0.0000\\
\rowcolor{PulsePale}
LingSante-Pulse patch mean & 0.0840\,$\pm$\,0.0000\\
TinyMyo pretrained & 0.0736\,$\pm$\,0.0000\\
NeuroRVQ-EMG & 0.2662\,$\pm$\,0.0001\\
PhysioWave-EMG Small & 0.0450\,$\pm$\,0.0000\\
MOMENT-small & 0.1874\,$\pm$\,0.0000\\
NormWear & 0.0211\,$\pm$\,0.0000\\
\bottomrule\end{longtable}

\begin{longtable}{@{}p{8.6cm}r@{}}\toprule
\rowcolor{PulsePale}Method & M01b\\\midrule\endhead
\rowcolor{PulsePale}
LingSante-Pulse CLS & 0.5110\,$\pm$\,0.0100\\
\rowcolor{PulsePale}
LingSante-Pulse patch mean & 0.5335\,$\pm$\,0.0118\\
TinyMyo pretrained & 0.4386\,$\pm$\,0.0138\\
NeuroRVQ-EMG & 0.4460\,$\pm$\,0.0077\\
MOMENT-small & 0.4413\,$\pm$\,0.0092\\
NormWear & 0.2971\,$\pm$\,0.0024\\
\bottomrule\end{longtable}\endgroup

\paragraph{Continuous hand-trajectory decoding.}

The NinaPro DB8 dataset comprises twelve participants and 36 sessions. Under the A1/A2/A3 protocol, A1 fits the ridge mapping, A2 selects its regularization parameter from 0.1/1/10/100, and A3 provides the test data. MAE and R-squared evaluate the same five-DoA trajectory predictions, which contribute one task to the domain-level ranking.

LingSante-Pulse retains each published 500-ms, 1,000-sample input at 2,000 Hz, pads 24 unobserved samples to 1,024, and maps the final-layer-normalized 32-sample patch features to 50 ten-millisecond target bins using real-sample coverage weights. Patch features from 16 physical channels are concatenated, excluding CLS and all four register tokens. The partitions, ridge solver, original A3 time axis and valid target masks match the baselines. All 6,370,270 valid original A3 samples are scored. This evaluates transfer from the 128-Hz pretraining grid to the published native-rate arrays.

\begin{longtable}{@{}p{1.15cm}p{8.0cm}p{4.5cm}@{}}\toprule\rowcolor{PulsePale}ID & Dataset / endpoint & Metric\\\midrule\endhead

M02a & NinaPro DB8 / Continuous five-DoA trajectory & MAE (glove units) $\downarrow$\\

M02b & NinaPro DB8 / Same five-DoA predictions & R-squared $\uparrow$\\

\bottomrule\end{longtable}

\begingroup\footnotesize\setlength{\tabcolsep}{4pt}\renewcommand{\arraystretch}{1.15}

\begin{longtable}{@{}p{5.1cm}rr@{}}\toprule

\rowcolor{PulsePale}Method & M02a & M02b\\\midrule\endhead

\rowcolor{PulsePale}
LingSante-Pulse patch sequence & 14.3060 & 0.2674\\
TinyMyo pretrained & 14.5514 & 0.2236\\
NeuroRVQ-EMG & 15.0324 & 0.1335\\
\bottomrule\end{longtable}\endgroup

\clearpage

\subsubsection{Dataset-balanced model rankings}\label{app:ranks}

Dataset-balanced mean ranks use the task sets and baseline models listed below. Task ranks are averaged within each dataset, and datasets receive equal weight. PPG, ECG and EEG tables are sorted by patch-mean rank; the EMG table is sorted by its overall mean rank.

\paragraph{PPG}

Included task IDs: P01a, P01b, P01c, P01d, P01e, P01f, P01g, P02a, P02b, P02c, P02d, P02e, P02f, P03a, P03b, P03c, P03d, P03e, P03f, P03g, P05a, P05b.

\begin{longtable}{@{}p{7cm}rr@{}}\toprule\rowcolor{PulsePale}Method & CLS rank $\downarrow$ & Patch-mean rank $\downarrow$\\\midrule\endhead
LingSante-Pulse & 2.551 & 2.316\\
AnyPPG & 2.653 & 2.741\\
PulsePPG & 3.580 & 3.607\\
MOMENT-small & 3.631 & 3.679\\
PaPaGei-P & 5.124 & 5.124\\
SIGMA-PPG & 5.189 & 5.189\\
PaPaGei-S & 5.272 & 5.344\\
\bottomrule\end{longtable}

\paragraph{ECG}

Included task IDs: E01a, E01b, E01c, E01d, E01e, E01f.

\begin{longtable}{@{}p{7cm}rr@{}}\toprule\rowcolor{PulsePale}Method & CLS rank $\downarrow$ & Patch-mean rank $\downarrow$\\\midrule\endhead
LingSante-Pulse & 6.833 & 3.083\\
ECGFounder-12lead & 4.083 & 4.167\\
ECG-JEPA Kim multi-block & 4.000 & 4.333\\
xECG & 4.250 & 4.583\\
ST-MEM & 4.250 & 5.083\\
ECG-LFM & 6.083 & 6.250\\
MOMENT-small & 6.083 & 6.417\\
ECGFounder-1lead & 7.833 & 8.167\\
SPOTR-Base & 8.250 & 8.417\\
NormWear & 8.917 & 9.333\\
PanLUNA & 9.250 & 9.417\\
ECG-CPC HEEDB & 10.250 & 10.667\\
HuBERT-ECG SSL Base & 10.917 & 11.083\\
\bottomrule\end{longtable}

\paragraph{EEG}

Included task IDs: B01a, B01b, B01c, B01d, B01e, B01f, B01g.

\begin{longtable}{@{}p{7cm}rr@{}}\toprule\rowcolor{PulsePale}Method & CLS rank $\downarrow$ & Patch-mean rank $\downarrow$\\\midrule\endhead
LingSante-Pulse & 2.300 & 3.000\\
CBraMod & 3.800 & 3.800\\
ST-EEGFormer-base & 4.600 & 4.500\\
CodeBrain EEGSSM & 4.600 & 4.600\\
REVE-base & 4.600 & 4.600\\
NormWear & 5.700 & 5.400\\
PanLUNA & 6.400 & 6.200\\
LaBraM-base & 6.600 & 6.600\\
EEGPT (Braindecode conversion) & 8.200 & 8.100\\
BIOT & 8.200 & 8.200\\
\bottomrule\end{longtable}

\paragraph{EMG}

Three datasets contribute equally: NinaPro DB5 (M01a, Macro-F1), WESAD (M01b, Macro-F1), and NinaPro DB8 (M02a, MAE). M02b reports a second metric on the same DB8 dataset task and is not counted again. Each task uses its fixed global or patch-sequence readout. The comparison fixes three models with all three results; missing results on the DB8 dataset are not imputed. LingSante-Pulse CLS and patch mean have identical classification ranks and share the patch-sequence result on the DB8 dataset.

\begin{longtable}{@{}p{7cm}rrrr@{}}\toprule\rowcolor{PulsePale}Method & DB5 & WESAD & DB8 & Mean rank\\\midrule\endhead
LingSante-Pulse & 2 & 1 & 1 & 1.333\\
NeuroRVQ-EMG & 1 & 2 & 3 & 2.000\\
TinyMyo pretrained & 3 & 3 & 2 & 2.667\\
\bottomrule\end{longtable}

\subsection{Detailed results for sleep assessment}\label{app:sleep}

The sleep tables retain each task's stated inputs, head and uncertainty summary.

\Needspace{18\baselineskip}
\subsubsection{Multimodal sleep staging on the HMC dataset}

The HMC dataset uses 121/15/15 participants and EEG/EOG/chin-EMG/ECG. Linear staging uses 13,770 30-s test epochs. Overnight SleepFM retains 82,620 native 5-s windows; the other overnight entries use 13,770 30-s epochs. All heads are selected on independent validation.

\begin{longtable}{@{}p{1.1cm}p{8.1cm}p{4.2cm}@{}}\toprule\rowcolor{PulsePale}ID & Endpoint / head & Metric\\\midrule\endhead

S04a & Five-stage staging; linear & Macro-F1\\

S04b & Five-stage staging; overnight BiLSTM & Macro-F1\\

\bottomrule\end{longtable}

\begingroup\footnotesize\setlength{\tabcolsep}{2pt}\renewcommand{\arraystretch}{1.12}

\begin{longtable}{@{}p{4.0cm}>{\raggedleft\arraybackslash}p{2.30cm}>{\raggedleft\arraybackslash}p{2.30cm}@{}}\toprule
\rowcolor{PulsePale}Method & S04a & S04b\\\midrule\endhead
\rowcolor{PulsePale}
LingSante-Pulse CLS & 0.7611\,$\pm$\,0.0025 & 0.7954\\
\rowcolor{PulsePale}
LingSante-Pulse patch mean & 0.7784\,$\pm$\,0.0068 & 0.7859\\
SleepFM Clinical & 0.6018\,$\pm$\,0.0099 & 0.7205\\
OSF & 0.7446\,$\pm$\,0.0035 & 0.7724\\
SleepLM signal encoder & 0.7218\,$\pm$\,0.0028 & 0.7440\\
\bottomrule\end{longtable}

\endgroup

\Needspace{30\baselineskip}
\subsubsection{EEG-only sleep staging}

Strict EEG-only, frozen balanced linear heads with five head fits. The Sleep-EDF20 dataset uses 16/2/2 participants and 11,031 test epochs; the HMC dataset uses the same participant partition as the multimodal staging evaluation. Native preprocessing and declared physical-channel adapters are retained.

\begin{longtable}{@{}p{1.1cm}p{8.1cm}p{4.2cm}@{}}\toprule\rowcolor{PulsePale}ID & Endpoint / head & Metric\\\midrule\endhead

S05a & Sleep-EDF20; frozen linear & Macro-F1\\

S05b & HMC; frozen linear & Macro-F1\\

\bottomrule\end{longtable}

\begingroup\footnotesize\setlength{\tabcolsep}{2pt}\renewcommand{\arraystretch}{1.12}

\begin{longtable}{@{}p{6cm}>{\raggedleft\arraybackslash}p{3.8cm}>{\raggedleft\arraybackslash}p{3.8cm}@{}}\toprule
\rowcolor{PulsePale}Method & S05a & S05b\\\midrule\endhead
\rowcolor{PulsePale}
LingSante-Pulse CLS & 0.7664\,$\pm$\,0.0016 & 0.7559\,$\pm$\,0.0041\\
\rowcolor{PulsePale}
LingSante-Pulse patch mean & 0.7941\,$\pm$\,0.0018 & 0.7678\,$\pm$\,0.0050\\
CBraMod & 0.6167\,$\pm$\,0.0061 & 0.3409\,$\pm$\,0.0231\\
NormWear & 0.6002\,$\pm$\,0.0085 & 0.5991\,$\pm$\,0.0195\\
PanLUNA & 0.6008\,$\pm$\,0.0148 & 0.5906\,$\pm$\,0.0294\\
REVE-base & 0.6904\,$\pm$\,0.0070 & 0.6561\,$\pm$\,0.0070\\
ST-EEGFormer-base & 0.6960\,$\pm$\,0.0053 & 0.6003\,$\pm$\,0.0073\\
LaBraM-base & 0.6148\,$\pm$\,0.0167 & 0.4226\,$\pm$\,0.0260\\
SleepFM Clinical & 0.7265\,$\pm$\,0.0093 & 0.6258\,$\pm$\,0.0043\\
OSF & 0.7891\,$\pm$\,0.0020 & 0.7470\,$\pm$\,0.0025\\
SleepLM signal encoder & 0.7017\,$\pm$\,0.0008 & 0.6785\,$\pm$\,0.0044\\
\bottomrule\end{longtable}

\endgroup

\Needspace{18\baselineskip}
\subsubsection{EOG-only sleep staging}

Strict EOG-only frozen balanced linear staging on the Sleep-EDF dataset and HMC dataset. Five head fits use training-only class weights and validation-only hyperparameter/epoch selection.

\begin{longtable}{@{}p{1.1cm}p{8.1cm}p{4.2cm}@{}}\toprule\rowcolor{PulsePale}ID & Endpoint / head & Metric\\\midrule\endhead

S06a & Sleep-EDF; frozen linear & Macro-F1\\

S06b & HMC; frozen linear & Macro-F1\\

\bottomrule\end{longtable}

\begingroup\footnotesize\setlength{\tabcolsep}{2pt}\renewcommand{\arraystretch}{1.12}

\begin{longtable}{@{}p{6cm}>{\raggedleft\arraybackslash}p{3.8cm}>{\raggedleft\arraybackslash}p{3.8cm}@{}}\toprule
\rowcolor{PulsePale}Method & S06a & S06b\\\midrule\endhead
\rowcolor{PulsePale}
LingSante-Pulse CLS & 0.7139\,$\pm$\,0.0009 & 0.6745\,$\pm$\,0.0071\\
\rowcolor{PulsePale}
LingSante-Pulse patch mean & 0.7299\,$\pm$\,0.0014 & 0.7158\,$\pm$\,0.0039\\
SleepFM Clinical & 0.6907\,$\pm$\,0.0036 & 0.5026\,$\pm$\,0.0051\\
OSF & 0.7252\,$\pm$\,0.0015 & 0.7126\,$\pm$\,0.0021\\
SleepLM signal encoder & 0.4949\,$\pm$\,0.0037 & 0.5470\,$\pm$\,0.0048\\
\bottomrule\end{longtable}

\endgroup

\Needspace{18\baselineskip}
\subsubsection{Chin-EMG-only sleep staging}

Chin-EMG-only staging uses the raw chin-EMG channel from the HMC dataset. Five head fits use training-only class weights and validation-only selection.

\begin{longtable}{@{}p{1.1cm}p{8.1cm}p{4.2cm}@{}}\toprule\rowcolor{PulsePale}ID & Endpoint / head & Metric\\\midrule\endhead

S07a & HMC; frozen linear & Macro-F1\\

\bottomrule\end{longtable}

\begingroup\footnotesize\setlength{\tabcolsep}{2pt}\renewcommand{\arraystretch}{1.12}

\begin{longtable}{@{}p{6cm}>{\raggedleft\arraybackslash}p{3.8cm}@{}}\toprule
\rowcolor{PulsePale}Method & S07a\\\midrule\endhead
\rowcolor{PulsePale}
LingSante-Pulse CLS & 0.4013\,$\pm$\,0.0204\\
\rowcolor{PulsePale}
LingSante-Pulse patch mean & 0.4254\,$\pm$\,0.0022\\
SleepFM Clinical & 0.3572\,$\pm$\,0.0081\\
OSF & 0.3672\,$\pm$\,0.0094\\
\bottomrule\end{longtable}

\endgroup

\Needspace{18\baselineskip}
\subsubsection{Single-lead ECG apnea screening}

Official Apnea-ECG dataset learning/x-test split; c05/c06 remain grouped within learning. LingSante-Pulse uses observed-valid patch mean for each 10-s ECG window, excluding CLS, other prefixes, padding and invalid patches; these vectors form the BiLSTM sequence. A fixed one-layer bidirectional LSTM (100 hidden units per direction) selects context and epoch by validation AUROC from 10/30/60/120/180/300/540 s. Test has 17,268 minutes, including 6,550 positives. Selected contexts are 300 s for LingSante-Pulse, 180 s for ECGFounder-1lead, and 540 s for xECG and ST-MEM. LingSante-Pulse achieves AUROC 0.9183, area under the precision--recall curve (AUPRC) 0.8939, balanced accuracy (BAC) 0.8383 and Macro-F1 0.8443.

\begin{longtable}{@{}p{1.1cm}p{8.1cm}p{4.2cm}@{}}\toprule\rowcolor{PulsePale}ID & Endpoint / head & Metric\\\midrule\endhead

S11a & Minute apnea screening; BiLSTM & AUROC\\

\bottomrule\end{longtable}

\begingroup\footnotesize\setlength{\tabcolsep}{2pt}\renewcommand{\arraystretch}{1.12}

\begin{longtable}{@{}>{\raggedright\arraybackslash}p{5.6cm}>{\raggedleft\arraybackslash}p{2.30cm}@{}}\toprule
\rowcolor{PulsePale}Method & S11a\\\midrule\endhead
\rowcolor{PulsePale}
LingSante-Pulse patch mean & 0.9183\\
xECG & 0.7647\\
ST-MEM & 0.7305\\
ECGFounder-1lead & 0.8089\\
\bottomrule\end{longtable}

\endgroup

\end{document}